\documentclass[trsc,nonblindrev]{informs3} 

\OneAndAHalfSpacedXI 

\usepackage{natbib}
 \bibpunct[, ]{(}{)}{,}{a}{}{,}%
 \def\bibfont{\small}%

\usepackage{bm}
\usepackage{booktabs}
\usepackage{tabularx}
\usepackage{threeparttable}
\usepackage{multicol}
\usepackage{multirow}
\usepackage{setspace}
\usepackage{graphicx}
\usepackage{subfigure}
\usepackage{algorithm} 
\usepackage{algorithmic}
\usepackage{amssymb}
\allowdisplaybreaks[4]
\usepackage[colorlinks=true, allcolors=blue]{hyperref}

\TheoremsNumberedThrough     

\EquationsNumberedThrough    

\begin{document}


\RUNAUTHOR{Authors} 
\RUNAUTHOR{Gao et al.}

\RUNTITLE{MMA for matching and
relocation}


\TITLE{Online Relocating and Matching of Ride-Hailing Services: A Model-Based Modular Approach}

\ARTICLEAUTHORS{%
\AUTHOR{Chang Gao}
\AFF{Department of Industrial Engineering, Tsinghua University \URL{}}
\AUTHOR{Xi Lin}
\AFF{Department of Civil Engineering, Tsinghua University \URL{}}
\AUTHOR{Fang He\thanks{Corresponding author: fanghe@tsinghua.edu.cn}}
\AFF{Department of Industrial Engineering, Tsinghua University \URL{}}
\AUTHOR{Xindi Tang}
\AFF{School of Management Science and Engineering, Central University of Finance and Economics \URL{}}
}  

\ABSTRACT{%
This study proposes an innovative model-based modular approach (MMA) to dynamically optimize order matching and vehicle relocation in a ride-hailing platform. MMA utilizes a two-layer and modular modeling structure. The upper layer determines the spatial transfer patterns of vehicle flow within the system to maximize the total revenue of the current and future stages. With the guidance provided by the upper layer, the lower layer performs rapid vehicle-to-order matching and vehicle relocation. MMA is interpretable, and equipped with the customized and polynomial-time algorithm, which, as an online order-matching and vehicle-relocation algorithm, can scale past thousands of vehicles. We theoretically prove that the proposed algorithm can achieve the global optimum in stylized networks, while the numerical experiments based on both the toy network and realistic dataset demonstrate that MMA is capable of achieving superior systematic performance compared to batch matching and reinforcement-learning based methods. Moreover, its modular and lightweight modeling structure further enables it to achieve a high level of robustness against demand variation while maintaining a relatively low computational cost. 
}%


\HISTORY{}

\maketitle
%
%
\section{Introduction}
The essential role of taxi services in urban transportation is characterized by their provision of personalized and flexible door-to-door services for passengers \citep{aguilera2022ridesourcing}. Conventional taxi services involve idle drivers cruising streets to locate prospective passengers, which results in diminished driver efficiency and exacerbates traffic congestion. The advent of mobile communication technologies has enabled Transportation Network Companies, such as Uber, Didi, and Lyft, to invest significantly in the development of accessible ride-hailing transport services via online platforms, thereby transforming the manner in which individuals travel \citep{shaaban2016assessment}. Online ride-hailing platforms facilitate real-time connections between passengers and drivers, thereby reducing information asymmetry and increasing service efficiency in ride-hailing services. 

This paper focuses on the examination of ride-hailing platforms that operate in dispatching mode, employing online algorithms. Numerous existing studies have demonstrated the profound impact of matching strategies on various aspects of system performance, such as passenger waiting times, vehicle utilization, and platform revenue \citep{vazifeh2018}. In practical settings, ride-hailing demand frequently exhibits distinct spatiotemporal heterogeneity, leading to imbalanced demand-supply relationships throughout the network and the emergence of several unfavorable outcomes. For example, excessive demand relative to supply in specific areas not only prolongs passenger waiting times but also contributes to lost trip requests, consequently undermining the long-term appeal of the platform. Conversely, an overabundance of supply in relation to demand in certain regions results in substantial idle time for drivers and decreased vehicle utilization. Therefore, an effective matching strategy should address these imbalances and improve overall performance. Implementing a filtering process based on established criteria for desirable demands can achieve these objectives. This concept is inspired by numerous studies that focus on optimizing future gains through reinforcement learning (RL) methods \citep{Xu2018}. Moreover, relocating idle vehicles represents another viable approach to attaining future spatial balance. Compared to matching where the vehicle transitions are contingent upon the origins and destinations of the outstanding requests, relocation is a proactive measure employed by ride-hailing platforms to enhance system performance by strategically deploying vacant vehicles to undersupplied areas in anticipation of future demand at the respective destinations. Optimal relocation can lead to reductions in fleet size and increased vehicle utilization. Nevertheless, relocation also generates deadhead miles, necessitating careful optimization. While matching and relocation strategies are typically addressed separately, their interrelated nature requires a simultaneous consideration, as both sets of decisions determine the future spatial distribution of supply. Therefore, this study concentrates on the integrated optimization problem of matching and relocation strategy for ride-hailing services. 

In practical settings, online matching and vehicle relocation present several fundamental challenges. First, ride-hailing system operations exhibit high levels of dynamism and stochasticity, primarily due to time-of-day characteristics of human travel patterns and the inherently random nature of demand and supply generation processes. Second, both supply and demand within platforms own considerable elasticity in practice, as drivers may choose to exit platforms and passengers may opt to cancel ride requests. Third, a strong endogeneity exists between current decision-making and the future evolution of the system, as strategies employed in the present stage significantly impact the temporal and spatial distribution of demand and supply in subsequent stages. Fourth, the process of matching and relocation necessitates an online approach. Decisions must be made expeditiously, typically within a matter of seconds, while the real-life decision-making process is prone to complications arising from the curse of dimensionality. Consequently, it is imperative that matching and relocation algorithms prove scalable and computationally feasible for practical implementation.

To tackle the aforementioned challenges, we propose an innovative model-based modular approach (MMA) to maximize the total profit, which is equal to the revenue from fulfilling service orders minus the cost of vehicle relocation. MMA utilizes a two-layer, step-by-step modular decision structure. The upper layer, operating on a larger timescale (e.g., five-minute intervals), integrates factors such as supply-demand prediction, vehicle relocation costs, and order prices to intelligently determine the spatial transfer patterns of vehicle flow within the system, aiming to maximize the total revenue of the current and future stages. Utilizing the spatial transfer patterns provided by the upper layer as guidance, the lower-layer model performs rapid vehicle-to-order matching within each matching-decision cycle (lasting several seconds). Each strategic-decision interval of the upper-layer model consists of a number of matching intervals of the lower-layer model, and the vehicle relocation is performed at the end of the last matching interval within each strategic interval. Essentially, the upper-layer model is formulated as a mixed-integer linear programming (MILP) problem, and we design a Lagrangian-relaxation-based (LR) algorithm to efficiently solve it to offer strategic and system-level matching and relocation recommendations through vehicle flow spatial transfer patterns. We theoretically prove that the LR algorithm yields a zero optimality gap for stylized cases, while its superior performance for general cases is numerically validated. We propose customized and polynomial-time algorithms for the lower-layer model. The main contributions of this work are summarized as follows. 

\begin{enumerate}
    \item From the modeling perspective, MMA utilizes a modular, two-layer, and lightweight modeling structure, where the model-free part requiring offline training is only limited to the supply-demand prediction within aggregated time intervals, and the other parts are all model-based equipped with polynomial solution algorithms. It decouples the complex and stochastic system and is capable of solving the large-scale online matching and relocation problem. Compared with the algorithms based on the RL approach, the MMA framework is of interpretability, and easy to implement, which is crucial for developing practical and industry-level applications. MMA effectively balances algorithm interpretability, scalability, and performance. We conduct extensive numerical experiments based on real-world data to verify the effectiveness and practicability of the MMA framework. MMA achieves remarkable improvements in system performance. In relocation-free scenarios, compared with the batch matching method, it can increase the requests completion rate by $10.79\%$. In the scenarios with relocation, MMA outperforms the RL method by a significant margin, with more than $3.09\%$ improvement in the number of completed requests. 

    \item We propose a polynomial-time LR method to efficiently search for the near-optimal solutions for the large-scale instances of the decision-making model in the strategic upper layer. In the LR method, the relaxed problem provides an upper bound, and a dedicated heuristic is proposed to provide a tight lower bound. We theoretically prove that the LR method is capable of delivering globally optimal strategies in stylized networks. In addition, the numerical studies show that it can always find satisfactory matching and relocation strategies with the objective function gap between the CPLEX solver and the LR method at around $0.6\%$ within a practically-implementable solution time in the realistic network. The execution lower-layer model assigns vehicles to requests in each matching interval such that the cumulative matched requests with specific O-D pairs in a strategic interval are consistent with the guidance suggested by the strategic layer. We design a two-step mechanism and devise a customized algorithm with polynomial time complexity to achieve this goal. With the specialized algorithms, the MMA framework is capable of scaling past thousands of vehicles in an online fashion. 

    \item With a modular, two-layer, and lightweight modeling structure, the upper and lower layers in MMA operate flexibly and interact effectively. The lower layer's decision-making mechanism is guided by, but not restricted to specific spatial transfer pattern guidance. The matching and relocation results determined by the lower layer distinctly shape the temporal and spatial distribution of demand and supply in subsequent stages, thereby exerting a pronounced influence on the decision-making processes in the upper layer. When sudden events occur in the traffic system, leading to significant changes in supply and demand, the relevant information can be promptly captured by the upper-layer model because the embedded prediction model can incorporate the multi-dimension temporal demand information including the one in adjacent intervals of the day. The intelligent decision-making process rapidly updates the spatial transfer patterns of vehicle flow. The updated transfer mode can guide the lower layer's vehicle operations to respond quickly, significantly enhancing the robustness of the system's matching and relocation strategy. In essence, the modular and flexible structure of MMA enables it to achieve a high level of robustness against demand variation while maintaining a relatively low computational cost. Numerical experiments based on the realistic dataset show that in the presence of irregular events, MMA with relocation further increases the number of completed requests by $4.69\%$ compared to the RL-based method with relocation. Moreover, it is also observed that MMA demonstrates superior resistance to the adverse impact of inaccurate prediction of platforms’ supply-demand. As prediction accuracy declines, it can still maintain an over 10$\%$ increase in completed requests compared to the batch matching method.


\end{enumerate}

The rest of this paper is organized as follows. Section \ref{literature} reviews the related literature. Section \ref{method} presents the overall framework of MMA. Section \ref{strategic} and \ref{execution} elaborate on the two-layer models. Section \ref{numerical} conducts extensive numerical experiments on both a toy network and a realistic network. Finally, Section \ref{conclusion} concludes the paper.

\section{Literature Review}
\label{literature}

Order dispatching and matching are core issues in the operations of ride-hailing services, and the utilized algorithms significantly affect the overall efficiency. Generally, two main categories of methods are utilized in practice, i.e., greedy matching and combinatorial optimization. Greedy algorithms, such as matching customers with the nearest vacant vehicles immediately or utilizing the first-come-first-serve (FCFS) principle \citep{Lee2004,jung2013design}, have been widely employed due to their ease of implementation. However, these algorithms neglect the system-level efficiency and result in the ``wild goose chase'' phenomenon where drivers have to travel long distances to pick up a customer when few vehicles are available \citep{Castillo2017}. 

A set of algorithms based on combinatorial optimization has been proposed to avoid suboptimal assignments and to prevent ``wild goose chase''. Order dispatching is typically formulated as a bipartite matching problem \citep{AGATZ20111450,QIN2021103287}. \citet{Zhang2017} utilized statistical learning methods to estimate the probability of an order being accepted by a particular driver. Based on the estimation, the total expected acceptance probability was maximized by the hill-climbing algorithm. Batch matching is another conventional approach, which accumulates waiting passengers and vacant vehicles for a few seconds, and then allocates them in a coordinated manner. The matching interval (a preset time duration to accumulate the unmatched passengers and drivers) and the matching radius (the maximum allowed pick-up distance) are critical parameters in batch matching strategy, which influence the expected pick-up distance, passenger waiting times, and matching rate. \citet{Yang2020} developed a spatial probability model to characterize batch matching process and theoretically derived the optimal matching interval and matching radius under various supply-demand conditions. Despite optimizing system performance within the current time interval, batch matching may be myopic and fail to provide satisfactory performance over a long time horizon. In view of the stochastic and dynamic nature of the problem, some researchers have attempted to adopt reinforcement learning approaches \citep{wang2018deep,shi2019operating}. \citet{Xu2018} developed a learning and planning method wherein the spatiotemporal quantity indicating the expected value of a driver being in a particular state was evaluated in the offline learning stage. In the online planning stage, the system optimizes the concrete assignment based on immediate rewards and future gains. A deep value network-based approach was proposed by \citet{Tang2021}, where Cerebellar value networks and the Lipschitz regularization schema were introduced to ensure the robustness and stability of the value iteration during policy evaluation. Although reinforcement learning places emphasis on long-term profit to avoid short-sighted decisions, its poor interpretability hinders its implementation in practice.

Appropriate relocation can improve vehicle utilization and platform profits by resolving the demand-supply imbalance. A growing body of literature employs optimization models to design relocation strategies. \citet{braverman2019empty} presented a fluid-based model in a queueing network to obtain the optimal vacant vehicle routing policy. In recent years, reinforcement learning has gained increasing popularity as a method to optimize the vacant vehicles' sequential decisions and vehicle relocation policies \citep{yu2019markov,liu2022deep}. \citet{lin2018efficient} proposed a contextual multi-agent reinforcement learning framework for relocating available vehicles to locations with larger demand-supply gaps. \citet{tang2020} proposed a novel \textit{advisor-student} reinforcement learning framework for assigning vehicles to serve demands, relocate, and get recharged at charging stations intelligently in the context of automated electric vehicles. In addition, rolling horizon control is frequently utilized to optimize relocation policies by incorporating explicit demand estimation to capture the complicated stochastic demand-supply variations and avoid short-term decisions \citep{Miao2016,Iglesias2018}. \citet{8105899} developed a data-driven vehicle dispatching framework that considered spatial-temporal demand uncertainty on a rolling horizon basis. A matching-integrated vehicle rebalancing model proposed by \citet{guo2021} adopted the rolling horizon manner and robustness optimization to develop the vehicle relocation strategies, while considering the impact of relocation on driver-customer matching. Despite these well-designed works addressing various challenges, several issues remain unresolved. The first aspect pertains to the joint optimization of order dispatching and vehicle relocation. In practice, matching and relocation decisions are highly interdependent and simultaneously affect the distribution of supply. However, most studies tend to investigate matching and relocation as separate entities, with only a limited number emphasizing the close interconnection between these two tasks. Consequently, the integration of matching and relocation strategies at the real-time operational level remains a challenging endeavor. Furthermore, the aforementioned algorithms mostly assume a fixed vehicle fleet, which overlooks the stochasticity of drivers' behavior. The third one is to design an interpretable and farsighted algorithm. The RL-based methods lack interpretability since they optimize long-term efficiency based on the value function, while the training process entails a lot of computation and data, which limits the application in realistic scenarios. To address the aforementioned issues, this paper presents a dedicated model-based approach that aims to resolve the joint optimization problem, encompassing matching and relocation strategies, while also accounting for the inherent stochastic nature of participants' behaviors.

Short-term demand forecasting and driver behavior analysis are crucial elements in the development of efficient operational strategies. Extensive research has been conducted on predicting demand distribution, employing techniques such as time-series forecasting methods
\citep{li2012prediction,moreira2012predictive} and regression models \citep{yang2014modeling,Tong2017}. The emergence of machine learning-based approaches has significantly enhanced the precision of demand prediction \citep{geng2019spatiotemporal, ke2021predicting}. For instance, \citet{Ke2017} proposed a fusion convolution long short-term memory network to address the temporal, spatial and exogenous dependencies. Another stream of related work involves modeling the online-offline behavior of vehicles. Most drivers are self-employed and are responsible for determining when to work \citep{Chen2017}. This flexibility enables drivers to obtain higher profits by working during more favorable periods; however, it also poses challenges for platforms in effectively managing the labor force \citep{Wang2019}. Therefore, the characterization of driver behavior holds paramount importance in the operational processes. Previous research by \citet{farber2005tomorrow} suggests that drivers' decision to cease working at a specific time primarily depends on their cumulative daily hours. Furthermore, \citet{Sun2019} used economic models to investigate the impact of hourly wages on labor supply.

\section{Methodological Framework}
\label{method}
\begin{figure}[hbt!]
\centering
\includegraphics[width=0.85\textwidth]{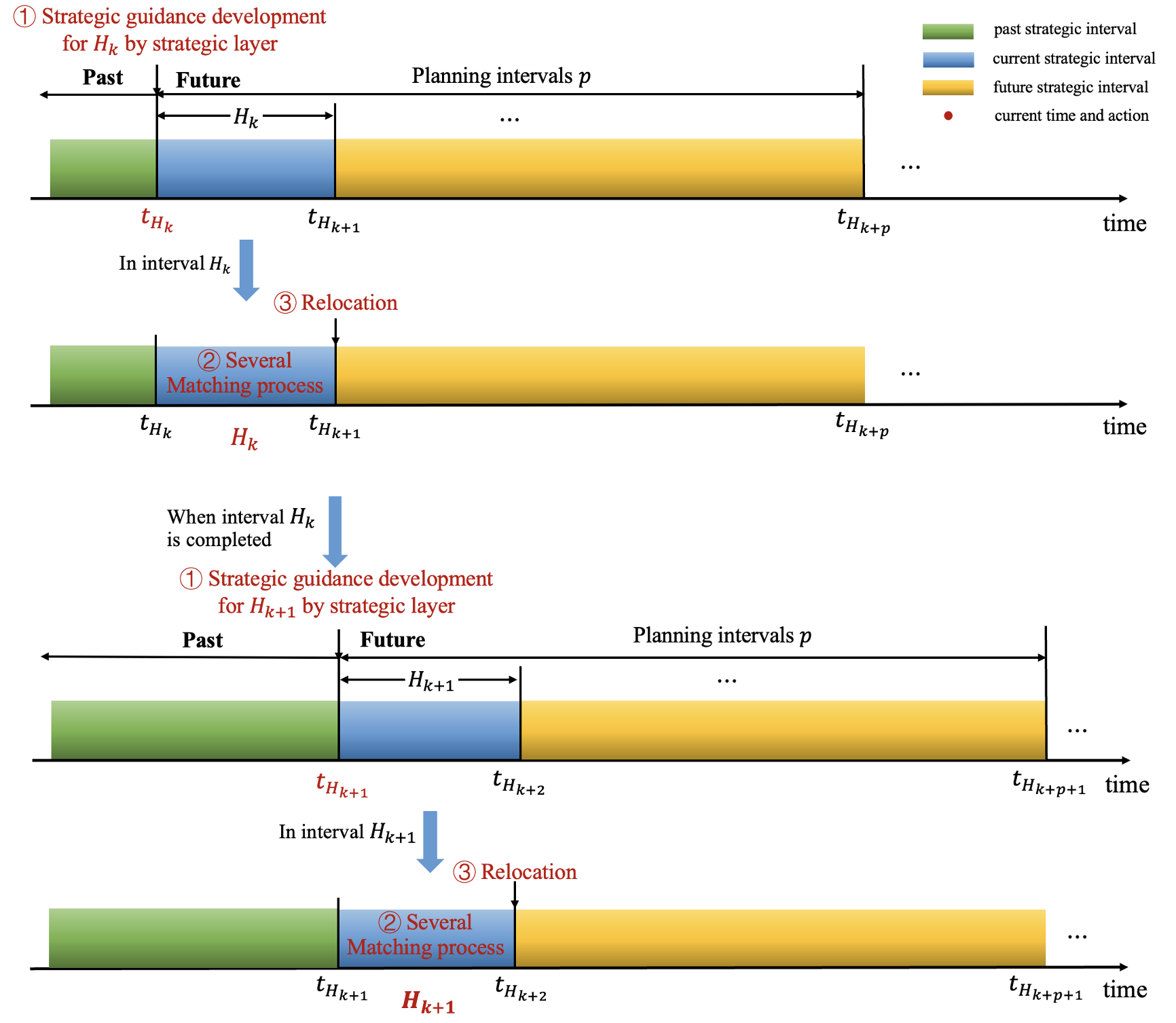}
\caption{\label{fig:operation_framework} Two-layer model-based modular approach (MMA) framework.\\}
\end{figure}

MMA adopts a novel two-layer framework based on the rolling-horizon optimization schema. The study region is partitioned into a set of disjoint zones, which are generally hexagonal grids. The set of all hexagonal grids is denoted by $\mathcal{R}$. The time domain is discretized into strategic intervals of equal duration, and $\mathcal{T}=\{1,2,\dots,\bar{t}\}$ represents the set of all the intervals. At the beginning of each interval, the upper layer (the strategic layer) first develops the strategic guidance for matching and relocation operations. The rolling horizon mechanism is adopted to avoid myopic decisions by incorporating explicit forecast information including the number of vacant vehicles and waiting demands in the subsequent intervals, estimated via the prediction model. The optimized strategic guidance encompasses the total number of vehicles to be matched to requests associated with each O-D pair and to be relocated to each hexagon for current and subsequent intervals within the rolling-horizon regime. Then, only the strategic guidance for the current interval is input into the lower-layer (execution-layer) model. The duration of each strategic interval should take into account the uncertainty of demand and supply, as excessively short intervals may result in inaccurate predictions. Moreover, for the effectiveness of dynamic decision-making, it is also advisable to avoid overly long time intervals. We opt for a duration ranging between five and 15 minutes for each strategic interval in this study. The matching of vehicles and passengers should occur every few seconds. Therefore, within each strategic interval of the strategic layer, there should be multiple matching intervals, and within each matching interval, the execution-layer model utilizes batch matching to assign vehicles to waiting demands in predetermined rules such that the accumulative assignment results are consistent with the strategic guidance from the strategic-layer model. At the end of each strategic-layer interval, the remaining vacant vehicles are relocated as per the strategic guidance. The time horizon rolls forward and the decision-making process as depicted above will be repeated, as shown in Figure \ref{fig:operation_framework} where $H_k$ represents the interval $k$ in the strategic-layer model. 

\begin{figure}[!htb]
\centering
\includegraphics[width=0.8\textwidth]{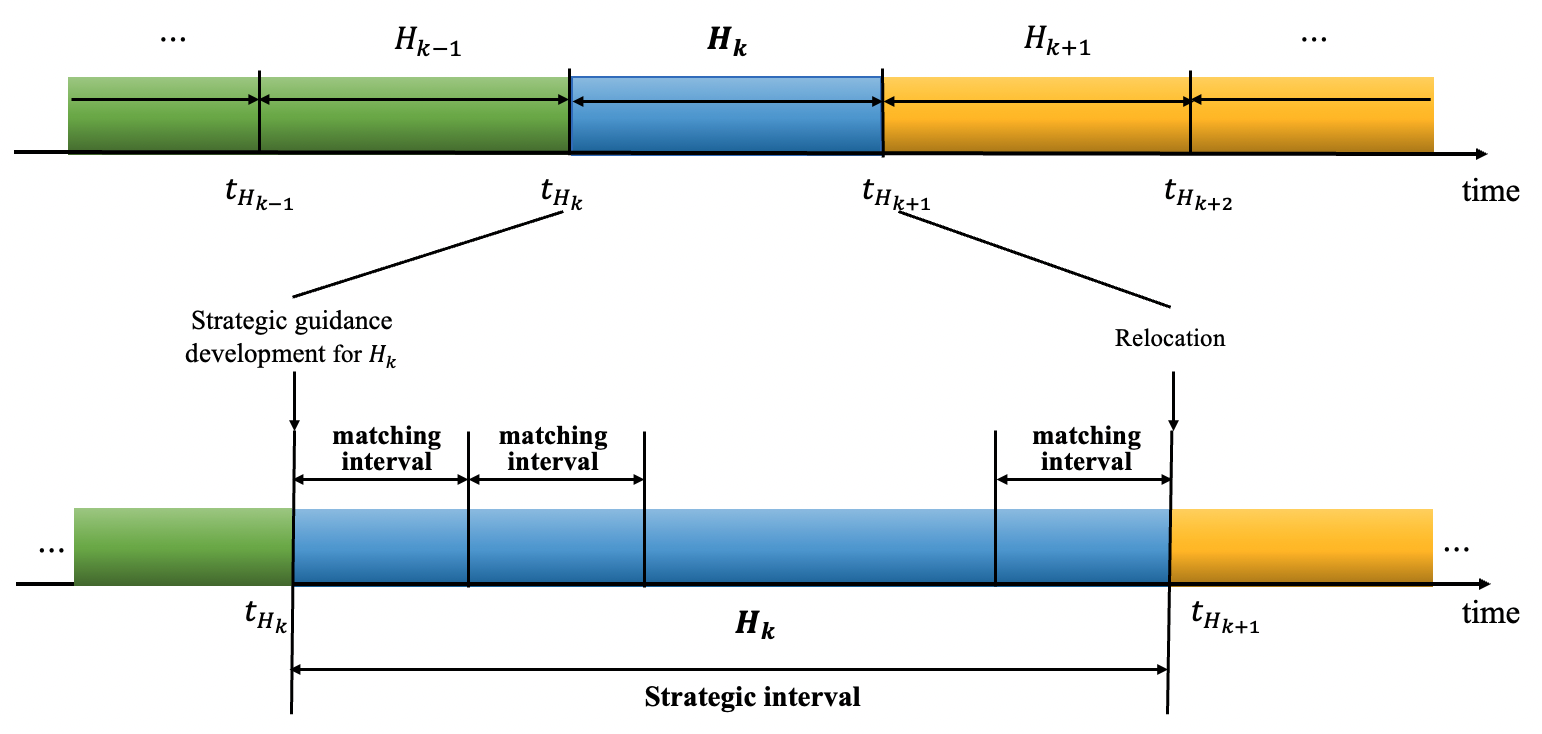}
\caption{\label{fig:time_horizon}Illustration of strategic intervals and matching intervals.}
\end{figure}

We note that each strategic interval in the strategic-layer model consists of several equal-length matching intervals as illustrated in Figure \ref{fig:time_horizon}. The details of the methodological components are presented in Section \ref{strategic} and \ref{execution}. The key notations are summarized in Table \ref{Notataion_tab} of appendix.

\section{The Strategic Layer}
\label{strategic}
\subsection{Prediction of Demand and Supply}
To assist in the development of strategic guidance in strategic interval $k$, the prediction on future demand and supply in interval $t \in \mathcal{T}_{k \rightarrow p}$ are required, and $\mathcal{T}_{k \rightarrow p} = \{k,k+1,\dots, k+p-1\}$ represents the set of $p$ consecutive time intervals starting from interval $k$. The predicted information includes the number of newly-emerging vehicles $\widehat{N}_{t,r}^{s}$, the number of newly-generated demands $\widehat{N}_{t,r}^{d}$ in hexagon $r$, the proportion of demands towards hexagon $j$ from $i$, $b_{i \rightarrow j}^{t}$, and the dropping rates of unmatched vehicles and demands, denoted by $\mu_{t}^{s}$ and $ \mu_{t}^{d}$, respectively. In this study, we consider that the accessible historical dataset encompasses vehicular trajectory and state data (vacant, occupied, and offline), as well as the generation time, origin, and destination of each order request. 

We adopt a series of  Lasso regression models with massive explanatory variables \citep{tibshirani1996regression} to predict the number of newly-emerging demands in hexagon $r$ in interval $t$ for any $r \in \mathcal{R}, t \in \mathcal{T}_{k \rightarrow p}$. It is worth noting that the proposed MMA framework is not limited to the use of Lasso; in fact, any other prediction model can be employed within MMA, provided that it exhibits satisfactory predictive performance. Lasso performs variable selection and regularization to enhance the prediction accuracy and interpretability. We denote the number of newly-emerging demands in hexagon $r$ in interval $k-j$ on the same day of week in the past $i$ weeks as $x_{i,(k-j),r}$. To predict the number of newly-emerging demands in hexagon $r$ in the next $h$ intervals, $h \in \{0,1,..,p-1\}$,  at the beginning of interval $k$, the explanatory variables $\vec{\bm{x}}_{(k+h),r}^{d}$ consist of the features that reveal the historical pattern and present-day variation, including (1) the number of newly-emerging demands in interval $k+h$ on the same day of the last four weeks in hexagon $r$, $x_{1,(k+h),r},x_{2,(k+h),r},x_{3,(k+h),r},x_{4,(k+h),r}$; and (2) the number of newly-emerging demands in the last six intervals on the current day and corresponding day of the last four weeks in hexagon $r$, $x_{i,(k-1),r},x_{i,(k-2),r},x_{i,(k-3),r},x_{i,(k-4),r},x_{i,(k-5),r},x_{i,(k-6),r}, i \in \{0,1,2,3,4\}$. Then, the predicted number of newly-emerging demands in interval $k+h$ in hexagon $r$, $\hat{N}_{(k+h),r}^{d}$ can be calculated as $\hat{N}_{(k+h),r}^{d} = (\vec{\bm{\zeta}}^{h})^{'} \vec{\bm{{x}}}_{(k+h),r}^{d}$, where $\vec{\bm{\zeta}}^{h}$ is the parameter vector to be learned through the equation  $\vec{\bm{\zeta}}^{h} = \mathop{\arg\min}_{\vec{\bm{\zeta}}}\sum_{i \in \mathcal{M}}\sum_{r \in \mathcal{R}}(\hat{N}_{(i+h),r}^{d}-N_{(i+h),r}^{d})^{2} + \lambda_1 \Vert \vec{\bm{\zeta}} \Vert_{1}$, and  $\lambda_1$ is a super-parameter for the $L_1$ regularization, and $\mathcal{M}$ represents the historical interval set. Likewise, the prediction of newly-emerging vehicles follows the same logic as that of the demand prediction. The explanatory variables utilized to predict the number of newly-emerging vehicles in interval $k+h \in \mathcal{T}_{k \rightarrow p}$ in hexagon $r$, $\widehat{N}_{(k+h),r}^{s}$, include the number of newly-emerging vehicles in the last six intervals on the current day and on the same day of last week, as well as the number of newly-emerging vehicles in interval $k+h$ on the same day of last week. We note that the choice of explanatory variables can be flexibly adjusted based on the available dataset. 

The proportion of demand from hexagon $i$ to $j$ in interval $t$, $b_{i \rightarrow j}^{t}$, is contingent upon the day of the week. Generally, the week is partitioned into four distinct categories, namely, (1) Monday, (2) Tuesday through Thursday, (3) Friday, and (4) the weekend, with the categorization founded on the principle of analogous demand patterns for each day within a given category. The transition proportion $b_{i \rightarrow j}^{t}$ is estimated by the historical transition frequency. 

In practice, an increase in waiting time may lead to the impatience of certain passengers and vehicles, resulting in their departure from the matching pool. To more accurately reflect real-life scenarios, the proposed framework incorporates abandonment behaviors for both passengers and vehicles. Consequently, it is assumed that the maximum waiting time for passengers and vehicles follows an exponential distribution with varying rate parameters. The proportion of passengers and vehicles, denoted as $\mu_{t}^{d}$ and $\mu_{t}^{s}$, respectively, that exit the matching pool during horizon $t$ with length $\Delta T$, can be inferred from historical data.

\subsection{Strategic-layer Decision-making Model}
Given the prediction results $\widehat{N}_{t,r}^{s}$, $\widehat{N}_{t,r}^{d}$, $b_{i \rightarrow
 j}^{t}$, $\mu_{t}^{s}$ and $\mu_{t}^{d}$ for interval $t \in \mathcal{T}_{k \rightarrow p}$, the strategic-layer model determines the matching and relocation strategic guidance for the entire planning intervals $\mathcal{T}_{k \rightarrow p}$. The strategic guidance encompasses the number of vehicles to serve the passengers from hexagon $i$ to hexagon $j$ in interval $t$, $M_{i,j}^{t}$, and the number of idle vehicles to be relocated from hexagon $i$ to hexagon $j$ in interval $t$, $E_{i,j}^{t}$. The performance of strategic-layer model is measured by the total number of completed requests and cost associated with relocation in the planning intervals, leading to a multi-objective optimization. The weighted-sum technique is adopted to scalarize them into a single objective optimization. To characterize the optimal matching and relocation strategic guidance, a mixed integer linear programming is formulated on the basis of the prediction information. The notations are summarized in Table~\ref{Notataion_tab} of the appendix. The decision-making model at the current strategic interval $k$ is subject to the following constraints.

(a) Supply constraints: constraints \eqref{supply_eq5} -- \eqref{supply_eq7} define and restrict the number of vacant vehicles in strategic interval $t$ in hexagon $r$, where $t \in \mathcal{T}_{k\rightarrow p}$ and $r \in \mathcal{R}$.
\begin{align}
& N_{k,r}^{s} = \widehat{N}_{k,r}^{s} + l_{(k-1),r}^{s}+ \tilde{E}_{k,r}^{s} + \tilde{O}_{k,r}^{s} & \forall r \in \mathcal{R} \label{supply_eq5} \\
& N_{t,r}^{s} = \widehat{N}_{t,r}^{s} + (L_{(t-1),r}^{s}-\sum_{j\in\mathcal{R}}E_{r,j}^{t-1})(1-\mu_{t-1}^{s}) + \tilde{E}_{t,r}^{s} + \tilde{O}_{t,r}^{s} & \forall t \in \mathcal{T}_{k\rightarrow p} \setminus \{k\}, r \in \mathcal{R} \label{supply_eq6} \\
& \tilde{E}_{t,r}^{s} = \check{E}_{t,r}^{s} + \check{e}_{t,r}^{s} & \forall t \in \mathcal{T}_{k\rightarrow p}, r \in \mathcal{R} \label{supply_eq3} \\
& \check{E}_{t,r}^{s} = \sum_{j \in \mathcal{R}} E_{j,r}^{t-a[jr]} & \forall t \in \mathcal{T}_{k\rightarrow p}, r \in \mathcal{R}  \label{supply_eq1} \\
& \tilde{O}_{t,r}^{s} = \check{O}_{t,r}^{s} + \check{o}_{t,r}^{s} & \forall t \in \mathcal{T}_{k\rightarrow p}, r \in \mathcal{R} \label{supply_eq4} \\
& \check{O}_{t,r}^{s} = \sum_{j \in \mathcal{R}} M_{j,r}^{t-a[jr]} & \forall  t \in \mathcal{T}_{k\rightarrow p}, r \in \mathcal{R}  \label{supply_eq2} \\
& N_{t,r}^{s} = L_{t,r}^{s} + F_{t,r} & \forall  t \in \mathcal{T}_{k\rightarrow p}, r \in \mathcal{R} \label{supply_eq7}
\end{align}

Constraints \eqref{supply_eq5} suggests that the vacant vehicles in hexagon $r$ in interval $k$ consist of the newly-emerging vacant vehicles $\hat{N}_{k,r}^{s}$, vehicles remaining from the previous interval, $l_{(k-1),r}^{s}$, relocated vehicles $\tilde{E}_{k,r}^{s}$ and occupied vehicles that complete the requests $\tilde{O}_{k,r}^{s}$. Constraints \eqref{supply_eq5} and \eqref{supply_eq6} are the same except that in interval $k$, the number of vehicles remaining from the previous interval $k-1$, $l_{(k-1),r}^{s}$, is known, whereas the corresponding numbers of future intervals are all variables and calculated by $(L_{(t-1),r}^{s}-\sum_{j\in\mathcal{R}}E_{r,j}^{t-1})(1-\mu_{t-1}^{s})$, and $L_{(t-1),r}^{s}$ denotes the number of vehicles that are not matched to customers in interval $t-1$, and $E_{r,j}^{t-1}$ represents the number of vehicles relocated from hexagon $r$ to hexagon $j$ at the end of interval $t-1$. Constraint \eqref{supply_eq3} indicates that the number of relocated vehicles arriving at hexagon $r$ in interval $t$, $\tilde{E}_{t,r}^{s}$, is composed of the vehicles relocated before interval $k$, $\check{e}_{t,r}^{s}$, which is known and was already decided in previous intervals, and the vehicles relocated from other hexagons after interval $k$, $\check{E}_{t,r}^{s}$, which is a decision variable . This study assumes that the number of intervals required to travel from hexagon $i$ to $j$, denoted as $a[ij]$, is deterministic. Then, the vehicles that are relocated or matched in hexagon $j$ in interval $t-a[jr]$ and head for hexagon $j$  will arrive at hexagon $r$ in interval $t$. Accordingly, constraint \eqref{supply_eq1} calculates $\check{E}^{s}_{t,r}$ as the summation of the vehicles that are relocated from other hexagons after interval $k$ and arrive at hexagon $r$ in interval $t$. Constraints \eqref{supply_eq4} -- \eqref{supply_eq2} are the same as constraints \eqref{supply_eq3} -- \eqref{supply_eq1} except that they are for the occupied vehicles, and the variable $M_{j,r}^{t-a[jr]}$ denotes the quantity of vehicles that are successfully matched in hexagon $j$ during interval $t-a[jr]$ and are heading for hexagon $r$. Constraint \eqref{supply_eq7} indicates that the vacant vehicles are either matched or 
unmatched to customers, where $F_{t,r}$ denotes the number of matched demands in hexagon $r$ in interval $t$. 

(b) Demand constraints: constraints \eqref{demand_eq1} -- \eqref{demand_eq3} define and constrain the number of customers in interval $t$ in each hexagon.
\begin{align}
& N_{k,r}^{d} = \widehat{N}_{k,r}^{d} + \sum_{j \in \mathcal{R}}l_{(k-1),r,j}^{d}  & \forall r \in \mathcal{R} \label{demand_eq1} \\
& N_{t,r}^{d} = \widehat{N}_{t,r}^{d} + \sum_{j \in \mathcal{R}}L_{(t-1),r,j}^{d}(1-\mu_{t-1}^{d})  & \forall t \in \mathcal{T}_{k\rightarrow p} \setminus \{k\}, r \in \mathcal{R} \label{demand_eq2} \\
& N_{t,r}^{d} = \sum_{j \in \mathcal{R}}L_{t,r,j}^{d} + F_{t,r} & \forall t \in \mathcal{T}_{k\rightarrow p}, r \in \mathcal{R} \label{demand_eq3}
\end{align}

Constraints \eqref{demand_eq1} and \eqref{demand_eq2} show that the waiting customers in specific hexagons and intervals consist of the newly-emerging demands and the unmatched demands from the previous interval. Constraint \eqref{demand_eq3} dictates that the customer demands are either matched or unmatched to vehicles, where $L_{t,r,j}^{d}$ represents the number of unmatched customers at the end of interval $t$ in hexagon $r$ heading for hexagon $j$.

(c) Variables constraints: constraints \eqref{variable_eq1} -- \eqref{variable_eq9} restrict the number of vacant vehicles matched or relocated from hexagon $r$ to hexagon $j$ in interval $t$ to maintain the consistency with reality.
\begin{align}
& \sum_{j \in \mathcal{R}} E_{r,j}^{t} \le L_{t,r}^{s} & \forall t \in \mathcal{T}_{k\rightarrow p}, r \in \mathcal{R} \label{variable_eq1}\\
& M_{r,j}^{k} \le l_{(k-1),r,j}^d + \widehat{N}_{k,r}^{d}b_{r \rightarrow  j}^{k} & \forall r \in \mathcal{R},j\in \mathcal{R} \label{variable_eq2} \\
& M_{r,j}^{t} \le L_{(t-1),r,j}^d(1-\mu_{t-1}^{d}) + \widehat{N}_{t,r}^{d}b_{r \rightarrow j}^{t} & \forall t \in \mathcal{T}_{k \rightarrow p} \setminus \{k\}, r \in \mathcal{R},j \in \mathcal{R} \label{variable_eq3} \\
& L_{k,r,j}^{d} = l_{(k-1),r,j}^{d} + \widehat{N}^{d}_{k,r}b_{r \rightarrow j}^{k} - M_{r,j}^{k} & \forall r \in \mathcal{R}, j \in \mathcal{R} \label{variable_eq4} \\
& L_{t,r,j}^{d} = L_{(t-1),r,j}^{d}(1-\mu_{t-1}^{d}) + \widehat{N}^{d}_{t,r}b_{r \rightarrow j}^{t} - M_{r,j}^{t} & \forall t \in \mathcal{T}_{k \rightarrow p} \setminus \{k\}, r \in \mathcal{R},j \in \mathcal{R} \label{variable_eq5} \\
& F_{t,r} = \sum_{j \in \mathcal{R}} M_{r,j}^{t} & \forall t \in \mathcal{T}_{k\rightarrow p}, r \in \mathcal{R} \label{variable_eq6} \\
& F_{t,r} = \min(N_{t,r}^{d},N_{t,r}^{s}) & \forall t \in \mathcal{T}_{k\rightarrow p}, r \in \mathcal{R} \label{variable_eq7} \\
& M_{i,j}^{t}, E_{i,j}^{t}, L_{t,i,j}^{d} \ge 0 & \forall t \in \mathcal{T}_{k\rightarrow p}, i \in \mathcal{R}, j \in \mathcal{R} \label{variable_eq8} \\
& L_{t,r}^{s} \ge 0 & \forall t \in \mathcal{T}_{k\rightarrow p}, r \in \mathcal{R} \label{variable_eq9} 
\end{align}

Constraint \eqref{variable_eq1} ensures that the number of vehicles relocated from hexagon $r$ in interval $t$ cannot exceed the number of available vehicles at the end of interval $t$. Constraints \eqref{variable_eq2} and \eqref{variable_eq3} restrict the number of matched demands from $r$ to $j$ in interval $t$ should be no greater than the waiting demands towards $j$ from $r$ in interval $t$, which are composed of two parts, namely, the remaining demands from the previous interval, $l_{(k-1),r,j}^d$ or $L_{(t-1),r,j}(1-\mu^{d}_{t-1})$, and the newly-emerging demands $b_{r \rightarrow j}^{t}\widehat{N}_{t,r}^{d}$. Constraints \eqref{variable_eq4} and \eqref{variable_eq5} dictate that demands are either matched or unmatched. Constraint \eqref{variable_eq6} shows that the number of matched demands in hexagon $r$ in interval $t$ is equal to the summation of the matched demands heading for all hexagons from hexagon $r$. The MMA framework requires that vacant vehicles and waiting passengers are fully matched, which implies that $F_{t,r} = \min(N_{t,r}^{d},N_{t,r}^{s})$, as shown in constraint \eqref{variable_eq7}. Constraints \eqref{variable_eq8} and \eqref{variable_eq9} are non-negative constraints. We note that considering that the strategic-layer model only provides guidance on vehicle operations, for computational tractability, we do not impose integer-variable constraints. 

The objective function is comprised of three components: (1) the total number of completed trips; (2) the cost of relocation; and (3) the penalties associated with the regional imbalance between supply and demand. We adopt a weighted-sum function to trade off among three objectives, whose mathematical expression is shown as follows. 
\begin{align}
& \mathnormal{P}(\vec{\bm{M}},
\vec{\bm{E}},\vec{\bm{L}}^{s},\vec{\bm{L}}^{d},\vec{\bm{N}}^{d},\vec{\bm{N}}^{s},\vec{\bm{F}},\vec{\bm{D}}) = \sum_{t \in \mathcal{T}_{k\rightarrow p}} \sum_{r \in \mathcal{R}} F_{t,r} - \alpha \sum_{t \in \mathcal{T}_{k\rightarrow p}} \sum_{i \in \mathcal{R}} \sum_{j \in \mathcal{R}} E_{i,j}^{t} - \beta \sum_{t \in \mathcal{T}_{k\rightarrow p}} \sum_{r \in \mathcal{R}}  D_{t,r} \nonumber
\end{align}
$\sum_{t \in \mathcal{T}_{k \rightarrow p}} \sum_{r \in \mathcal{R}} D_{t,r}$ represents the spatiotemporal imbalance between demand and supply, and is calculated as $D_{t,r} = \left\lvert N_{t,r}^{s} - N_{t,r}^{d} - \frac{1}{\lvert \mathcal{R} \rvert} \sum_{i\in\mathcal{R}} (N_{t,i}^{s} - N_{t,i}^{d}) \right\rvert \ , \forall t \in \mathcal{T}_{k\rightarrow p}, r \in \mathcal{R}$. It is adopted in the objective function of strategic-layer model to further emphasize the importance of mitigating the future supply-demand imbalance in determining the strategic guidance for vehicle flow transfer pattern. The values for $\alpha$ and $\beta$ can be adjusted to reflect different preferences. In the numerical experiments presented in Appendix \ref{sensi_toy_model}, we conduct extensive sensitivity analyses concerning the values of $\alpha$ and $\beta$. 

Let $\tilde{M}$ denote a large constant number. We introduce auxiliary binary variables to linearize the absolute terms $D_{t,r} = \left\lvert N_{t,r}^{s} - N_{t,r}^{d} - \frac{1}{\lvert \mathcal{R} \rvert} \sum_{i\in\mathcal{R}} (N_{t,i}^{s} - N_{t,i}^{d}) \right\rvert \ , \forall t \in \mathcal{T}_{k\rightarrow p}, r \in \mathcal{R}$ in the objective function, as well as the minimum operators $F_{t,r} = \min(N_{t,r}^{d},N_{t,r}^{s}),\forall t\in \mathcal{T}_{k \rightarrow p}, r\in \mathcal{R}$ in constraint \eqref{variable_eq7}. The corresponding linearized constraints are shown as follows. 
\begin{align}
& F_{t,r} \le N_{t,r}^{d} & \forall t \in \mathcal{T}_{k\rightarrow p}, r \in \mathcal{R} \label{aux_eq2} \\
& F_{t,r} \le N_{t,r}^{s} & \forall t \in \mathcal{T}_{k\rightarrow p}, r \in \mathcal{R} \label{aux_eq3} \\
& F_{t,r} \ge N_{t,r}^{d} - \tilde{M} (1-A_{t,r}^{d}) & \forall t \in \mathcal{T}_{k\rightarrow p}, r \in \mathcal{R} \label{aux_eq4} \\
& F_{t,r} \ge N_{t,r}^{s} - \tilde{M}(1-A_{t,r}^{s}) & \forall t \in \mathcal{T}_{k\rightarrow p}, r \in \mathcal{R} \label{aux_eq5} \\
& A_{t,r}^{d} + A_{t,r}^{s} = 1 & \forall t \in \mathcal{T}_{k\rightarrow p}, r \in \mathcal{R} \label{aux_eq6} \\
& D_{t,r} \ge N_{t,r}^{s} - N_{t,r}^{d} - \frac{1}{\lvert \mathcal{R} \rvert}\sum_{i\in\mathcal{R}} (N_{t,i}^{s} - N_{t,i}^{d}) & \forall t \in \mathcal{T}_{k\rightarrow p}, r \in \mathcal{R} \label{aux_eq7} \\
& D_{t,r} \ge -N_{t,r}^{s} + N_{t,r}^{d} + \frac{1}{\lvert \mathcal{R} \rvert} \sum_{i\in \mathcal{R}} (N_{t,i}^{s} - N_{t,i}^{d}) & \forall t \in \mathcal{T}_{k\rightarrow p}, r \in \mathcal{R} \label{aux_eq8} \\
& A_{t,r}^{d}, A_{t,r}^{s}\in \{0,1\} & \forall t \in \mathcal{T}_{k\rightarrow p}, r \in \mathcal{R} \label{aux_eq12}
\end{align}

In consequence, the strategic-layer model (SLM) can be formulated as follows. 
\begin{align}
\textbf{SLM} & \nonumber \\
&\max_{\vec{\bm{M}},
\vec{\bm{E}},\vec{\bm{L}}^{s},\vec{\bm{L}}^{d},\vec{\bm{N}}^{d},\vec{\bm{N}}^{s},\vec{\bm{F}},\vec{\bm{D}},\vec{\bm{A}}^{d}, \vec{\bm{A}}^{s}} \quad \sum_{t \in \mathcal{T}_{k \rightarrow p}} \sum_{r \in \mathcal{R}} F_{t,r} - \alpha \sum_{t \in \mathcal{T}_{k \rightarrow p}} \sum_{i \in \mathcal{R}} \sum_{j \in \mathcal{R}} E_{i,j}^{t} - \beta \sum_{t \in \mathcal{T}_{k \rightarrow p}} \sum_{r \in \mathcal{R}}  D_{t,r} \nonumber \\
&\mbox{s.t.}\quad  \quad \quad \mathrm{Constraints} \eqref{supply_eq5} \sim \eqref{variable_eq6}, \eqref{variable_eq8} \sim  \eqref{aux_eq12} \nonumber
\end{align} 

\subsection{Solution Approach}
SLM is a MILP problem and can be solved by commercial solvers (e.g., CPLEX). Considering that practical scenarios typically involve a substantial number of constraints and MMA necessitates an online implementation, it is essential to devise a customized and efficient solution algorithm. Recent advances in solution methods, including Benders decomposition \citep{lamorgese2016optimal, yin2023integrated}, column generation \citep{jiang2023column,wang2023integrated} and the Lagrangian relaxation method  \citep{ozsen2009facility,xie2016reliable}, have gained widespread popularity in addressing large-scale mixed integer programming problems. In this study, we thus develop a tailored and computationally efficient solution algorithm founded on the Lagrangian relaxation method. We relax constraints \eqref{aux_eq4} and \eqref{aux_eq5}, depicting sufficient matching, with Lagrangian multipliers $\vec{\bm{\lambda}}^{d} = \{\lambda_{t,r}^{d} \mid t \in \mathcal{T}_{k \rightarrow p}, r \in \mathcal{R}\}$ and $\vec{\bm{\lambda}}^{s} = \{\lambda_{t,r}^{s} \mid t \in \mathcal{T}_{k \rightarrow p}, r \in \mathcal{R}\}$, respectively. The relaxed problem is formulated as follows. 
\begin{align}
V(\vec{\bm{\lambda}}^{d},\vec{\bm{\lambda}}^{s}) & = \max_{\vec{\bm{M}},
\vec{\bm{E}},\vec{\bm{L}}^{s},\vec{\bm{L}}^{d},\vec{\bm{N}}^{d},\vec{\bm{N}}^{s},\vec{\bm{F}},\vec{\bm{D}},\vec{\bm{A}}^{d}, \vec{\bm{A}}^{s}}\quad  \sum_{t \in \mathcal{T}_{k \rightarrow p}} \sum_{r \in \mathcal{R}} (1+\lambda_{t,r}^{d} + \lambda_{t,r}^{s})F_{t,r} - \alpha \sum_{t \in \mathcal{T}_{k \rightarrow p}} \sum_{i \in \mathcal{R}} \sum_{j \in \mathcal{R}} E_{i,j}^{t} \nonumber \\
& - \beta \sum_{t \in \mathcal{T}_{k \rightarrow p}} \sum_{r \in \mathcal{R}}  D_{t,r} - \sum_{t \in \mathcal{T}_{k \rightarrow p}} \sum_{r \in \mathcal{R}}\lambda_{t,r}^{d}N_{t,r}^{d} - \sum_{t \in \mathcal{T}_{k \rightarrow p}} \sum_{r \in \mathcal{R}}\lambda_{t,r}^{s}N_{t,r}^{s} \nonumber \\
& + \sum_{t \in \mathcal{T}_{k \rightarrow p}} \sum_{r \in \mathcal{R}}\lambda_{t,r}^{d}\tilde{M}(1-A_{t,r}^{d}) + \sum_{t \in \mathcal{T}_{k \rightarrow p}} \sum_{r \in \mathcal{R}}\lambda_{t,r}^{s}\tilde{M}(1-A_{t,r}^{s}) \nonumber \\
\mbox{s.t.}\quad  &  \mathrm{Constraints} \ \eqref{supply_eq5} \sim \eqref{variable_eq6}, \eqref{variable_eq8} \sim \eqref{aux_eq3}, \eqref{aux_eq6} \sim \eqref{aux_eq12} \nonumber
\end{align}

Obviously, the optimal objective function value of the relaxed problem,  $V(\vec{\bm{\lambda}}^{d},\vec{\bm{\lambda}}^{s})$, is a valid upper bound on the SLM's optimal objective function value. Essentially, the relaxed problem can be equivalently decomposed into two subproblems, shown as follows. 

(1) \textbf{Subproblem 1.}
The first subproblem, shown as follows, is a linear programming, which is polynomial time solvable.
\begin{align}
\max_{\vec{\bm{M}},
\vec{\bm{E}},\vec{\bm{L}}^{s},\vec{\bm{L}}^{d},\vec{\bm{N}}^{d},\vec{\bm{N}}^{s},\vec{\bm{F}},\vec{\bm{D}}} \quad & \sum_{t \in \mathcal{T}_{k \rightarrow p}} \sum_{r \in \mathcal{R}} (1+\lambda_{t,r}^{d} + \lambda_{t,r}^{s})F_{t,r} - \alpha \sum_{t \in \mathcal{T}_{k \rightarrow p}} \sum_{i \in \mathcal{R}} \sum_{j \in \mathcal{R}} E_{i,j}^{t} \nonumber \\
& - \beta \sum_{t \in \mathcal{T}_{k \rightarrow p}} \sum_{r \in \mathcal{R}}  D_{t,r} - \sum_{t \in \mathcal{T}_{k \rightarrow p}} \sum_{r \in \mathcal{R}}\lambda_{t,r}^{d}N_{t,r}^{d} - \sum_{t \in \mathcal{T}_{k \rightarrow p}} \sum_{r \in \mathcal{R}}\lambda_{t,r}^{s}N_{t,r}^{s} \nonumber  \\
\mbox{s.t.}\quad  &  \mathrm{Constraints} \ \eqref{supply_eq5} \sim \eqref{variable_eq6}, \eqref{variable_eq8} \sim \eqref{aux_eq3}, \eqref{aux_eq7}, \eqref{aux_eq8} \nonumber
\end{align}

(2) \textbf{Subproblem 2.} The second subproblem, shown as follows, only contains binary variables $A_{t,r}^{d}$ and $A_{t,r}^{s}$.
\begin{align}
\max_{\vec{\bm{A}}^{d}, \vec{\bm{A}}^{s}} \quad &  \sum_{t \in \mathcal{T}_{k \rightarrow p}} \sum_{r \in \mathcal{R}}\lambda_{t,r}^{d}\tilde{M}(1-A_{t,r}^{d}) + \sum_{t \in \mathcal{T}_{k \rightarrow p}} \sum_{r \in \mathcal{R}}\lambda_{t,r}^{s}\tilde{M}(1-A_{t,r}^{s}) \nonumber \\
\mbox{s.t.}\quad  &  \mathrm{Constraints} \ 
\eqref{aux_eq6}, \eqref{aux_eq12}\nonumber
\end{align}

Subproblem 2 can be optimally solved in polynomial time $O(p\lvert \mathcal{R} \rvert)$ with the following steps: for each interval $t \in \mathcal{T}_{k \rightarrow p}$ and hexagon $r \in \mathcal{R}$, (i) if $\lambda_{t,r}^{d} \ge \lambda_{t,r}^{s}$, then $A_{t,r}^{d} = 0$, $A_{t,r}^{s} = 1$; (ii) otherwise, $A_{t,r}^{d} = 1, A_{t,r}^{s} = 0$.

(3) \textbf{Multiplier Updates.} 
The Lagrangian relaxation algorithm aims at solving the Lagrangian dual problem, $\min_{\vec{\bm{\lambda}}^{d},\vec{\bm{\lambda}}^{s}} V(\vec{\bm{\lambda}}^{d},\vec{\bm{\lambda}}^{s})$. The function $V(\vec{\bm{\lambda}}^{d},\vec{\bm{\lambda}}^{s})$ is convex, as it is a pointwise maximum over affine functions in $\vec{\bm{\lambda}}^{d},\vec{\bm{\lambda}}^{s}$. We thus adopt the subgradient method to solve the Lagrangian dual problem. The initial values of multipliers $(\vec{\bm{\lambda}}^{s})^{0}$ and $(\vec{\bm{\lambda}}^{d})^{0}$ are set to be zero. At the $i^{th}$ iteration of the subgradient optimization, we denote the optimal solution of the relaxed SLM problem by $\vec{\bm{\overline{M}}},\vec{\bm{\overline{E}}},\vec{\bm{\overline{L}}^{s}},\vec{\bm{\overline{L}}^{d}},\vec{\bm{\overline{N}}^{d}},\vec{\bm{\overline{N}}^{s}}$, $\vec{\bm{\overline{F}}},\vec{\bm{\overline{D}}},\vec{\bm{\overline{A}}^{d}}, \vec{\bm{\overline{A}}^{s}}$. Then, the value of multipliers $\vec{\bm{\lambda}}^{d},\vec{\bm{\lambda}}^{s}$ at the $(i+1)^{th}$ iteration are updated as follows.
\begin{align}
(\lambda_{t,r}^{d})^{i+1} = \max \left \{ 0, (\lambda_{t,r}^{d})^{i} +\alpha_{i}(\overline{N}_{t,r}^{d} - \tilde{M}(1-\overline{A}_{t,r}^{d}) - \overline{F}_{t,r}) \right \}, \ & \forall t \in \mathcal{T}_{k \rightarrow p}, r \in \mathcal{R} \nonumber \\
(\lambda_{t,r}^{s})^{i+1} = \max \left \{ 0, \left (\lambda_{t,r}^{s})^{i} +\alpha_{i}(\overline{N}_{t,r}^{s} - \tilde{M}(1-\overline{A}_{t,r}^{s}) - \overline{F}_{t,r} \right ) \right \}, \ & \forall t \in \mathcal{T}_{k \rightarrow p}, r \in \mathcal{R} \nonumber 
\end{align}

where the step size $\alpha_{i}$ is updated as follows. 
\begin{align}
    \alpha_{i} = \xi_{i}(V_{UB}^{i} - V_{LB}^{i}) \cdot \left ( \sum_{t \in \mathcal{T}_{k \rightarrow p}} \sum_{r \in \mathcal{R}} \left (\overline{N}_{t,r}^{d} - \tilde{M}(1-\overline{A}_{t,r}^{d}) - \overline{F}_{t,r} \right )^{2} + \sum_{t \in \mathcal{T}_{k \rightarrow p}} \sum_{r \in \mathcal{R}} \Big(\overline{N}_{t,r}^{s} - \tilde{M}(1-\overline{A}_{t,r}^{s}) - \overline{F}_{t,r} \Big )^{2} \right )^{-1} \nonumber
\end{align}

Here, $\xi_{i}$ is a step size control parameter; $V_{UB}^{i}$ is the upper bound obtained at the $i^{th}$ iteration, the value of which is given by the optimal objective value of $V((\vec{\bm{\lambda}}^{s})^{i},(\vec{\bm{\lambda}}^{d})^{i})$; and $V_{LB}^{i}$ denotes the lower bound and its value is given by a SLM's feasible solution's objective function value. The initial value of $\xi_{i}$ is set to be 2.0 and decreases by a constant discount factor of 0.8 whenever the best upper bound does not decrease within ten iterations. The subgradient algorithm is terminated when the Lagrangian duality gap is less than a threshold or the maximum number of iterations has been reached.

(4) \textbf{Primal Heuristic.}
We devise a primal heuristic to generate a feasible solution to SLM and obtain a lower bound. The computational complexity of SLM is attributed to binary variables 
$\vec{\bm{A}^{d}}$ and $\vec{\bm{A}^{s}}$. An arbitrarily-given values of them may lead to an infeasible solution. Thus, the primal heuristic contains two main steps, (1) finding a set of $\vec{\bm{A}^{d}}$ and $\vec{\bm{A}^{s}}$ which can yield  feasible solutions,  and (2) obtaining the optimal matching and relocation strategy for given $\vec{\bm{A}^{d}}$ and $\vec{\bm{A}^{s}}$. The primal heuristic based on the optimal solution to the relaxed SLM, $\vec{\bm{\overline{M}}}, \vec{\bm{
\overline{E}}}$ and $\vec{\bm{\overline{L}}}$, is described in Algorithm \ref{primal_lower_bound}. 
As the development of feasible matching strategy $\vec{\bm{\underline{M}}}$ involves randomness, we can utilize primal heuristic to generate multiple groups of feasible solutions. Then, the highest lower bound will be selected as $V_{LB}^{i}$ in iteration $i$.

\begin{algorithm}
    \footnotesize
    \caption{The primal heuristic to obtain the lower bound of SLM} 
    \label{primal_lower_bound} 
    \begin{algorithmic}[1] 
    \STATE Initialization: $t = k$
    \WHILE{$t \in \mathcal{T}_{k \rightarrow{p}}$}
    \STATE For each hexagon $r \in \mathcal{R}$, calculate the number of waiting demands $N_{t,r}^{d}$ subject to equation \eqref{demand_eq1} -- \eqref{demand_eq2}, the number of vacant vehicles $N_{t,r}^{s}$ on the basis of equation \eqref{supply_eq5} -- \eqref{supply_eq2} and the number of waiting demands towards each hexagon from hexagon $r$ in strategic interval $t$ based on the right-hand side (RHS) of equation \eqref{variable_eq2} -- \eqref{variable_eq3}
    \IF{$N_{t,r}^{d} \le N_{t,r}^{s}$}
    \STATE All demands in hexagon $r$, interval $t$ can be met and the auxiliary variables are $\underline{A}_{t,r}^{d}=1, \underline{A}_{t,r}^{s} = 0$
    \IF{$t == k$}
    \STATE The new matching strategy for interval $t$ is $M_{r,j}^{k} = l_{(k-1),r,j}^{d} + \hat{N}_{k,r}^{d}b_{r \rightarrow j}^{k}$
    \ELSE 
    \STATE The new matching strategy for interval $t$ is $M_{r,j}^{t} = L_{(t-1),r,j}^{d}(1-\mu_{t-1}^{d}) + \hat{N}_{t,r}^{d}b_{r\rightarrow j}^{t}$
    \ENDIF 
    \STATE Calculate the number of  remaining vacant vehicles $L_{t,r}^{s} = N_{t,r}^{s} - \sum_{i \in \mathcal{R}}M_{r,j}^{t}$
    \STATE The new relocation strategy is $E_{r,j}^{t} = L_{t,r}^{s}\frac{\overline{E}_{r,j}^{t}}{\overline{L}_{t,r}^{s}}, \forall j \in \mathcal{R}$ 
    \ELSE 
    \STATE All vehicles will be matched and the auxiliary variables are $\underline{A}_{t,r}^{d}=0, \underline{A}_{t,r}^{s} = 1$
    \STATE Initially allocate vehicles to demands towards each hexagon from $r$ in proportion with $\overline{M}_{r,j}^{t}$
    \IF{$t == k$}
    \STATE $\dot{M}_{r,j}^{k} = \min \left\{ l_{(k-1),r,j}^{d} + \hat{N}_{k,r}^{d}b_{r\rightarrow j}^{k},\frac{\overline{M}_{r,j}^{k}}{\sum_{i \in \mathcal{R}}\overline{M}_{r,i}^{k}}N_{k,r}^{s}\right\}$
    \ELSE
    \STATE $\dot{M}_{r,j}^{t} = \min \left\{ L_{(t-1),r,j}^{d}(1-\mu_{t-1}^{d}) + \hat{N}_{t,r}^{d}b_{r\rightarrow j}^{t},\frac{\overline{M}_{r,j}^{t}}{\sum_{i \in \mathcal{R}}\overline{M}_{r,i}^{t}}N_{t,r}^{s}\right\}$
    \ENDIF
    \STATE $N_{t,r}^{s} - \sum_{j \in \mathcal{R}}\dot{M}_{r,j}^{t}$ remaining vacant vehicles are randomly assigned to demands in hexagon $r$ and obtain new matching strategy $M_{r,j}^{t}$ with restriction \eqref{variable_eq2} and \eqref{variable_eq3}
    \STATE The new relocation strategy is $E_{r,j}^{t} = 0, \forall j \in \mathcal{R}$
    \ENDIF
    \STATE Update the number of demand and vehicles remaining to the next interval, denoted as $L_{t,r,j}^{d}$, $L_{t,r}^{s}$ by the equation \eqref{variable_eq4}, \eqref{variable_eq5} and \eqref{supply_eq7}
    \STATE $t = t + 1$
    \ENDWHILE
    \STATE Given $\vec{\bm{\underline{A}}^{d}} =\{\underline{A}_{t,r}^{d} \mid t \in \mathcal{T}_{k\rightarrow p}, r \in \mathcal{R}\}, \vec{\bm{\underline{A}}^{s}} = \{\underline{A}_{t,r}^{s} \mid t \in \mathcal{T}_{k\rightarrow p}, r \in \mathcal{R} \}$, the conditional optimal strategies $\vec{\bm{\underline{M}}} = \{ \underline{M}_{i,j}^{t} \mid t \in \mathcal{T}_{k \rightarrow p}, i,j \in \mathcal{R}\}, \vec{\bm{\underline{E}}} 
 = \{\underline{E}_{i,j}^{t} \mid t \in \mathcal{T}_{k \rightarrow p}, i,j \in \mathcal{R} \}$ are solved by the linear programming which has the same objective function with SLM and the constraints $\eqref{supply_eq5} \sim \eqref{variable_eq6}, \eqref{variable_eq8} \sim \eqref{variable_eq9}, \eqref{aux_eq2} \sim \eqref{aux_eq5}, \eqref{aux_eq7}, \eqref{aux_eq8}$
    \end{algorithmic}
\end{algorithm}

\subsection{Theoretical Performance Evaluation of the Lagrangian Relaxation Method}
It is extremely difficult (if not impossible) to theoretically characterize the effectiveness of the LR method on solving SLM. To shed light on its effectiveness, we theoretically investigate its optimality gap in a stylized network. There are $\bar{n}$ nodes, and the vehicle travel time between each pair of nodes equals the length of one relocation interval in the network. For analytical tractability, the following simplifications are further set.
\begin{itemize}
    \item We set $\alpha = \beta = 0$ in the SLM's objective function. That is,  SLM aims at maximizing the total number of completed trips.
    \item The dropping rate of demands is one, suggesting that all customers will only wait for one strategic interval.
    \item The vehicle fleet is deterministic throughout the analysis intervals.
\end{itemize}

We hereinafter refer to the case with the above simplifications as the \textit{stylized case}. The following proposition asserts the strong performance of LR approach to solve SLM in the stylized case, the proof of which is provided in Appendix \ref{proof_pro}.

\begin{proposition}
\label{LRoptimality_prop}
   In the stylized case with a $\bar{n}$-node network, the Lagrangian dual of SLM owns the same optimal objective function value as that of the original SLM problem. 
\end{proposition} 

Proposition \ref{LRoptimality_prop} provides a partial explanation for the effectiveness of the LR approach in solving SLM problems, as the stylized network used in the proposition captures some essential properties of many real-world urban road networks. For instance, a stylized case with a three-node network can be treated as a simplified representation of the road network of some cities with explicit functional zoning, i.e., one residential district, one commercial district, and one industrial district. Nevertheless, the networks in realistic cases are more complex, with differentiated trip distances and times, and objective functions that consider multiple terms. Therefore, achieving a zero duality gap in larger-scale problems is challenging. However, with Proposition \ref{LRoptimality_prop}, we can envision that the LR approach is still capable of achieving good performance solutions in more generalized cases, and this will be validated numerically in Section \ref{numerical}.

\section{The Execution Layer}
\label{execution}

After the SLM model determines the strategic guidance for interval $k$, the execution-layer model proceeds with specific order matching and relocation based on the suggested strategies in each hexagon independently. As previously noted, a strategic interval typically spans a duration of five to 15 minutes and comprises multiple matching intervals, with each matching interval lasting between two to 20 seconds. During each matching interval, both the vacant vehicles and waiting passengers are gathered by the platform, which then utilizes a batch matching mechanism to assign specific vehicles to the requests based on the strategic guidance from the strategic layer and the real-time information. With a well-designed matching mechanism, the cumulative number of matched demands traveling from hexagon $r$ to $j$ during strategic interval $k$ can closely approximate the suggested target of $M_{r,j}^{k}$. At the end of a strategic interval, vacant vehicles are relocated. To ensure consistency between the actual relocation results and the suggested relocation strategy $E_{r,j}^{k}$, it is crucial to design an appropriate mechanism. In the subsequent subsections, we develop a two-step matching mechanism and a real-time relocation mechanism that address these requirements.



\subsection{Two-step Mechanism for Order Matching}

Given the strategic matching guidance, $M_{r,j}^{k}$, in strategic interval $k$, we design the following two-step mechanism to determine the vehicle-passenger matching in hexagon $r$ in each matching interval .

\subsubsection{The First Step.}

\paragraph{}
The first step in the matching mechanism  determines the actual number of vehicles to be assigned to passengers heading for each hexagon in the current matching interval. These quantities must be proportional to the gaps between the strategic-layer guidance and the number of completed trips by the current matching interval in strategic interval $k$.  In hexagon $r$,  we denote $u_{r,j}^{k}$ as the number of successfully matched demands from hexagon $r$ to $j$ immediately before the current matching interval. $d_{r,j}^{k}$ represents the uncompleted target number of demands from hexagon $r$ to $j$, and we have $M_{r,j}^{k} = u_{r,j}^{k} + d_{r,j}^{k}$. The number of vacant vehicles, $n_{r}^{s}$, and the number of waiting demands towards hexagon $j$ in hexagon $r$, $n_{r,j}^{d}$, are both known parameters at the end of a matching interval. Consider the following three scenarios in real operations.

Case 1: The guidance from the strategic layer has not been completed, and the available vehicles dominate the waiting demands, i.e., $n_{r}^{s} \ge \sum_{j \in \mathcal{R}}n_{r,j}^{d}$. In this case, each waiting passenger can be matched with a vacant vehicle, and no passenger needs to be selected for pick-up.

Case 2: The target has not been completed, and the number of vacant vehicles is less than the number of waiting demands, i.e., $n_{r}^{s} < \sum_{j \in \mathcal{R}}n_{r,j}^{d}$. In this case, the platform must select a subset of passengers to serve in this matching interval. To ensure that the cumulative number of matched demands in strategic interval $k$ approaches the strategic guidance from SLM, the number of vehicles to be matched with the demands from $r$ to $j$ in the current matching interval, denoted by $x_{r,j}$, is proportional to $d_{r,j}^{k}$. To achieve the above goal and ensure that $x_{r,j}$ is an integer simultaneously, the matching vehicle allocation (MVA) problem is formulated as follows.
\begin{align}
\textbf{MVA} \nonumber \\
\min_{\bm{\vec{x}},\bm{\vec{c}}} \quad & \sum_{j \in \mathcal{R}} c_{r,j}  \nonumber\\
\mbox{s.t.} \quad & c_{r,j} + x_{r,j} \ge n_{r}^{s} \times \frac{d_{r,j}^{k}}{\sum_{i \in \mathcal{R}}d_{r,i}^{k}} & \forall j \in \mathcal{R} \label{eq29} \\
 & x_{r,j} \le n_{r,j}^{d} & \forall j \in \mathcal{R} \label{eq30}\\
 & \sum_{j \in \mathcal{R}} x_{r,j} = n_{r}^{s} \label{eq31} \\
 & c_{r,j} \ge 0 & \forall j \in \mathcal{R} \label{eq32}\\
 & x_{r,j} \in \mathbb{N} & \forall j \in \mathcal{R} \label{eq33}
\end{align}

Constraints \eqref{eq29} and \eqref{eq32}, together with the objective function, dictate that $c_{r,j}^{*} = \mathrm{max}\left(0, n_{r}^{s} \times \frac{d_{r,j}^{k}}{\sum_{i \in \mathcal{R}}d_{r,i}^{k}}-x_{r,j}^{*} \right)$, that is, $c_{r,j}^{*}$ represents the gap between $x_{r,j}^{*}$ and the number of vehicles assigned to the demands from hexagon $r$ to $j$ calculated based on the ratio of uncompleted target number of demands. Constraint \eqref{eq30} ensures that $x_{r,j}^{*}$ is no more than the actual number of waiting demands. Constraint \eqref{eq31} shows that all vacant vehicles should be matched to orders to maximize the fleet utilization. MVA is a mixed integer linear programming model and the number of variables increases linearly with the number of hexagons. We propose Algorithm \ref{middle_algo} to solve MVA with a polynomial-time complexity, as demonstrated by Proposition \ref{optimal_MDM}. 

\begin{algorithm}
    \footnotesize
    \caption{The algorithm to solve MVA} 
    \label{middle_algo} 
    \begin{algorithmic}[1] 
    \REQUIRE ~~\\ 
    The number of vacant vehicles in hexagon $r$, $n_{r}^{s}$\\
    The number of waiting demands towards hexagon $j$ in hexagon $r$, $n_{r,j}^{d}$\\
    The number of target demands to be completed from hexagon $r$ to $j$ in strategic interval $k$, $d_{r,j}^{k}$\\
    \ENSURE The number of vehicles to be matched with the demands from $r$ to $j$ in the current matching interval $x_{r,j}$\\ 
    \STATE Initialization: $x_{r,j}$ = min ($\left \lfloor  \frac{d_{r,j}^{k}}{\sum_{i \in \mathcal{R}}d_{r,i}^{k}}  \times n_{r}^{s} \right\rfloor$, $n_{r,j}^{d}$), $\forall j \in \mathcal{R}$
    \STATE Denote $\mathcal{S} = \{j \ | \  n_{r,j}^{d} > \left \lfloor  \frac{d_{r,j}^{k}}{\sum_{i \in \mathcal{R}}d_{r,i}^{k}}  \times n_{r}^{s} \right\rfloor, j \in \mathcal{R} \}$
    \STATE Calculate $\theta_{j} = \frac{d_{r,j}^{k}}{\sum_{i \in \mathcal{R}}d_{r,i}^{k}}  \times n_{r}^{s} - \left \lfloor  \frac{d_{r,j}^{k}}{\sum_{i \in \mathcal{R}}d_{r,i}^{k}}  \times n_{r}^{s} \right\rfloor\ $, $j \in \mathcal{S} $; $m = n_{r}^{s} - \sum_{j \in \mathcal{R}} x_{r,j} $
    \IF{$| \mathcal{S} | > m $}
    \STATE Rearrange the values of $\theta_{j}$ in descending order; identify the hexagons with top $m$  largest $\theta_{j}$ and update the corresponding $x_{r,j} = x_{r,j} + 1$
    \ELSE
    \STATE For all $j \in \mathcal{S}$, $x_{r,j} = x_{r,j} + 1 $ and update $m = m - | S |$
    \WHILE{ $m > 0$}
    \STATE Calculate $\delta_{j}$ = max($d_{r,j}^{k} - x_{r,j} $,0)\textbf{I}($n_{r,j}^{d} > x_{r,j}$) $\forall j \in \mathcal{R}$
    \STATE Let $\tilde{\mathcal{S}}$ represent the set of hexagons with with unmatched demands and uncompleted targets, defined as  $\tilde{\mathcal{S}} = \{ j \ | \  \delta_{j} > 0, j \in \mathcal{R} \}$
    \IF{ $m > |\tilde{\mathcal{S}}| > 0$}
    \STATE For all $j \in \tilde{\mathcal{S}}$, $x_{r,j} = x_{r,j} + 1$
    \STATE Update $m = m - |\tilde{\mathcal{S}}|$
    \ELSIF{$m > |\tilde{\mathcal{S}}| = 0$}
    \STATE Denote $\hat{\mathcal{S}} = \{j \ | \ n_{r,j}^{d} - x_{r,j} > 0\}$ as the set of hexagons with unmatched demands
    \STATE Randomly sample one hexagon $i \in \hat{\mathcal{S}}$ and the corresponding $x_{r,j} = x_{r,j} + 1$
    \STATE Update $m = m - 1$
    \ELSE
    \STATE Arrange $\delta_j$ in descending order, and for hexagons with top $m$ largest $\delta_{j}$, update $x_{r,j} = x_{r,j} + 1$
    \STATE Update $m = 0$
    \ENDIF
    \ENDWHILE
    \ENDIF
    \RETURN $x_{r,j}$
    \end{algorithmic}
\end{algorithm}

\begin{proposition}
\label{optimal_MDM}
    The optimal solution of MVA can be obtained by Algorithm \ref{middle_algo} with a polynomial time complexity.
\end{proposition}

Case 3: When the target number of matched demands from SLM is achieved, the strategic guidance ceases to impose constraints on the matching process within the current matching interval. Consequently, we proceed directly to the second step, which involves fully matching waiting passengers with vacant vehicles.

\subsubsection{The Second Step.}
\paragraph{}

Given the matched vehicle number from the MVA model in the first step, this step assigns specific vehicles to customers via batch matching. In a matching interval, the set of waiting passengers in hexagon $r$ is denoted by $\mathcal{C}_{r} = \{c_{p}^{r}\}$, where $c_{p}^{r}$ represents the $p^{th}$ passenger with destination $\theta_{p}^{r} \in \mathcal{R}$ and arrival sequence $\nu_{p}^{r} \in \mathbb{N}^{+}$. The set of vacant vehicles is denoted by $\mathcal{V}_{r} =\{v_{q}^{r} \} $, where $v_{q}^{r}$ represents the $q^{th}$ vacant vehicle. $w_{q,p}^{r}$ represents the pick-up distance when passenger $c_{p}^{r}$ is assigned to vehicle $v_{q}^{r}$. Let binary variable $y_{q,p}^{r}$ represents whether vehicle $ v_{q}^{r}$ is assigned to passenger $c_{p}^{r}$ or not. Recall that the MVA model in case 2 of the first step imposes a constraint on the assignment of $x_{r,j}$ vehicles to be matched with the demands from $r$ to $j$ in the current matching interval. Then, the vehicle and order matching (VOM) problem is formulated to effectively satisfy this constraint as follows.
\begin{align}
\textbf{VOM} \nonumber \\
\min_{\bm{\vec{y}}} \quad & \sum_{c_{p}^{r} \in \mathcal{C}_{r}} \sum_{v_{q}^{r} \in \mathcal{V}_{r}} y_{q,p}^{r} \times w_{q,p}^{r} \times \mathnormal{G}(\nu_{p}^{r}) \nonumber\\
\mathrm{s.t.} \quad & \sum_{v_{q}^{r} \in \mathcal{V}_{r}} y_{q,p}^{r} \le 1  &  \forall c_{p}^{r} \in \mathcal{C}_{r} \label{eq34} \\
& \sum_{c_{p}^{r} \in \mathcal{C}_{r}} y_{q,p}^{r} = 1 & \forall v_{q}^{r} \in \mathcal{V}_{r} \label{eq35}\\
& \sum_{c_{p}^{r} \in \mathcal{C}_r} \sum_{v_{q}^{r} \in \mathcal{V}_{r}} \mathbf{I}(\theta_p^r = j) y_{q,p}^{r}=x_{r,j} & \forall j \in \mathcal{R} \label{eq36} \\
& y_{q,p}^{r} \in \{ 0,1\} & \forall v_{q}^{r} \in \mathcal{V}_{r}, c_{p}^{r} \in \mathcal{C}_{r} \label{eq37}
\end{align}

The objective function aims at minimizing the total weighted pick-up distance, where $G(\cdot)$ is a monotonically increasing function and suggests that those passengers with earlier arrivals are given higher priority. Constraint \eqref{eq34} indicates that at most one vehicle is assigned to each demand, and constraint \eqref{eq35} ensures that only one request can be picked by each vacant vehicle. Constraint \eqref{eq36} restricts that $x_{rj}$ demands towards hexagon $j$ will be matched in the current matching interval, where $\mathbf{I}(\theta_p^r = j)$ is the indicator function that takes a value of one if the destination of passenger $c_{p}^{r}$ is hexagon $j$. Constraint \eqref{eq37} requires $y_{q,p}^{r}$ to be binary. The typical batch matching entails solving a bipartite graph matching problem where vehicles and passengers are two sets of nodes, and links only connect nodes in separate sets. As a variation of batch matching, VOM introduces supplementary constraints provided by the first step. To facilitate a better understanding of VOM's properties, we further construct an extended graph $\mathcal{G}(\mathcal{V}_{r},\mathcal{C}_{r},\mathcal{Z}_{r},E)$ for illustrative purposes. Specifically, for each hexagon $r$, we introduce the virtual nodes  $z_1^{r},z_2^{r},\ldots,z_{\mid \mathcal{R} \mid}^{r}$ corresponding to each hexagon in set $\mathcal{R}$; for each customer $c_p^{r}$ with destination $\theta_{p}^{r}$, we add the edge $(c_p^r,z_{\theta_{p}^{r}}^{r})$ with the cost of zero and the capacity of one. Then, the following proposition states that the linear-programming relaxation of VOM is equivalent to the minimum cost flow problem in the extended graph $\mathcal{G}(\mathcal{V}_{r},\mathcal{C}_{r},\mathcal{Z}_{r},E)$, and the optimal solution of VOM can be solved by the corresponding linear-programming relaxation. 

\begin{proposition}
\label{execution_pro}
    The linear-programming relaxation of VOM is equivalent to the minimum cost flow problem in the extended graph $\mathcal{G}(\mathcal{S}_{r},\mathcal{C}_{r},\mathcal{Z}_{r},E)$. The optimal solution of VOM can be obtained by solving its linear-programming relaxation problem. 
\end{proposition}


Recall that for cases 1 and 3, the first step does not require specific number of vehicles assigned to passengers, and thus VOM without constraint \eqref{eq36} can be utilized to decide vehicle and order matching, which can be solved by linear-programming relaxation or the auction algorithm \citep{bertsekas1992forward}. Upon the completion of the current matching interval, the uncompleted target is updated. The order requests and vehicles will accumulate for a new matching interval and the two-step mechanism will be executed repeatedly. To better illustrate the core idea of the two-step mechanism, a toy example with three regions $A,B,C$ is provided in Appendix \ref{toy_illu}. 

It is essential to emphasize that the lower layer's decision-making mechanism is guided by, but not restricted to specific spatial transfer pattern guidance. In the first step, we outline three distinct cases.
The first two cases pertain to the situations where the strategic target remains unattained, necessitating the matching of the waiting passengers and vehicles. If this additional matching overshoots the target provided by the strategic layer, the lower layer prioritizes completing the strategic target, followed by the assignment of the remaining vehicles to fulfill the outstanding demands, ultimately achieving the fully matching. Conversely, if the target is met while waiting demands and vacant vehicles exist, as stated in Case 3, the lower layer fully matches demands and vehicles rather than deferring the use of these resources.

\subsection{Real-time Mechanism for Relocation}
At the end of strategic interval $k$, the vacant vehicles are relocated in accordance with the guidance specified by the strategic layer. For simplicity, the suggested number of relocated vehicles from hexagon $r$ to $j$ in interval $k$, $E_{r,j}^{k}$, is not necessarily an integer in SLM. Let $l_{k,r}^{s}$ denote the number of vacant vehicles in hexagon $r$ at the end of interval $k$. If the vacant vehicles in hexagon $r$ can meet the relocation goal, that is, $l_{k,r}^{s} \ge \sum_{j \in \mathcal{R}}\lceil E_{r,j}^{k} \rceil$, then the vacant vehicles are relocated as per the suggested relocation strategy after rounding up. Otherwise, the number of relocated vehicles from hexagon $r$ to $j$, denoted as $z_{r,j}$, is an integer and roughly proportional to  $E_{r,j}^{k}$, which is determined by solving the following integer programming model. 
\begin{align}
\max_{\vec{\bm{z}}} \quad & \sum_{j \in \mathcal{R}} E_{r,j}^{k} \times \left(z_{r,j} - \left\lfloor  \frac{E_{r,j}^{k}}{\sum_{i \in \mathcal{R}}E_{r,i}^{k}}  \times l_{k,r}^{s} \right\rfloor \right) \nonumber\\
\mathrm{s.t.} \quad & z_{r,j} \ge \left\lfloor  \frac{E_{r,j}^{k}}{\sum_{i \in \mathcal{R}}E_{r,i}^{k}}  \times l_{k,r}^{s} \right\rfloor & \forall j \in \mathcal{R} \label{eq101} \\
& z_{r,j} \le \left\lceil  \frac{E_{r,j}^{k}}{\sum_{i \in \mathcal{R}}E_{r,i}^{k}} \times l_{k,r}^{s} \right\rceil  & \forall j \in \mathcal{R} \label{eq102} \\
& \sum_{j \in \mathcal{R}} z_{r,j} = l_{k,r}^{s} & \label{eq103} \\
& z_{r,j} \in \mathbb{N} &  \forall j \in \mathcal{R} \label{eq104}
\end{align}

Constraints \eqref{eq101} and \eqref{eq102} jointly ensure that $z_{r,j}$ is proportional to $E_{r,j}^{k}$ in rough. Constraint \eqref{eq103} indicates that all vacant vehicles are relocated if the relocation goal cannot be met. The objective function tends to relocate more vehicles to hexagons with higher relocation targets. The integer programming model can be globally solved by the simple greedy algorithm, which is presented in Appendix \ref{greedy_relo}.

\section{Numerical Studies}
\label{numerical}
In this section, we perform extensive numerical experiments to validate the effectiveness of the MMA framework on two networks, a three-node toy network and a realistic network in Chengdu, a major city in Southwest China. We assume that passengers and drivers are homogeneous in the simulation. All computational experiments are implemented on a computer with an Intel(R) Core(TM) i7-4770HQ CPU@2.20 GHz and 16 GB RAM. The algorithms are coded in Python 3.7.

\subsection{A Toy Network}
\label{toy_network}
We design a three-node toy network to demonstrate how the proposed framework improves the performance of the ride-hailing platform. A simulator is established to approximate realistic passenger and driver behavior, and to simulate the matching, pick-up and delivery processes in the real world. The detailed simulation setting is presented in Appendix \ref{toy_model_design}. 

\subsubsection{Benchmarks.} 
\paragraph{}
We compare the MMA framework with the following efficient benchmarks. 

\begin{itemize}
    \item FCFS: it is a greedy and myopic method, and matches the closest vehicle with the first passenger in the waiting queue. It is simple to operate and does not consider the overall efficiency of system. 
    \item Batch matching: batch matching accumulates passengers and vacant vehicles for a matching interval, and then determines specific assignment by minimizing the weighted pick-up distance. Different from classical batch matching, we prioritize matching requests with a longer waiting time in the system.
    
    \item RL-based method: the third benchmark is the learning and planning framework proposed by \citet{Xu2018}. The offline learning stage obtains the state-value function $\tilde{V}(s)$ indicating the expected value of a driver being in the spatiotemporal state $s = (t,r), \forall t \in \mathcal{T}, r \in \mathcal{R} $. The online learning stage takes the value function as the edge weights in the bipartite graph and determines the assignment. 
    
    \item RL-based method with relocation: to fairly compare MMA with the baselines, the third benchmark can incorporate the relocation based on the value function and discrete choice model to make joint order matching and relocation decisions \citep{tang2021value}. Specifically, the relocation is triggered every 10 minutes which is consistent with MMA. The probability for each vacant vehicle in hexagon $i$ relocating to hexagon $j$ in strategic interval $t$, denoted as $\tilde{p}^{t}_{i \rightarrow j}$, is proportional to the discounted state value function, calculated as follows. 
    
    $$\tilde{p}^{t}_{i \rightarrow j} = \frac{e^{{\gamma}^{a[ij]}\tilde{V}(t+a[ij],j)}}{\sum_{r \in \mathcal{R}}e^{{\gamma}^{a[ir]}\tilde{V}(t+a[ir],r)}}$$
    
    where $0<\gamma \le 1$ is the discounted factor and $a[ij]$ represents the number of strategic intervals required to travel from hexagon $i$ to $j$.
\end{itemize}

In the following experiments, we set the batch matching interval to 10 seconds and the planning intervals to encompass nine strategic intervals in the MMA framework.

\subsubsection{Experimental Results.}
\paragraph{}

The simulation is conducted for ten days and the simulation period for each day is 24 hours, which is divided into 144 strategic intervals with identical lengths of ten minutes in MMA. We compare the performance of algorithms concerning two major measures: (1) the number/rate of completed requests, and (2) the average pick-up distance. Table \ref{res_toy_model1} displays the average values of these measures over the course of ten days.

\begin{table}[!ht]
  \footnotesize
  \centering
  \linespread{-0.1} 
  \caption{Performance of MMA and benchmarks in the toy model.}
  \begin{tabularx}{\textwidth}{m{1.6cm}m{2.7cm}m{2.2 cm}m{1.6cm}m{1.6cm}m{3.4cm}m{1.8cm}}
    \hline
   \textbf{Method} & \textbf{Number of completed requests}  &\textbf{Requests completion rate ($\%$)} &\textbf{Improve- ment ($\%$)} &\textbf{Average \newline pick-up \newline distance}  &\textbf{Transition matrix of actually-completed requests} & \textbf{Average of relocation times}\\
    \hline
    FCFS & 8346  & 55.64  & 0.23 & 1.5911 & $$\begin{pmatrix} 0.200 & 0.301 & 0.499 \\ 0.300 & 0.202 & 0.498 \\ 0.200 & 0.198 & 0.602 
    \end{pmatrix}$$ & 0\\
    Batch matching & 8327  & 55.51  & 0.00 & 1.1015 & $$\begin{pmatrix} 0.197 & 0.303 & 0.500 \\ 0.296 & 0.202 & 0.502 \\ 0.201 & 0.199 & 0.600
    \end{pmatrix}$$ & 0\\
    RL-based \newline without \newline relocation & 9529  & 63.53  & 14.43 & 1.2062 & $$\begin{pmatrix} 0.303 & 0.415 & 0.282\\ 0.514 & 0.370 & 0.116 \\ 0.171 & 0.085 & 0.744 \end{pmatrix}$$ & 0 \\
    RL-based \newline with \newline relocation & 10383  & 69.22  & 24.69 & 1.1897 & $$\begin{pmatrix} 0.277 & 0.399 & 0.324 \\ 0.511 & 0.365 & 0.124 \\ 0.203 & 0.106 & 0.691 \end{pmatrix}$$ & 186 \\
    MMA  \newline without \newline relocation & \textbf{10346}  & 68.97  & 24.25 & 1.0210 & $$\begin{pmatrix} 0.291 & 0.399 & 0.310 \\ 0.437 & 0.293 & 0.270 \\ 0.191 & 0.151 & 0.658 \end{pmatrix}$$ & 0 \\
    MMA \newline $\alpha=0.5$\newline$\beta=0$ & 12256  & 81.70  & 47.18 & 1.1218 & $$\begin{pmatrix} 0.233 & 0.341 & 0.426 \\ 0.365 & 0.235 & 0.400 \\ 0.203 & 0.157 & 0.640 \end{pmatrix}$$ & 1391 \\
    MMA \newline $\alpha=0.5$\newline$\beta=0.2$ & 12458 & 83.05  & 49.61 & 1.1079 & $$\begin{pmatrix} 0.228 & 0.323 & 0.449 \\ 0.359 & 0.232 & 0.409 \\ 0.211 & 0.164 & 0.625 \end{pmatrix} $$ & 1597 \\
    \hline
  \end{tabularx}
  \label{res_toy_model1}
\end{table}

The results indicate that the FCFS strategy yields the longest average pick-up distance, because it does not adopt a batch-matching structure. Furthermore, even without relocation, both MMA and the RL-based method still yield substantially higher performance than FCFS and batch matching. To elucidate the reasons, we initiate our analysis by examining the characteristics of the simulation setup. An imbalanced transition probability matrix $P=(p_{i\rightarrow j})$ is designed in the simulation, as shown below, where $p_{i \rightarrow j}$ represents the proportion of generated requests heading for region $j$ from $i$ accounting for the total generated requests in region $i$.

\begin{equation}
    P = 
    \begin{vMatrix}{ccc}
     0.2 & 0.3 & 0.5\\
     0.3 & 0.2 & 0.5 \\
     0.2 & 0.2 & 0.6     
    \end{vMatrix}
\end{equation}

Matrix $P$ implies that 50$\%$ of the demands in regions $A$ and $B$ are directed towards $C$, while 60$\%$ of the demands in $C$ also proceed to $C$. FCFS and batch matching do not incorporate the requests' destination information into the short-sighted decision-making process. In consequence, a large proportion of occupied vehicles will travel from $A$ and $B$ to $C$, but only a small proportion of vehicles will travel to $A$ and $B$, as shown by the transition matrix of actually-completed requests under different methods in Table \ref{res_toy_model1}. This imbalanced transition inevitably leads to an accumulation of vacant vehicles in $C$ and a shortage of supplies in regions $A$ and $B$. In contrast, MMA utilizes the prediction information and develops the matching strategy to optimize long-term efficiency. Based on the nature of demand generation and the future imbalance of demand and supply in each region, MMA suggests more requests towards $A$ and $B$ should be matched, as shown by the transition matrix of actually-completed requests under MMA in Table \ref{res_toy_model1}. The RL-based method learns the value function of spatiotemporal states and applies it as weights in the bipartite graph to maximize the long-term revenue. Based on the value function, the RL-based method is capable of adjusting the matched probability of requests towards different regions. However, the improvement of the RL-based method is less significant than that of MMA. This is not to mention the significant limitations inherent in the RL-based method, including the lack of interpretability and the requirement of highly computation-intensive process. 

\begin{figure}[!htb]
\centering
\subfigure[Batch matching]{
    \begin{minipage}[b]{0.28\linewidth} 
    \includegraphics[width=5cm]{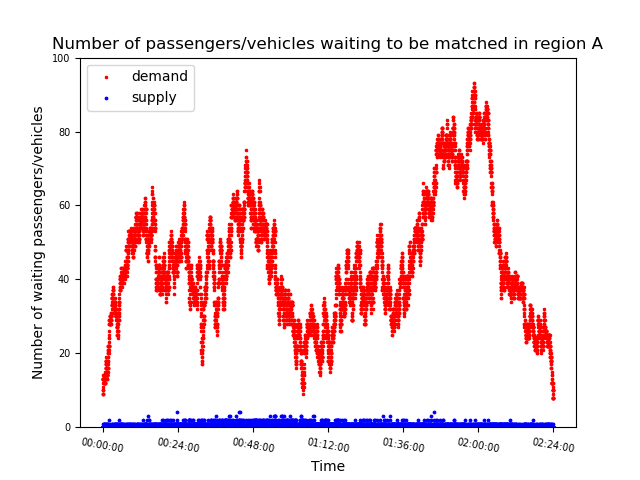}\vspace{1.5pt}
    \includegraphics[width=5cm]{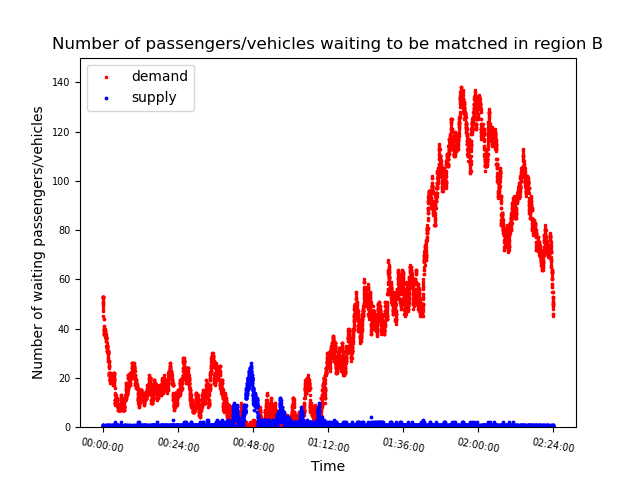}\vspace{1.5pt}
    \includegraphics[width=5cm]{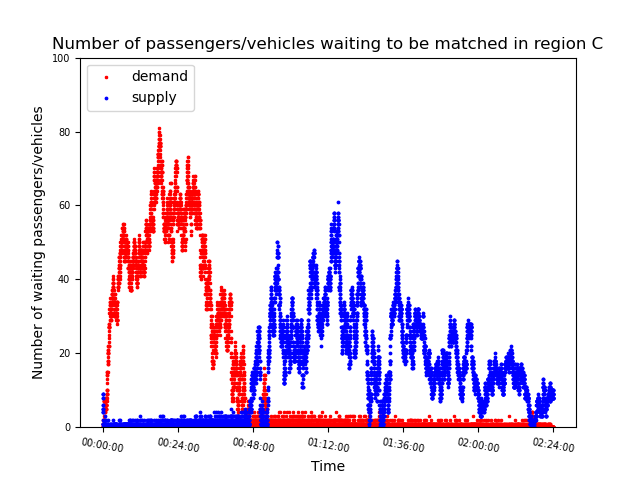}\vspace{1.5pt}
    \includegraphics[width=5cm]{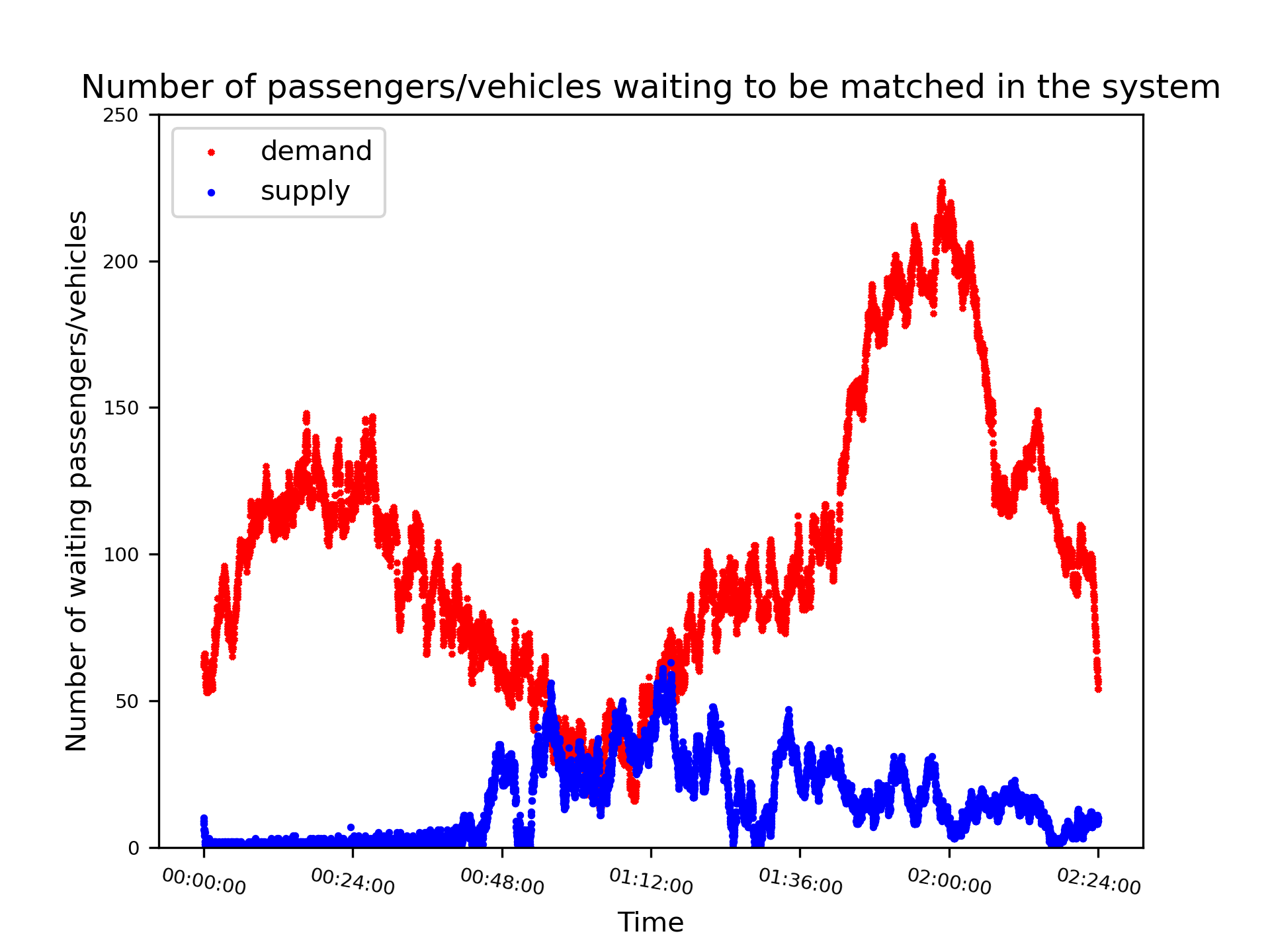}\vspace{1pt}
    \end{minipage}
}
\quad 
\subfigure[MMA without relocation]{
    \begin{minipage}[b]{0.28\linewidth}
    \includegraphics[width=5cm]{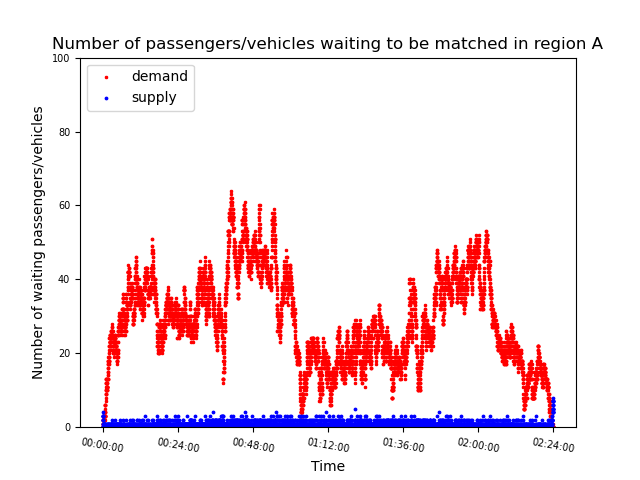}\vspace{1.5pt} 
    \includegraphics[width=5cm]{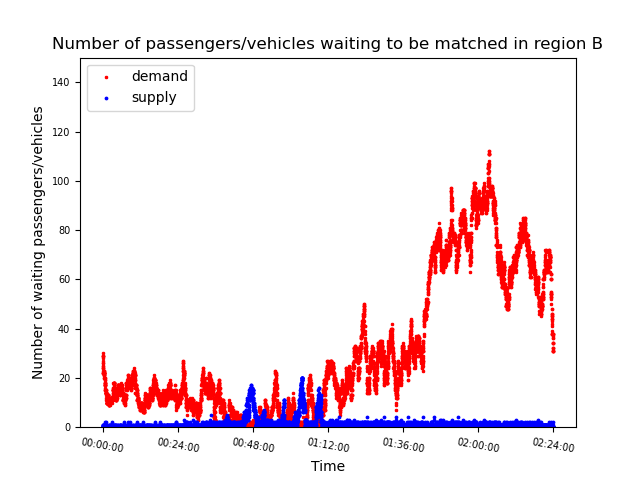}\vspace{1.5pt}
    \includegraphics[width=5cm]{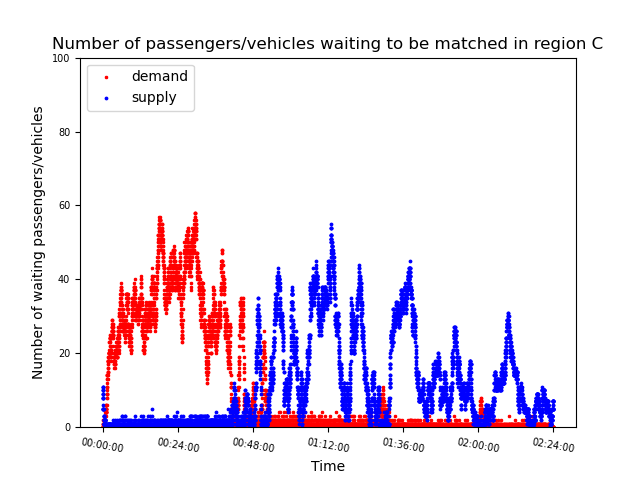}\vspace{1.5pt}
    \includegraphics[width=5cm]{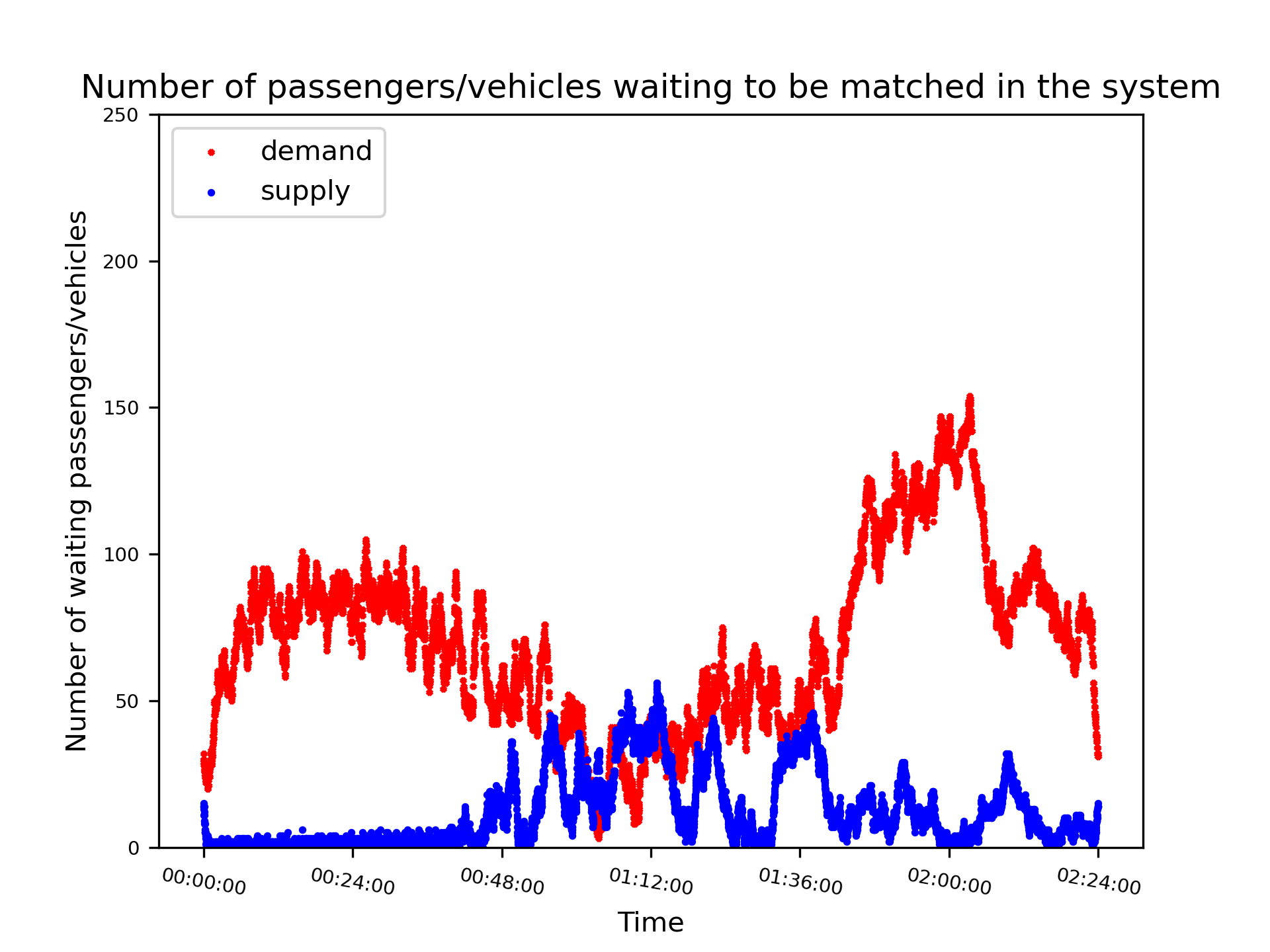}\vspace{1pt}
    \end{minipage}
}
\quad 
\subfigure[MMA with $\alpha=0.5$,\newline$\beta=0.2$]{
    \begin{minipage}[b]{0.28\linewidth}
    \includegraphics[width=5cm]{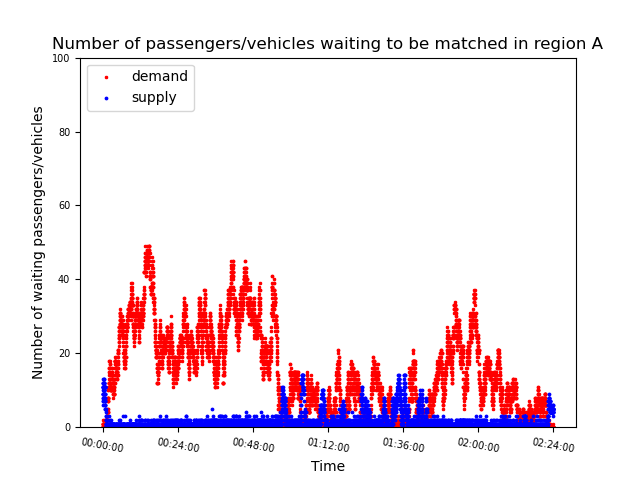}\vspace{1.5pt} 
    \includegraphics[width=5cm]{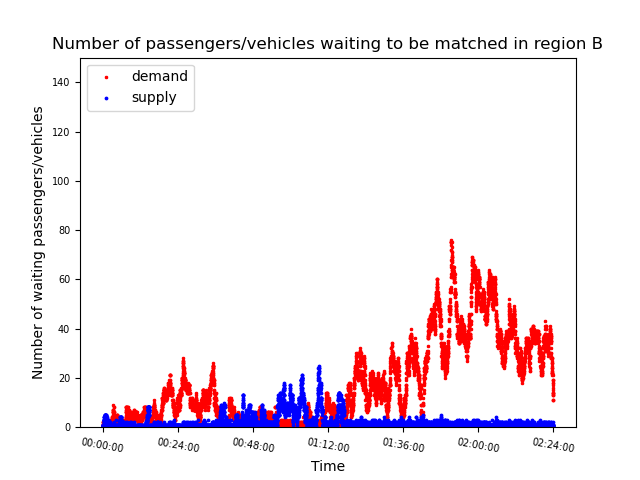}\vspace{1.5pt}
    \includegraphics[width=5cm]{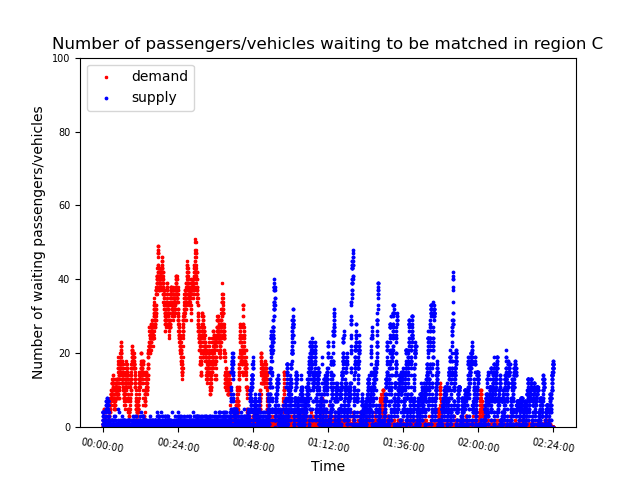}\vspace{1.5pt}
    \includegraphics[width=5cm]{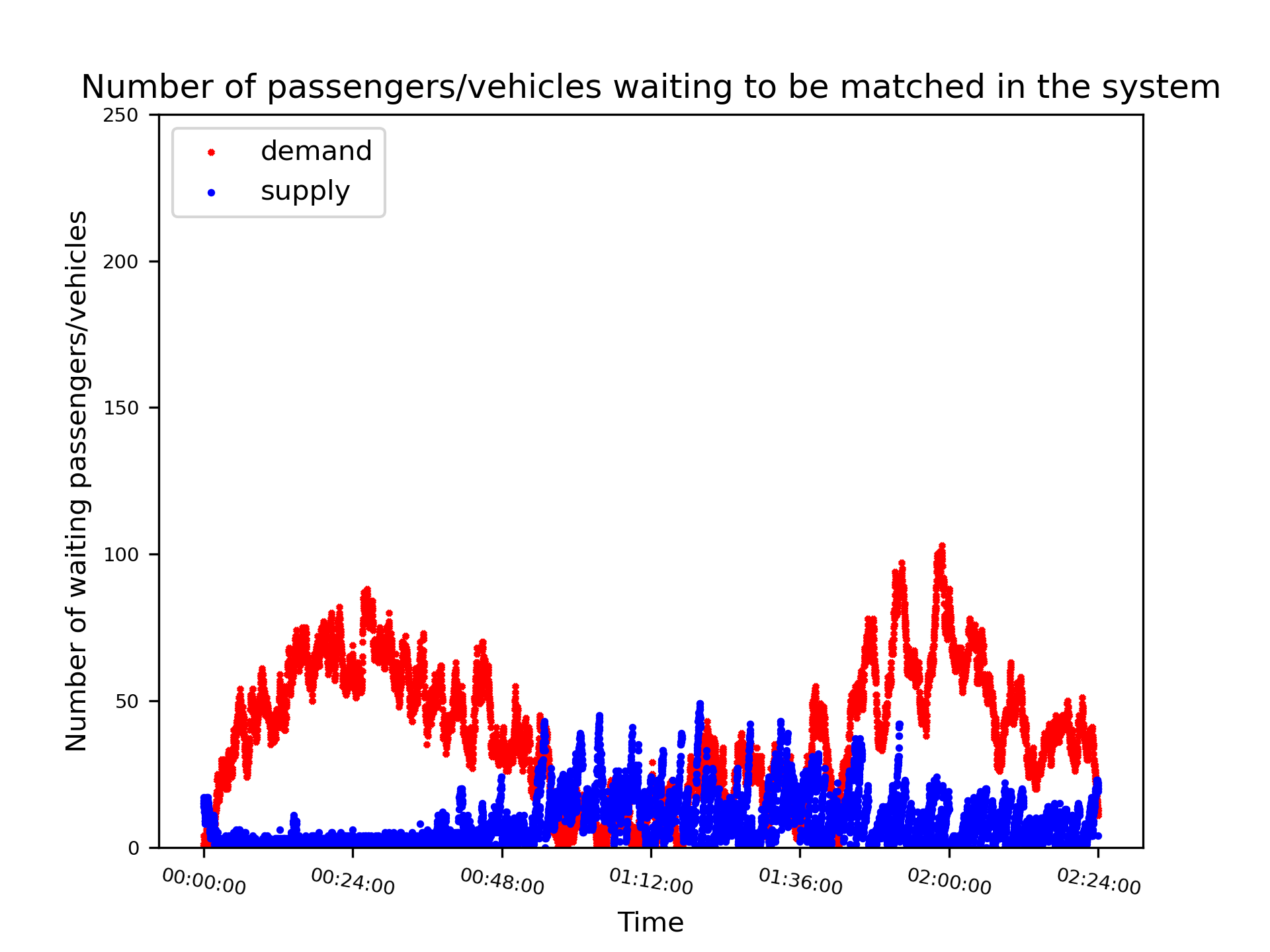}\vspace{1pt}
    \end{minipage}
}
\caption{Number of waiting passengers and vehicles under three methods in the toy model.}
\label{fig:number_t1}
\end{figure}

The numerical results also demonstrate the effectiveness of the MMA framework as a centralized method for exploiting joint matching and relocation. The relocation is triggered every 10 minutes in MMA and the RL-based method. When relocation is included, the MMA also shows an advantage over the RL-based method on the number of completed requests and the number of relocations, with more than $18\%$ improvement in requests completion rate by relocating vacant vehicles from $C$ to $A$ and $B$ to satisfy more requests. Apart from the quantitative results, we also visualize the temporal dynamics of the number of waiting passengers and vacant vehicles in each region and overall system when utilizing batch matching and MMA, as shown in Figure \ref{fig:number_t1}. It is obvious that region $C$ has an excessive supply, while regions $A$ and $B$ face a shortage of available vehicles if the destination information of requests is not incorporated into the myopic decision-making process (e.g., using batch matching). In contrast, MMA utilizes the prediction information to develop a farsighted matching strategy, leading to a noticeable reduction in the number of waiting passengers. Thus, we can claim that MMA effectively integrates the requests' destination information and future prediction information to enhance system performance.

Moreover, the strategic-layer decision making process in the MMA framework is a multi-objective optimization problem to optimize the number of completed requests, the relocation cost, and the penalty for regional imbalance. The multi-objective optimization is converted to a single-objective optimization by the weighted sum method with weight 1, $\alpha$, and $\beta$. The sensitivity analysis on the values of $\alpha$ and $\beta$ is conducted in Appendix \ref{sensi_toy_model}.

\subsection{A Realistic Network}
We test the capability of MMA in improving system performance in a more complicated environment using the real-world data in Chengdu, which was collected by the Software-as-a-Service platform of Gaode. The details on the dataset are shown in Appendix \ref{addi_real_net}. 

\subsubsection{Experimental Results.}
\label{results_real}
\paragraph{}
There are 78947 requests on average and 3,536 vehicles per day in the system. Table \ref{table:real_result} compares the MMA's average performance with the benchmarks.  

\begin{table}[!ht]
  \footnotesize
  \renewcommand\arraystretch{1.5}
  \centering
  \caption{Results of the realistic network.}
  \begin{tabularx}{\textwidth}{m{3cm}m{2.8cm}m{2.6cm}m{1.6cm}m{2.2cm}m{2.2cm}}
    \hline
   \textbf{Method} & \textbf{Number of completed requests}  &\textbf{Requests completion rate ($\%$)} &\textbf{Improve- ment ($\%$)} &\textbf{Average pick-up distance}  & \textbf{Average relocation times}\\
    \hline
    FCFS & 56613 & 72.06 & -1.94 & 2.7477 & 0\\
    Batch matching & 57731 & 73.45 & 0.00 & 1.7679 & 0\\
    RL-based without \newline relocation & 63554 & 80.69& 10.09 & 1.8362 & 0 \\
    RL-based with \newline relocation & 63935 & 81.16 & 10.75 & 1.8561 & 883 \\
    MMA without \newline  relocation & \textbf{63960} & 81.26  & 10.79 & 1.8491 & 0 \\
    MMA \newline $\alpha=0.5,\beta=0$ & 65552 & 83.30 & 13.55 & 1.8621 & 1400 \\
    MMA \newline $\alpha=0.5,\beta=0.2$ & \textbf{65910} & 83.76 & 14.17 & 1.8794 & 3216 \\
    \hline
  \end{tabularx}
  \label{table:real_result}
\end{table}

\begin{figure}[!htb]
\centering
\subfigure[Batch matching] {\includegraphics[width=8 cm]{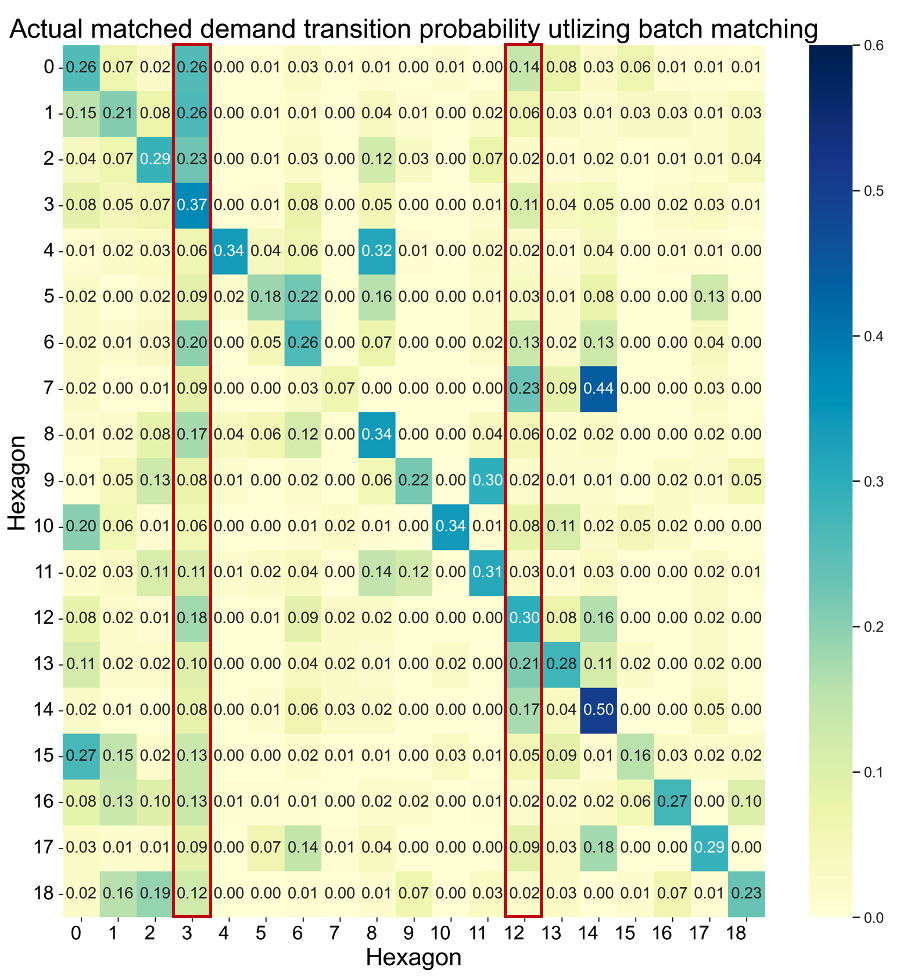}}
\subfigure[MMA without relocation] {\includegraphics[width=8cm]{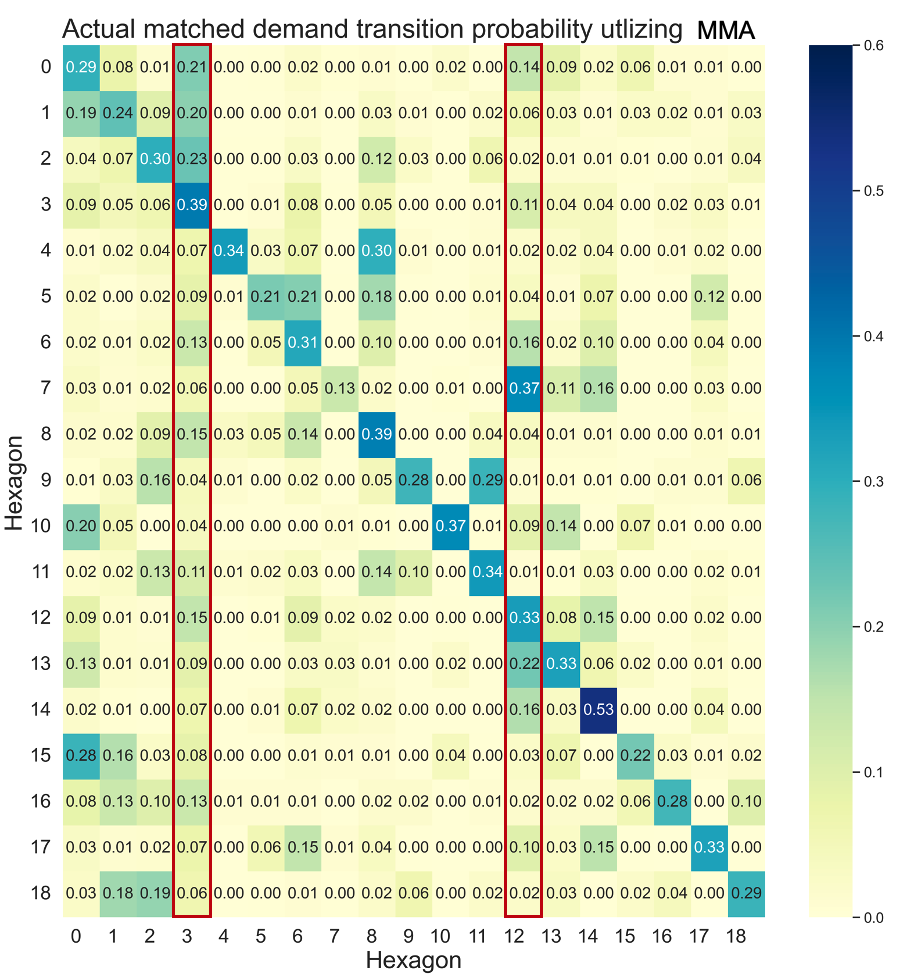}}
\caption{The transition probability matrix of actual matched demand.}
\label{fig:acutal_tran_heatmap}  
\end{figure}

For the cases without relocation, MMA consistently exhibits superior performance compared to all other strategies. The heat map in Figure \ref{fig:acutal_tran_heatmap} demonstrates the transition probability of actual matched demand from 8:00 a.m. to 12:00 a.m. under batch matching and the MMA without relocation. The number in row $i$ and column $j$ represents the proportion of requests towards hexagon $j$ accounting for total matched requests in hexagon $i$. It is evident that MMA alters the actual matching probability of requests heading for different hexagons. For instance, the matching probability of requests for hexagon 12 increases and that of hexagon 3 drops. The phenomenon is consistent with the fact that the strategic layer prioritizes requests which end in more populated hexagons, thereby minimizing the imbalance. 

When incorporating relocation, MMA yields a notable performance improvement. The number of completed requests exhibits an increment ranging from $13.55\%$ to $14.17\%$ compared to batch matching. Furthermore, MMA demonstrates a consistent advantage over the RL-based method when relocation exists. The possible explanation is that the vehicles in the RL-based method with relocation tend to maximize their own long-term profit, which may lead to unnecessary competition among vehicles. In contrast, MMA is a centralized controller that utilizes SLM to systematically coordinate vehicle flow spatial transfer patterns. Finally, it is also observed that the improvement brought by MMA in the realistic network is not as significant as that in the toy network. The possible reasons could include two folds. First, in the realistic dataset, supply shortages occur in most hexagons during peak hours, yielding limited space for MMA to relocate and filter requests by destination to improve system performance. Second, the MMA framework is more effective when regional imbalances are obvious. However, compared to the toy network, demands in Chengdu are relatively even distributed during the off-peak period.

\subsubsection{Performance of Lagrangian relaxation method.}
\paragraph{}

We compare the performance of the LR method with the off-the-shelf solver CPLEX. The tests are conducted on the setting of $\alpha=0.5$ and $\beta=0.2$, which yields the best performance in Table \ref{table:real_result}. The LR method terminates if the duality gap is less than 0.03 or 50 iterations are completed, and CPLEX is terminated if the optimal solution is found or the computation time reaches five minutes. Table \ref{LR_iteration} shows that the average results of the Lagrangian duality gap (between the upper and lower bounds obtained by the LR method), the gap of the SLM objective function values between CPLEX and LR (CPLEX yields a higher value because of more stringent termination criteria), and CPU time utilizing LR method. The average CPU time required for CPLEX to solve the real-world instances of SLM is 97.662 seconds. It is observed that compared to CPLEX, the LR method is capable of finding a solution with a marginal gap while requiring significantly less CPU time. We further reduce the maximum iteration number of LR and test its impact on LR's performance. Table \ref{LR_iteration} shows that after reducing the iteration limit from 50 to 10, the LR method is capable of achieving a saving of $66.66\%$ in computation time at the cost of only a $0.22\%$ increase in the duality gap and a $0.42\%$ increase in the objective-function gap. The results demonstrate the capability of the LR method in striking a good balance between computational efficiency and effectiveness, thereby highlighting its great potential for practical implementation in real-word applications.

\begin{table}[!ht]
  \footnotesize
\renewcommand\arraystretch{1.5}
  \centering
  \caption{Performance of the LR method with different maximum iteration number.}
  \begin{tabularx}{\textwidth}{m{2cm}m{4.4cm}m{4.8cm}m{4.0cm}}
   \hline
   \textbf{Instance} & \textbf{Lagrangian duality gap ($\%$)}  &\textbf{Objective-function gap ($\%$)} &\textbf{Average CPU time \newline utilizing LR method (s)} \\
    \hline
    50 & 1.1313 & 0.5736 & 34.3911 \\
    40 & 1.1313 & 0.5736 & 28.6786 \\
    30 & 1.1316 & 0.5738 & 22.9964 \\
    20 & 1.1330 & 0.5752 & 17.2965 \\
    10 & 1.1338 & 0.5760 & 11.4675\\

    \hline
  \end{tabularx}
  \label{LR_iteration}
\end{table}

\subsubsection{Impact of Inaccurate Prediction.} 
\paragraph{}

Considering that the prediction accuracy of demand and supply in practice may fluctuate in different networks, we test the robustness of the MMA framework against the prediction accuracy. Set $\alpha = 0.5$, and $\beta = 0$. We adopt the dataset on November $2^{nd}$, and we multiply the original predictions by random variables sampled from the uniform distribution with ranges [0.9,1.1], [0.8,1.2], [0.7,1.3], [0.6,1.4], and [0.5,1.5], respectively, to mimic different levels of prediction inaccuracy. The perturbed predictions are referred to as fluctuations within 10$\%$, 20$\%$, 30$\%$, 40$\%$, and 50$\%$, respectively. 

Table \ref{inaccurate_prediction} shows the performance of MMA utilizing inaccurate prediction results in the SLM model. The number of completed requests varies within a narrow range as prediction accuracy decreases, and the MMA maintains an over $10\%$ improvement compared to batch matching. The negative impact of inaccurate prediction can be primarily seen in the increase in ineffective relocation. Overall, MMA achieves a robust performance against the prediction inaccuracy. The possible explanation includes two folds. First, whether a hexagon is undersupplied or oversupplied depends on the estimation of vacant vehicles and waiting requests, which contain both the prediction results and the state transition from the previous intervals. For many hexagons, the state transition, instead of the prediction results, may account for a large proportion of the total number of requests or vehicles. Second, MMA framework deliberately incorporates the priority of fulfilling the current orders over vehicle relocation. Thus, the impact of inefficient vehicle relocation due to inaccurate prediction is rather limited.

\begin{table}[!ht]
  \footnotesize
  \renewcommand\arraystretch{1.5}
  \centering
  \caption{Results of additional experiments with inaccurate predictions.}
  \begin{tabularx}{\textwidth}{m{3.5cm}m{2.7cm}m{2.7cm}m{1.6cm}m{2.3cm}m{1.8cm}}
    \hline
   \textbf{Perturbation \newline amplitude} & \textbf{Number of completed requests}  &\textbf{Requests completion rate ($\%$)} &\textbf{Improve- ment ($\%$)} &\textbf{Average pick-up distance}  & \textbf{Relocation times}\\
    \hline
    Batch matching & 55300 & 75.26 & 0.00 & 1.8103 & 0 \\
    Original prediction & 62691 & 85.31 & 13.37 & 1.9411 & 1578\\
    Fluctuation within 10$\%$ & 62999 & 85.73 & 13.92 & 1.9442 & 2171\\
    Fluctuation within 20$\%$ & 62985 & 85.71 & 13.90 & 1.9194 & 2444 \\
    Fluctuation within 30$\%$ & 62925 & 85.63 & 13.79 & 1.8958 & 2518 \\
    Fluctuation within 40$\%$ & 61513 & 83.71  & 11.24 & 1.8439 & 1964 \\
    Fluctuation within 50$\%$ & 61288 & 83.40 & 10.83 & 1.8711 & 2246 \\
   \hline
  \end{tabularx}
  \label{inaccurate_prediction}
\end{table}

\subsubsection{Impact of Irregular Events.} 
\paragraph{}
The purpose of this numerical experiment is to assess the robustness of the MMA framework against the irregular events that are out of the scope of any prior knowledge. To mimic the impact of irregular events on the demand pattern, 400 new requests are generated every hour in hexagons 2 and 5, and 200 requests are dropped every hour in hexagons 3, 6, 12, and 14 in the morning peak. In the evening peak, 200 new requests are generated in hexagon 8, and 200 requests are dropped in hexagon 3 every 30 minutes from 16:30 to 20:00. Such manipulation changes the hot-spot location in the study area during rush hours. MMA and RL-based methods still utilize the originally-trained prediction model and state-value function, respectively, because the irregular events are not in the prior knowledge. Table \ref{table:sudden event} shows the performance of different strategies under irregular events.  


\begin{table}[!ht]
  \footnotesize  \renewcommand\arraystretch{1.5}
  \caption{Results of realistic network with irregular events}
  \begin{tabularx}{\textwidth}{m{3cm}m{2.8cm}m{2.6cm}m{1.6cm}m{2.2cm}m{2.2cm}}
    \hline
   \textbf{Method} & \textbf{Number of completed requests}  &\textbf{Requests completion rate ($\%$)} &\textbf{Improve- ment ($\%$)} &\textbf{Average pick-up distance}  & \textbf{Average relocation times}\\
    \hline
    FCFS & 54518 & 69.34 & -2.27 & 2.7503 & 0\\
    Batch matching & 55784 & 70.92 & 0.00 & 1.7707 & 0\\
    RL-based without \newline relocation & 61231 & 77.70 & 9.76 & 1.8343 & 0 \\
    RL-based with \newline relocation & 61619 & 78.19 & 10.46 & 1.8561 & 969 \\
    MMA without \newline  relocation & \textbf{61929} & 78.68  & 11.02 & 1.8451 & 0 \\
    MMA \newline $\alpha=0.5,\beta=0$ & 64180 & 81.55 & 15.05 & 1.8801 & 1523 \\
    MMA \newline $\alpha=0.5,\beta=0.2$ & \textbf{64511} & 81.99 & 15.64 & 1.8878 & 3311 \\
    \hline
  \end{tabularx}
  \label{table:sudden event}
\end{table}

\begin{figure*}[!htb]
    \centering
    \subfigure[Hexagon 2]{\includegraphics[width=5.6cm]{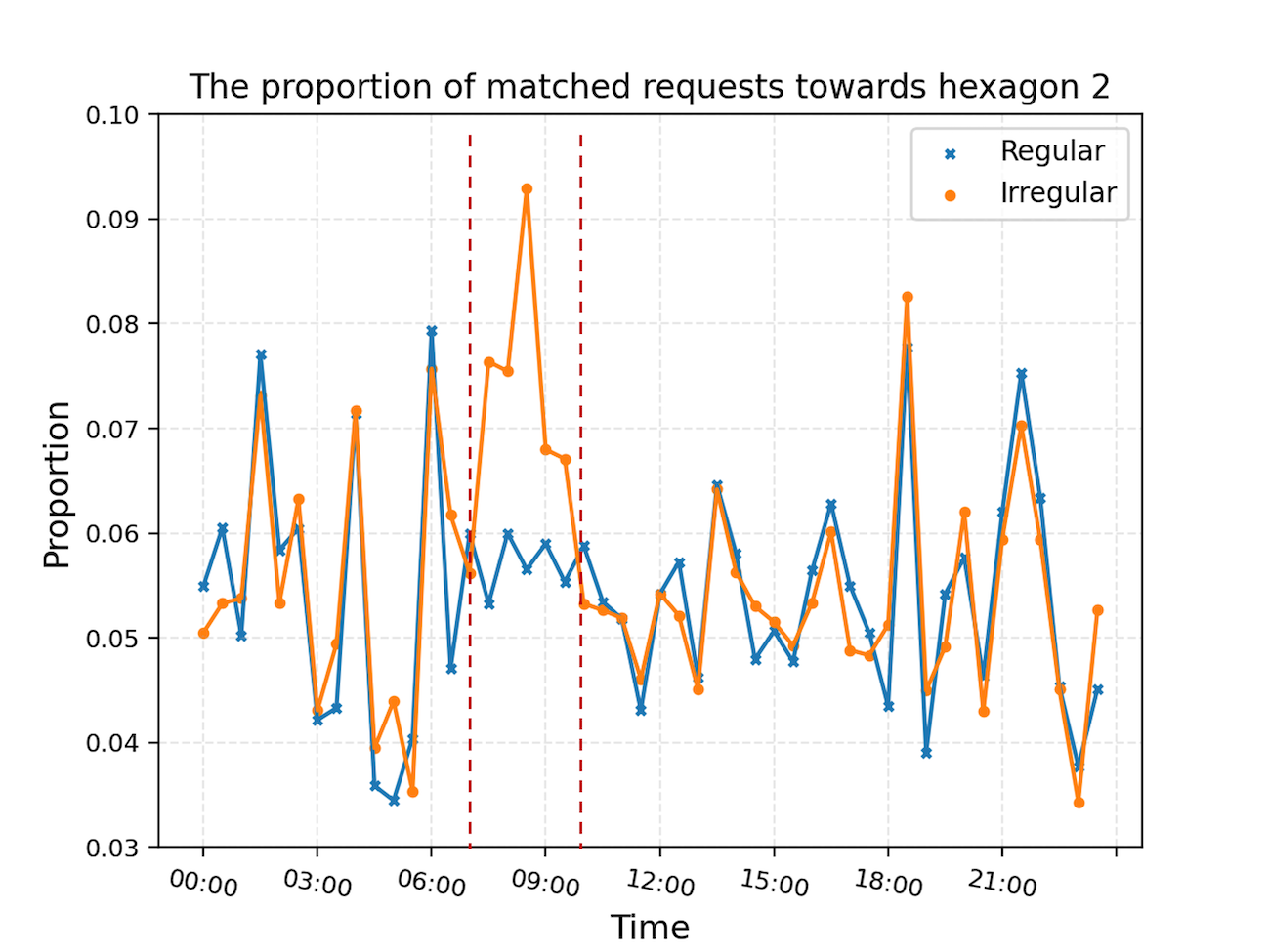}\label{figure/Re_hexagon2}}\hspace{-0.4cm}
    \subfigure[Hexagon 3]{\includegraphics[width=5.6cm]{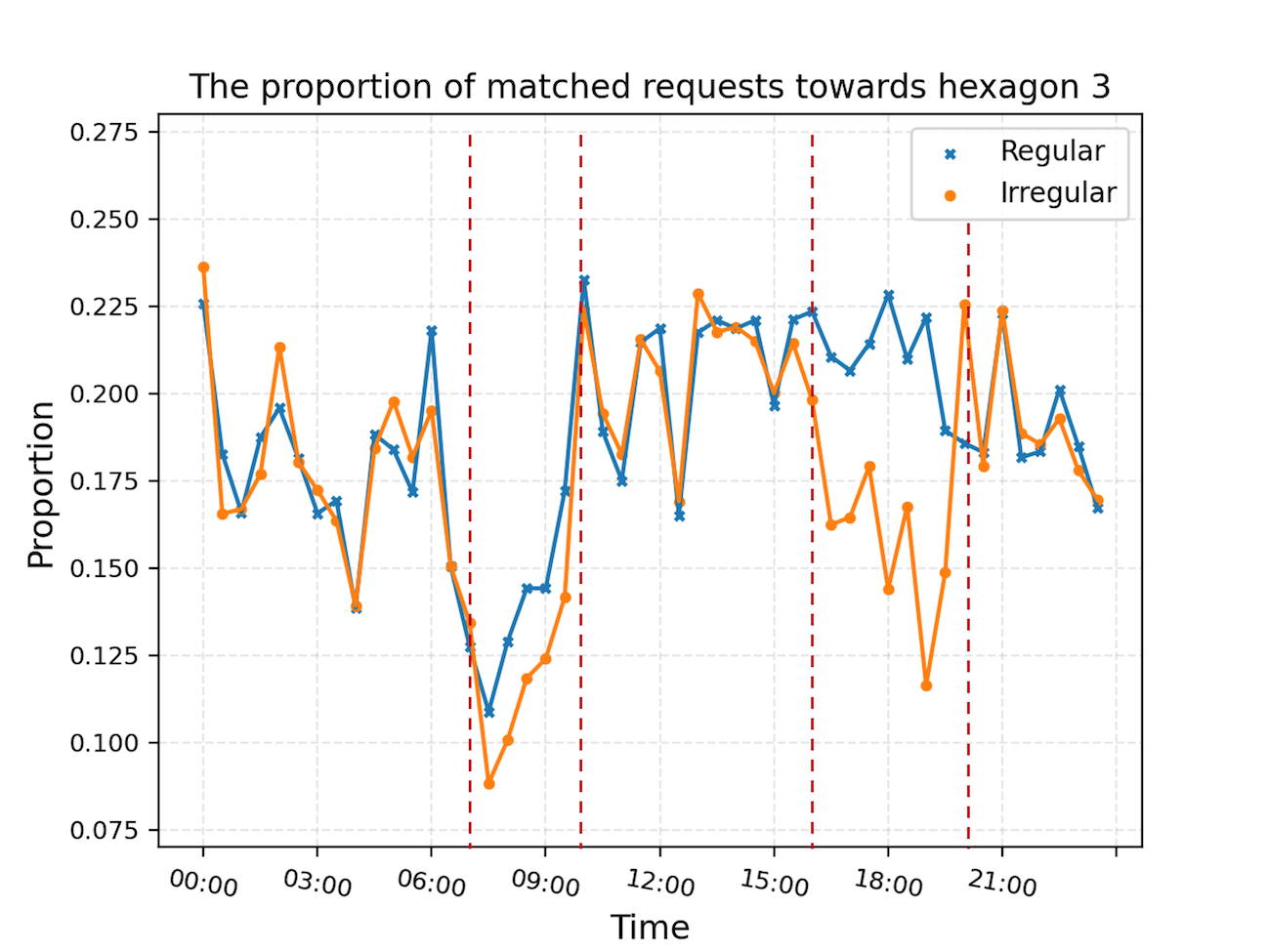}\label{figure/Re_hexagon3}} \hspace{-0.4cm}
    \subfigure[Hexagon 8]{\includegraphics[width=5.6cm]{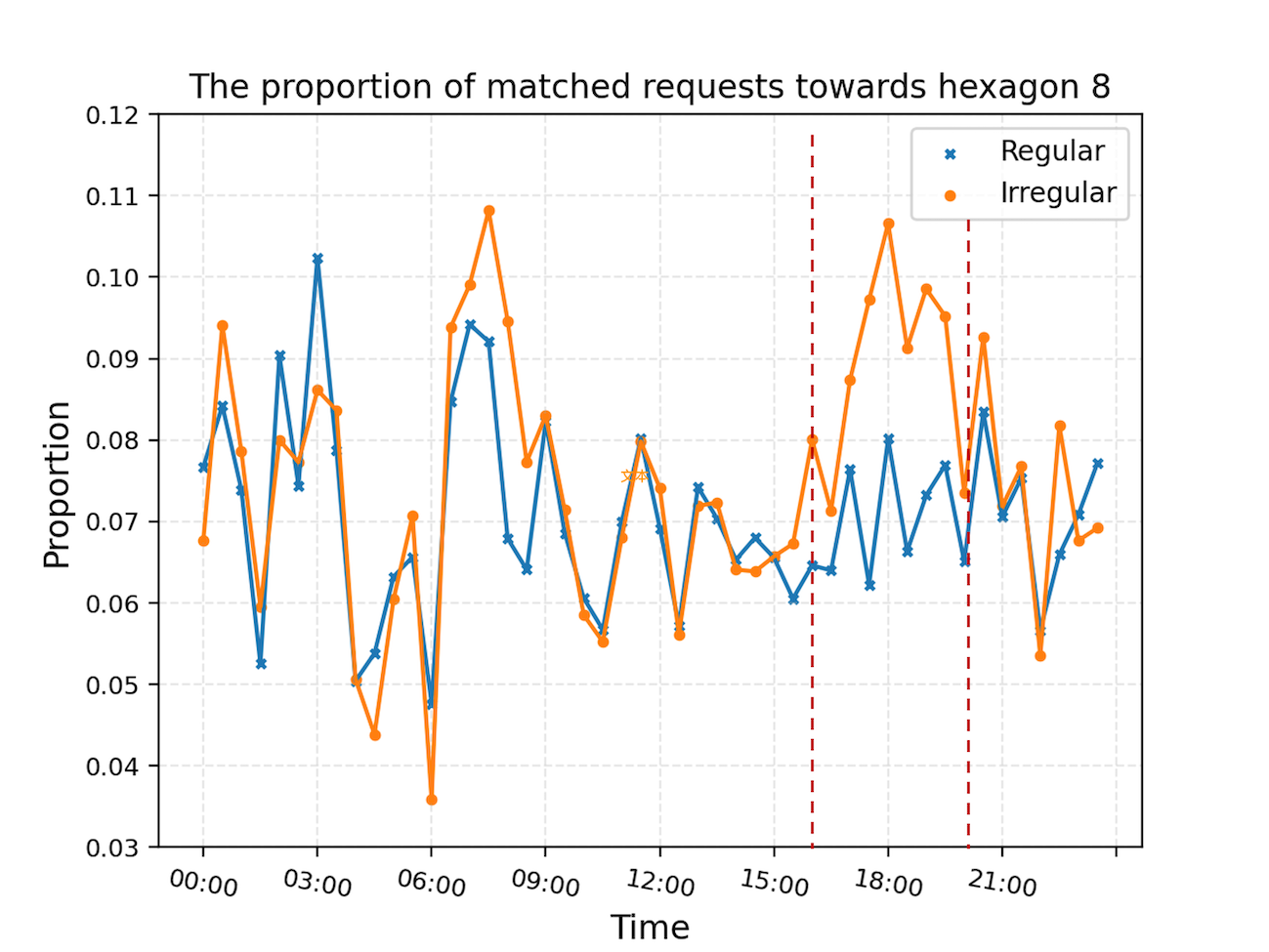}\label{figure/Re_hexagon8}}
    \caption{Proportion of matched requests towards each hexagon accounting for the total matched requests.}
    \label{hexagon_rate}
\end{figure*}

Comparing Tables \ref{table:real_result} and \ref{table:sudden event}, MMA achieves a more significant advantage in system performance over all the other benchmark policies. The MMA framework is capable of adapting rapidly to surges or declines in demand because the prediction model in SLM incorporates the multi-dimensional temporal demand information including the one in adjacent intervals of the day. To visualize this, we plot the curves of the proportion of matched requests ending in specific hexagons accounting for the total matched requests under MMA with and without irregular events in Figure \ref{hexagon_rate}. It can be observed that in response to the newly-generated requests in hexagon 2 from 8:00 a.m. to 10:00 a.m., MMA rapidly increases the successful-matching probability of requests towards hexagon 2. A similar phenomenon can be observed in hexagons 3 and 8 during the evening peak hours. Overall, MMA is capable of rapidly adapting to the variations in demand through capturing the change in the prediction model and then adjusting the matching probability of requests, accordingly. Furthermore, irregular events known in advance have the potential to significantly alter supply and demand patterns as well. For instance, events like concerts and sports competitions can lead to spikes in demand. The incorporation of prior knowledge pertaining to these temporary events into the prediction model, equips the MMA framework with the capability to formulate adaptive strategies in response to the dynamics of demand and supply. We note that the RL-based model can also increase its robustness to irregular events by expanding the dimension of the state vector. However, doing this will inevitably bring huge model-training efforts as the state-action space along with the temporal dimension could be prohibitively large. MMA utilizes a two-layer modular and lightweight model structure, enabling it to achieve a high level of robustness against demand variations while maintaining a relatively low computational cost.

\section{Conclusion}
\label{conclusion}

This study proposes an innovative MMA approach to optimize online matching and relocation strategies for ride-hailing service systems. MMA employs a two-layer and modular modeling structure. The strategic layer operates on a large timescale and optimizes the spatial transfer patterns of vehicle flow within the system to maximize the total revenue of the current and future stages. The lower-layer model performs rapid vehicle-to-order matching and relocates vacant vehicles based on the guidance provided by the strategic layer. We design an LR-based algorithm to efficiently solve the upper-layer model, which is proven to achieve global optimum in stylized networks and shows superior performance in the numerical experiments based on the realistic dataset. We propose customized and polynomial-time algorithms to globally solve the lower-layer model. MMA is interpretable, equipped with polynomial solution algorithms, and easy to implement, which is crucial for practical applications. Numerical studies on the real-world data in Chengdu reveal that in relocation-free scenarios, MMA increases the request completion rate by 10.79\% compared with batch matching; in the scenarios with relocation, MMA outperforms RL-based methods by more than 3.09\% improvement in the number of completed requests. Furthermore, it is also observed that the independence between the upper and lower-layer models enables MMA to achieve a high level of robustness against demand variations while maintaining a relatively low computational cost. With irregular events, compared to the RL-based method with relocation, MMA with relocation further increases the number of completed requests by 4.69\%. As the prediction accuracy on supply and demand decreases, MMA maintains a relatively stable performance in the order completion rate.

Future work may consider developing operational strategies in a competitive mobility market with multiple ride-hailing platforms. The multi-homing behaviors of drivers and customers, and the game-theoretical interactions among different platforms will further complicate the modeling framework. In addition, we may explore adopting and extending the proposed two-layer, and modular modeling structure to solve other dynamic operational problems in demand-responsive mobility services, such as the dynamic scheduling and passenger assignment in flexible bus service.

\ACKNOWLEDGMENT{
The research is supported in part by grants from Tsinghua-Toyota Joint Research Institution.}

\bibliographystyle{apalike} 
\bibliography{sample} 

\begin{APPENDICES}

\section{Nomenclature}
\setcounter{table}{0}
\renewcommand{\thetable}{A-\arabic{table}}
\renewcommand*{\theHtable}{\thetable}

\begin{table}[!ht]
  \footnotesize
  \centering
  \caption{Notations for the strategic layer models}
  \begin{tabularx}{\textwidth}{lX}
    \hline
    \textbf{Notations} & \textbf{Explanation} \\
    \hline
    \multicolumn{2}{l}{\textit{Sets}} \\
    $\mathcal{R}$ & The set of all hexagonal grids\\
    $\mathcal{T}$ & The set of all the strategic intervals\\
	$\mathcal{T}_{k\rightarrow p}$ & The set of $p$ consecutive strategic intervals starting from interval $k$ \\
		  & \\
		  
	\multicolumn{2}{l}{\textit{Parameters}} \\
	$x_{i,(k-j),r}$ & The number of newly-emerging demands in hexagon $r$ in interval $k-j$ on the same day of week in the past $i$ weeks \\
	$p$ & The number of planning intervals\\
	$\check{e}_{t,r}^{s}$ &  The number of relocated vehicles arriving at hexagon $r$ in interval $t$ that leaving their origin before interval $k$\\
	$\check{o}_{t,r}^{s}$ & The number of occupied vehicles arriving at hexagon $r$ in interval $t$ that leaving their origin before interval $k$\\
	$l_{(k-1),r}^{s}$ & The number of vehicles remaining from the previous interval $k-1$ in hexagon $r$\\
	$l_{(k-1),r,j}^{d}$ & The number of unmatched passengers at the end of interval $k-1$ in hexagon $r$ heading for $j$\\
	$\mu_{t}^{s}$ & The dropping rate of unmatched vehicles in interval $t$\\
	$\mu_{t}^{d}$ & The dropping rate of unmatched demands in interval $t$\\
	$b_{i\rightarrow j}^{t}$ & The proportion of demands towards hexagon $j$ from $i$ in interval $t$ \\
	$a[ij]$ & The number of intervals required to travel from hexagon $i$ to $j$\\
	$\alpha$ & The weight to penalize number of relocation in SLM\\
	$\beta$ & The weight to penalize the imbalance of vehicles and demands in operating area in SLM \\
	$\Delta T$ & The length of a strategic interval \\
		  & \\
		  
	\multicolumn{2}{l}{\textit{Variables}} \\
	$N_{t,r}^{s}$ & The number of vacant vehicles in hexagon $r$ in interval $t$ \\
	$N_{t,r}^{d}$ & The number of waiting customers in hexagon $r$ in interval $t$\\
	$\widehat{N}_{t,r}^{s}$ & The predicted number of newly-emerging vehicles in hexagon $r$ in interval $t$\\
	$\widehat{N}_{t,r}^{d}$ & The predicted number of newly-generated demands in hexagon $r$ in interval $t$\\
    $\check{E}_{t,r}^{s}$ & The number of relocated vehicles arriving at hexagon $r$ in interval $t$ that leaving their origins after interval $k$\\
	$\tilde{E}_{t,r}^{s}$ & The number of relocated vehicles arriving at hexagon $r$ in interval $t$\\
	$\check{O}_{t,r}^{s}$ & The number of occupied vehicles arriving at hexagon $r$ in interval $t$ that leaving their origin after interval $k$\\
	$\tilde{O}_{t,r}^{s}$ & The number of occupied vehicles arriving at hexagon $r$ in interval $t$\\
	$L_{t,r}^{s}$ & The number of vehicles remaining from interval $t$ in hexagon $r$\\
	$L_{t,r,j}^{d}$ & The number of unmatched passengers at the end of interval $t$ in hexagon $r$ heading for hexagon $j$\\
	$D_{t,r}$ & Absolute value of the difference between the gap in the numbers of idle vehicles and waiting passengers in hexagon $r$ and the average gap in all hexagons in interval $t$ \\
	$F_{t,r}$ & The number of matched demands in hexagon $r$ in interval $t$\\
	$A_{t,r}^{d}$,$A_{t,r}^{s}$ & Auxiliary variables, $A_{t,r}^{d},A_{t,r}^{s} \in \{0,1\}$ \\

	  & \\
	
    \multicolumn{2}{l}{\textit{Decision variables}} \\
    $M_{i,j}^{t}$ & The number of vehicles to serve the passengers from hexagon $i$ to $j$ in interval $t$ \\
    $E_{i,j}^{t}$ & The number of vacant vehicles to be relocated from hexagon $i$ to $j$ in interval $t$ \\
    \hline
  \end{tabularx}
  \label{Notataion_tab}
\end{table}

\section{Proof}
\label{proof_pro}
This section provides complete proofs for the propositions presented in the main paper.
\subsection{Proof of Proposition \ref{LRoptimality_prop}}

Before proceeding with the proof, we first need to introduce some notations. The travel demand in interval $t$ is denoted as a vector $\vec{\mathbf{D}}_{t} = \{D_{t, i \rightarrow j}\mid i,j \in \mathcal{R}\}$; the demand information for all intervals is denoted as a vector $\vec{\mathbf{D}}$; the vehicle distribution at the beginning of interval $t$ is denoted as a vector $\vec{\mathbf{N}}_{t}^{s} = \{ N_{t,i}^{s} \mid i \in \mathcal{R}\}$; the matching decision in interval $t$ is denoted as a vector $\vec{\mathbf{M}}_{t} = \{M_{i,j}^{t} \mid i,j \in \mathcal{R}\}$; the relocating decision in interval $t$ is denoted as a vector $\vec{\mathbf{E}}_{t} = \{ E_{i,j}^{t} \mid i,j \in \mathcal{R}\}$. We define two value functions as follows:
\begin{itemize}
    \item The value function $V_{t}(\vec{\mathbf{N}}_{t}^{s}; \vec{\mathbf{D}})$ indicates the maximum number of completed requests in intervals $\{ t,t+1,\dots,\bar{t} \}$ with the original SLM model given the full-scale demand information $\vec{\mathbf{D}}$ and the current vehicle distribution information $\vec{\mathbf{N}}_{t}^{s}$.

    \item The value function $\widehat{V}_{t}(\vec{\mathbf{N}}_{t}^{s}; \vec{\mathbf{D}})$ indicates the maximum number of completed requests in intervals $\{ t,t+1,\dots,\bar{t} \}$ with the relaxed SLM model without constraints \eqref{aux_eq4} and \eqref{aux_eq5} given the full-scale demand information $\vec{\mathbf{D}}$ and the current vehicle distribution information $\vec{\mathbf{N}}_{t}^{s}$.
\end{itemize}

The primary goal of this proof is to assert that $\widehat{V}_{t}(\vec{\mathbf{N}}_{t}^{s};\vec{\mathbf{D}}) = V_{t}(\vec{\mathbf{N}}_{t}^{s}; \vec{\mathbf{D}})$ for all $\vec{\mathbf{D}}, \vec{\mathbf{N}}_{t}^{s}$ and $t$. To achieve this aim, it is necessary to introduce the subsequent lemma.

\begin{lemma} \label{boundedness_lem}
Given any demand distribution $\vec{\mathbf{D}}$ and interval $t \in \mathcal{T}$, we have:
\begin{align*}
& \left |\widehat{V}_{t}(\vec{\mathbf{N}}_{t}^{s}; \vec{\mathbf{D}}) - \widehat{V}_{t}(\vec{\tilde{\mathbf{N}}}_{t}^{s}; \vec{\mathbf{D}})\right | \le \frac{1}{2} \left \|\vec{\mathbf{N}}_{t}^{s} - \vec{\tilde{\mathbf{N}}}_{t}^{s
} \right \|, \forall \ \vec{\mathbf{N}}_{t}^{s}, \vec{\tilde{\mathbf{N}}}_{t}^{s} .
\end{align*}
\end{lemma}

To prove Lemma \ref{boundedness_lem}, let us denote $\vec{\tilde{\mathbf{N}}}_{t}^{s} = \vec{\mathbf{N}}_{t}^{s} + \vec{\boldsymbol{\epsilon}}_{t}$, where $\vec{\boldsymbol{\epsilon}}_{t}$ has $\bar{n}-1$ patterns: $\bar{p}$ non-negative elements with $\bar{n}-\bar{p}$ negative elements, $\bar{p} = \{1,2, \dots, \bar{n}-1\}$. Without losing generality, we assume that $\widehat{V}_{t}(\vec{\tilde{\mathbf{N}}}_{t}^{s};\vec{\mathbf{D}}) \ge \widehat{V}_{t}(\vec{\mathbf{N}}_{t}^{s};\vec{\mathbf{D}})$. We investigate one of the patterns where there are $\bar{p}$ non-negative elements and $\bar{n}-\bar{p}$ negative elements, and without loss of generality we set $\epsilon_{t,1},\dots, \epsilon_{t,\bar{p}} \ge 0$. We set: $K_{t,p} = 0$, for $p \in \{ 1, \dots, \bar{p}\}$, and $K_{t,q} = \left | \epsilon_{t,q} \right | $, for $q \in \{ \bar{p}+1, \dots, \bar{n}\}$. Then, we have:
\begin{align*}
& N_{t,p}^{*} = N_{t,p}^{s} - K_{t,p} = \tilde{N}_{t,p}^{s} - \epsilon_{t,p}, \ \forall p \in \{1,\dots,\bar{p}\} \\
& N_{t,q}^{*} = N_{t,q}^{s} - K_{t,q} = \tilde{N}_{t,q}^{s}, \ \forall q \in \{\bar{p}+1, \dots, \bar{n}\}
\end{align*}

Vehicles in the vector $\vec{\mathbf{N}}_{t}^{*}$ are the ``freely controllable part" in $\vec{\mathbf{N}}_{t}^{s}$, and $\vec{\mathbf{N}}_{t}^{*} \le \vec{\tilde{\mathbf{N}}}_{t}^{s}$ (here ``$\le$" indicates that each element of the left-hand side vector is less than or equal to the associated element of the RHS vector). In the  $\vec{\mathbf{N}}_{t}^{s}$ case, we stipulate that each vehicle in $\vec{\mathbf{N}}_{t}^{*}$ replicates the optimal action of the associated vehicle in the $\vec{\tilde{\mathbf{N}}}_{t}^{s}$ case; and for vehicles in $\vec{\mathbf{K}}_{t}$, we instruct them to emulate the destination distribution of the associated vehicles in $\vec{\tilde{\mathbf{N}}}_{t}^{s}$ case, but without carrying any customers. With the above arrangement, the vehicle distributions in interval $t+1$ for both $\vec{\tilde{\mathbf{N}}}_{t}^{s}$ and $\vec{\mathbf{N}}_{t}^{s}$ cases are identical, and in interval $t$, the number of completed requests in the $\vec{\mathbf{N}}_{t}^{s}$ case is at most $\sum_{i = 1}^{\bar{p}}\epsilon_{t,i}$ fewer than that of the $\vec{\tilde{\mathbf{N}}}_{t}^{s}$ case. Since $\sum_{i = 1}^{\bar{p}}\epsilon_{t,i} = \frac{1}{2} \sum_{i =1}^{\bar{n}} \left | \epsilon_{t,i}\right | = \frac{1}{2} \left \| \vec{\mathbf{N}}_{t}^{s} - \vec{\tilde{\mathbf{N}}}_{t}^{s} \right \|$, we then acquire that $\widehat{V}_{t}(\vec{\tilde{\mathbf{N}}}_{t}^{s}; \vec{\mathbf{D}}) - \widehat{V}_{t}(\vec{\mathbf{N}}_{t}^{s}; \vec{\mathbf{D}}) \le \frac{1}{2} \left \|\vec{\mathbf{N}}_{t}^{s} - \vec{\tilde{\mathbf{N}}}_{t}^{s}\right \|$. \par 

The remaining patterns can be analyzed with similar techniques; hence, we omit the details here. By combining the all patterns, we conclude $\left | \widehat{V}_{t}(\vec{\tilde{\mathbf{N}}}_{t}^{s}; \vec{\mathbf{D}}) - \widehat{V}_{t}(\vec{\mathbf{N}}_{t}^{s}; \vec{\mathbf{D}})\right | \le \frac{1}{2} \left \|\vec{\mathbf{N}}_{t}^{s} - \vec{\tilde{\mathbf{N}}}_{t}^{s}\right \|$. Therefore, the proof of Lemma \ref{boundedness_lem} is complete.  \hfill $\square$ \\ \par 

With Lemma \ref{boundedness_lem}, we then proceed to prove the main proposition by mathematical induction. We aim to demonstrate that $\widehat{V}_{t}
(\vec{\mathbf{N}}_{t}^{s}; \vec{\mathbf{D}}) = V_{t}(\vec{\mathbf{N}}_{t}^{s}; \vec{\mathbf{D}})$ for all $\vec{\mathbf{D}},\vec{\mathbf{N}}_{t}^{s}$ and $t$. It straightforwardly holds for $t = \bar{t}$. Now, assuming that the above claim holds for all $t \ge t'$, we focus on the case of $t = t' - 1$. Given the vehicle distribution $\vec{\mathbf{N}}_{t'-1}^{s}$, we suppose that the optimal matching decision for the relaxed case is $\vec{\widehat{\mathbf{M}}}_{t'-1}$, and the corresponding vehicle distribution in interval $t'$ is $\vec{\widehat{\mathbf{N}}}_{t'}^{s}$. We further denote:
\begin{align*}
& \widehat{K}_{t',i} = \widehat{N}_{t',i}^{s} - \sum_{j =1}^{\bar{n}} \widehat{M}^{t'-1}_{j,i} & \forall i \in \{1,\dots,\bar{n}\}
\end{align*}

On the other hand, for the original case, we can always find a matching decision $\vec{\mathbf{M}}_{t'-1}$ which satisfies  $\vec{\mathbf{M}}_{t'-1} \ge \vec{\widehat{\mathbf{M}}}_{t'-1}$ in the vector sense, as the original case necessitates vehicles to serve all existing demands within a node. The vehicle distribution in interval $t'$ is denoted as $\vec{\mathbf{N}}_{t'}^{s}$ when matching decision $\vec{\mathbf{M}}_{t'-1}$ is applied. We further denote:
\begin{align*}
& J_{t',i} = \sum_{j =1}^{\bar{n}} \left( M^{t'-1}_{j,i} - \widehat{M}^{t'-1}_{j,i} \right) \ge 0 & \forall i \in \{1,\dots, \bar{n}\} \\
& K_{t',i} = N_{t',i}^{s} - \sum_{j =1}^{\bar{n}} M^{t'-1}_{j,i} & \forall i \in \{1,\dots,\bar{n}\} \\
& \bar{\epsilon}_{t',i} = \widehat{K}_{t',i} - J_{t',i}=\widehat{N}_{t',i}^{s} - \sum_{j=1}^{\bar{n}}M_{j,i}^{t'-1} & \forall i \in \{1,\dots,\bar{n}\}
\end{align*}

Apparently $\sum_{j =1}^{\bar{n}} \bar{\epsilon}_{t',i} \ge 0$. Subsequently, we proceed to discuss the following two main cases. \par 

\textit{Case I: $\bar{\epsilon}_{t',i} \ge 0$ for all $i \in \{1,2,\dots,\bar{n}\}$.} In this case, the original model possesses the flexibility to adjust the relocation decision to fulfill the condition $\vec{\mathbf{K}}_{t'} = \vec{\bar{\boldsymbol{\epsilon}}}_{t'}$, thereby resulting in $\vec{\mathbf{N}}_{t'}^{s} = \vec{\widehat{\mathbf{N}}}_{t'}^{s}$. Since $\sum_{i =1}^{\bar{n}} J_{t',i} \ge 0$, it follows that $V_{t'-1}(\vec{\mathbf{N}}_{t'-1}^{s}; \vec{\mathbf{D}}) \ge \widehat{V}_{t'-1}(\vec{\mathbf{N}}_{t'-1}^{s}; \vec{\mathbf{D}})$. Moreover, owing to the relaxed nature, we always have $V_{t'-1}(\vec{\mathbf{N}}_{t'-1}^{s}; \vec{\mathbf{D}}) \le \widehat{V}_{t'-1}(\vec{\mathbf{N}}_{t'-1}^{s}; \vec{\mathbf{D}})$. Consequently, we can conclude that $V_{t'-1}(\vec{\mathbf{N}}_{t'-1}^{s}; \vec{\mathbf{D}}) = \widehat{V}_{t'-1}(\vec{\mathbf{N}}_{t'-1}^{s}; \vec{\mathbf{D}})$. \par 

\textit{Case II: there is at least one $\bar{\epsilon}_{t',i} < 0, i \in \{1,2,\dots,\bar{n}\}$}. We assume that there are $\bar{p}$ ($\bar{p} \ge 1$) negative elements and $\bar{n} - \bar{p}$ non-negative elements. With loss of generality, we set $\bar{\epsilon}_{t',p} < 0, p \in \{1,\dots,\bar{p}\}$, and $\bar{\epsilon}_{t',q} \ge 0, q \in \{\bar{p}+1,\dots,\bar{n}\}$. In this case, the original model can freely arrange the relocation decision such that:
\begin{align*}
& K_{t',p} = 0, \  \forall 
p \in \{1,\dots,\bar{p}\} \\
& K_{t',q} \le \ \bar{\epsilon}_{t',q}  , \ \forall q \in \{\bar{p}+1, \dots, \bar{n}\}
\end{align*}

Furthermore, we have:
\begin{align*}
& N_{t',p}^{s} = \sum_{j =1}^{\bar{n}} M^{t'-1}_{j,p}>\widehat{N}_{t',p}^{s}, \  \forall p \in \{1,\dots,\bar{p}\} \\ 
& N_{t',q}^{s} \le \widehat{N}_{t',q}^{s}   , \ \forall q \in \{\bar{p}+1, \dots, \bar{n}\}
\end{align*}

The relocation decisions also result in $\left \| \vec{\mathbf{N}}_{t'}^{s} - \vec{\widehat{\mathbf{N}}}_{t'}^{s}\right \| = - 2 \sum_{p = 1}^{\bar{p}}\bar{\epsilon}_{t',p}$. The number of completed requests with matching decision $\vec{\mathbf{M}}_{t'-1}$ must be less than the maximum number of completed requests $V_{t'-1}(\vec{\mathbf{N}}_{t'-1}^{s};\vec{\mathbf{D}})$. With Lemma \ref{boundedness_lem}, we yield:
\begin{align*}
V_{t'-1}(\vec{\mathbf{N}}_{t'-1}^{s}; \vec{\mathbf{D}}) -\widehat{V}_{t'-1}(\vec{\mathbf{N}}_{t'-1}^{s}; \vec{\mathbf{D}}) 
& \ge \sum_{i =1}^{\bar{n}} J_{t',i} + V_{t'}(\vec{\mathbf{N}}_{t'}^{s}; \vec{\mathbf{D}}) - \widehat{V}_{t'}(\vec{\widehat{\mathbf{N}}}_{t'}^{s}; \vec{\mathbf{D}}) \\
& = \sum_{i =1}^{\bar{n}} J_{t',i} + \widehat{V}_{t'}(\vec{\mathbf{N}}_{t'}^{s}; \vec{\mathbf{D}}) - \widehat{V}_{t'}(\vec{\widehat{\mathbf{N}}}_{t'}^{s}; \vec{\mathbf{D}}) \\
&\ge \sum_{i =1}^{\bar{n}} J_{t',i} - \left | \widehat{V}_{t'}(\vec{\mathbf{N}}_{t'}^{s}; \vec{\mathbf{D}}) - \widehat{V}_{t'}(\vec{\widehat{\mathbf{N}}}_{t'}^{s}; \vec{\mathbf{D}}) \right | \\
&\ge  \sum_{i =1}^{\bar{n}} J_{t',i} + \sum_{p = 1}^{\bar{p}}\bar{\epsilon}_{t',p} \\
&= \sum_{p=1}^{\bar{p}}\widehat{K}_{t',p} + \sum_{i = \bar{p}+1}^{\bar{n}}J_{t'.i}\ge 0
\end{align*}

Combining $V_{t'-1}(\vec{\mathbf{N}}_{t'-1}^{s}; \vec{\mathbf{D}}) \le \widehat{V}_{t'-1}(\vec{\mathbf{N}}_{t'-1}^{s}; \vec{\mathbf{D}})$, we yield $V_{t'-1}(\vec{\mathbf{N}}_{t'-1}^{s}; \vec{\mathbf{D}}) = \widehat{V}_{t'-1}(\vec{\mathbf{N}}_{t'-1}^{s}; \vec{\mathbf{D}})$ in this case. \par 

In conclusion, based on the two main cases and employing the rule of induction, we have successfully proved that $\widehat{V}_{t}(\vec{\mathbf{N}}_{t}^{s}; \vec{\mathbf{D}}) = V_{t}(\vec{\mathbf{N}}_{t}^{s}; \vec{\mathbf{D}})$ for all $\vec{\mathbf{D}}, \vec{\mathbf{N}}_{t}^{s}$ and $t$. This result implies that the relaxation of constraints \eqref{aux_eq4} and \eqref{aux_eq5} does no yield any improvement in the number of completed requests in the stylized case. Furthermore, the optimal objective value of the Lagrangian dual of SLM does not exceed the optimal objective value of the relaxed SLM, which means that the Lagrangian duality gap of SLM is zero. \hfill $\square$

\subsection{Proof of Proposition \ref{optimal_MDM}}
The proposed algorithm pre-assigns $\min\big( \left \lfloor \frac{d_{r,j}^{k}}{\sum_{i \in \mathcal{R}} d_{r,i}^{k}}\times n_{r}^{s} \right \rfloor,n_{r,j}^{d})$ vehicles to serve the demands traveling from hexagon $r$ to $j$. We identify the set of hexagons with additional waiting demands after the pre-assignment, denoted as  $\mathcal{S} = \{ j \mid \left \lfloor \frac{d_{r,j}^{k}}{\sum_{i \in \mathcal{R}} d_{r,i}^{k}}\times n_{r}^{s}\right \rfloor < n_{r,j}^{d},  j \in \mathcal{R} \}$, and the number of available vehicles after pre-assignment, denoted as $\dot{n}_{r}^{s} = n_{r}^{s} - \sum_{j \in \mathcal{R}} \min\big( \left \lfloor \frac{d_{r,j}^{k}}{\sum_{i \in \mathcal{R}} d_{r,i}^{k}}\times n_{r}^{s} \right \rfloor, n_{r,j}^{d}\big)$. Then, we discuss the following two main cases, distinguished by the relationship between the size of $\mathcal{S}$ and $\dot{n}_{r}^{s}$.

\textit{Case 1: $\dot{n}_{r}^{s}< \mid \mathcal{S} \mid$.} We define $\theta_{j} =  \frac{d_{r,j}^{k}}{\sum_{i \in \mathcal{R}} d_{r,i}^{k}}\times n_{r}^{s} - \left \lfloor \frac{d_{r,j}^{k}}{\sum_{i \in \mathcal{R}} d_{r,i}^{k}}\times n_{r}^{s} \right \rfloor$, for all $j \in \mathcal{S}$. The set of hexagons with top $\dot{n}_{r}^{s}$ largest $\theta_{j}$ is denoted by $\mathcal{S}_{1}$ and $\mathcal{S}_{2} = \mathcal{S} \setminus \mathcal{S}_{1}$.

In this case, the matching vehicles allocation $\vec{\mathbf{x}}^{*} = \{x_{r,j}^{*}\}$ and the corresponding $\vec{\mathbf{c}}^{*} = \{c_{r,j}^{*}\}$ determined by Algorithm \ref{middle_algo} are presented below.  
\begin{itemize}
    \item For $j \in \mathcal{R}_{1} = \{ j \mid \left \lfloor \frac{d_{r,j}^{k}}{\sum_{i \in \mathcal{R}} d_{r,i}^{k}}\times n_{r}^{s} \right \rfloor \ge n_{r,j}^{d}\}$, $x_{r,j}^{*} = n_{r,j}^{d}, c_{r,j}^{*} = \frac{d_{r,j}^{k}}{\sum_{i \in \mathcal{R}} d_{r,i}^{k}} \times n_{r}^{s} - n_{r,j}^{d}$
    \item For $j \in \mathcal{R}_{2} = \{ j \mid \left \lfloor \frac{d_{r,j}^{k}}{\sum_{i \in \mathcal{R}d_{r,i}^{k}}} \right \rfloor < n_{r,j}^{d} \} \cap \mathcal{S}_{1}$ , $x_{r,j}^{*} = \left \lceil \frac{d_{r,j}^{k}}{\sum_{i \in \mathcal{R}}d_{r,i}^{k}} \right \rceil,  c_{r,j}^{*} = 0$
    \item For $j \in \mathcal{R}_{3} = \{ j \mid \left \lfloor \frac{d_{r,j}^{k}}{\sum_{i \in \mathcal{R}d_{r,i}^{k}}} \right \rfloor < n_{r,j}^{d} \} \cap \mathcal{S}_{2}$, $x_{r,j}^{*} =  \left \lfloor \frac{d_{r,j}^{k}}{\sum_{i \in \mathcal{R}}d_{r,i}^{k}} \right \rfloor, c_{r,j}^{*} = \frac{d_{r,j}^{k}}{\sum_{i \in \mathcal{R}} d_{r,i}^{k}} \times n_{r}^{s} - \left \lfloor \frac{d_{r,j}^{k}}{\sum_{i \in \mathcal{R}} d_{r,i}^{k}} \times n_{r}^{s} \right \rfloor$
\end{itemize}

The optimality of solution $(\vec{\mathbf{x}}^{*},\vec{\mathbf{c}}^{*})$ can be proved by contradiction. Let us assume that the optimal solution for MVA model is $(\vec{\check{\mathbf{x}}},\vec{\check{\mathbf{c}}})$. We denote $\vec{\check{\mathbf{x}}} = \vec{\mathbf{x}}^{*} + \vec{\boldsymbol{\varepsilon}}$, where the elements of $\vec{\boldsymbol{\varepsilon}} = \{\varepsilon_{j}\}$ are integers. To explicitly express $\vec{\check{\mathbf{c}}}$, we define the following sets: $\mathcal{R}^{+}_{2} = \{j \mid j \in \mathcal{R}_{2}, \varepsilon_{j} \ge 0 \}$, $\mathcal{R}^{-}_{2} = \{j \mid j \in \mathcal{R}_{2}, \varepsilon_{j} < 0 \}$, $\mathcal{R}^{+}_{3} = \{j \mid j \in \mathcal{R}_{3}, \varepsilon_{j} > 0 \}$, $\mathcal{R}^{-}_{3} = \{j \mid j \in \mathcal{R}_{3}, \varepsilon_{j} \le 0 \}$. The expression of $\vec{\check{\mathbf{c}}}$ is provided below.

\begin{itemize}
    \item For $j \in \mathcal{R}_{1}$, $\varepsilon_{j}$ should be non-negative to ensure the feasibility of constraint \eqref{eq30}, thus,  $\check{c}_{r,j} = c_{r,j}^{*} - \varepsilon_{j}$
    \item For $j \in \mathcal{R}_{2}^{+}$, $\check{c}_{r,j} = c_{r,j}^{*} = 0$
    \item For $j \in \mathcal{R}_{2}^{-}$, $\check{c}_{r,j} = \frac{d_{r,j}^{k}}{\sum_{i \in \mathcal{R}} d_{r,i}^{k}}\times n_{r}^{s} - \left(\left\lceil \frac{d_{r,j}^{k}}{\sum_{i \in \mathcal{R}} d_{r,i}^{k}}\times n_{r}^{s}\right\rceil + \varepsilon_{j}\right)= \theta_{j} - 1 -\varepsilon_{j}$
    \item For $j \in \mathcal{R}^{+}_{3}$, $\check{c}_{r,j} = 0$
    \item For $j \in \mathcal{R}^{-}_{3}$, $ \check{c}_{r,j} = \frac{d_{r,j}^{k}}{\sum_{i \in \mathcal{R}} d_{r,i}^{k}}\times n_{r}^{s} - \left(\left\lfloor \frac{d_{r,j}^{k}}{\sum_{i \in \mathcal{R}} d_{r,i}^{k}}\times n_{r}^{s}\right\rfloor + \varepsilon_{j}\right) = c_{r,j}^{*} - \varepsilon_{j}$
    
\end{itemize}

We denote the objective function values of MVA when decision variables are $(\vec{\check{\mathbf{x}}},\vec{\check{\mathbf{c}}})$ and $(\vec{\mathbf{x}}^{*},\vec{\mathbf{c}}^{*})$ as $V_{1}$ and $V_{2}$, respectively. $V_1$ and $V_2$ can be calculated as follows. 
\begin{align*}
V_1 & = \sum_{j \in \mathcal{R}_1}(c_{r,j}^{*}-\varepsilon_{j}) + \sum_{j \in R_{2}^{-}}(\theta_{j} - 1 - \varepsilon_{j}) + \sum_{j \in \mathcal{R}_{3}^{-}}(c_{r,j}^{*} - \varepsilon_{j}) \\
V_2 & = \sum_{j \in \mathcal{R}_1}c_{r,j}^{*} + \sum_{j \in \mathcal{R}_{3}}c_{r,j}^{*}
\end{align*}

The difference between $V_{1}$ and $V_{2}$ can be expressed as: 
\begin{align*}
   V_1 - V_2 = \sum_{j \in \mathcal{R}_1} (-\varepsilon_{j}) + \sum_{j \in \mathcal{R}_{2}^{-}}(\theta_j - 1 -\varepsilon_{j}) + \sum_{j \in \mathcal{R}_{3}^{+}}(-\theta_{j}) + \sum_{j\in \mathcal{R}_{3}^{-}}(-\varepsilon_j) 
\end{align*}

To ensure the feasibility of constraint \eqref{eq31}, it is necessary that $\sum_
{j \in \mathcal{R}} \varepsilon_{j} = 0 $. The non-positive and integer nature of $\varepsilon_{j}$ for all $j \in \mathcal{R}_{1} \cup \mathcal{R}_2^{-} \cup \mathcal{R}_{3}^{-}$ implies that $\sum_{j \in \mathcal{R}_1} -\varepsilon_{j} \ge 0$, $\sum_{j \in \mathcal{R}_2^{-}} \theta_j - 1 -\varepsilon_{j} \ge 0$ and $\sum_{j \in \mathcal{R}_3^{-}} -\varepsilon_{j} \ge 0$. If $\mathcal{R}_{2}^{+}$ is nonempty, there is at least one element $\varepsilon_j \le -1$ where $ j \in \mathcal{R}_{1} \cup \mathcal{R}_2^{-} \cup \mathcal{R}_{3}^{-}$, which leads to an improvement in the objective function value. Thus, we can conclude that $\mathcal{S}_{2}^{+} = \phi$, simplifying the difference between $V_1$ and $V_2$ as follows.
$$
V_1 - V_2 = \sum_{j \in \mathcal{R}_{2}^{-}} (\theta_{j} - 1) + \sum_{j \in \mathcal{R}_{3}^{+}}(\varepsilon_{j} - \theta_{j}) 
$$

The optimality of $(\vec{\check{\mathbf{x}}},\vec{\check{\mathbf{c}}})$ implies that $\varepsilon_{j} = 1, j \in \mathcal{R}_{3}^{+}$. Otherwise, the reduction in objective function value contributed to $R_{3}^{+}$ remains unchanged and the improvement in objective function value contributed to $\mathcal{R}_{1}\cup \mathcal{R}_{2}^{-} \cup \mathcal{R}_{3}^{-}$ increases. Therefore, it is evident that $\mid \mathcal{R}_{2}^{-} \mid < \mid \mathcal{R}_{3}^{+} \mid$ because $\varepsilon_{j} = 1, j \in \mathcal{R}_{3}^{+}$ , $\varepsilon_{j} \le -1, j \in \mathcal{R}_1 \cup \mathcal{R}_{2}^{-} \cup \mathcal{R}_{3}^{-}$ and the constraint $\sum_{j \in \mathcal{R}} \varepsilon_{j} = 0$. Recall that $\theta_{i} > \theta_{j}, \forall i \in \mathcal{R}_{2}^{-}, j\in \mathcal{R}_{3}^{+}$. Thus, $\varepsilon_{j} - \theta_{j} > \mid \theta_{i} - 1 \mid$, for all $i \in \mathcal{R}_{2}^{-}, j \in \mathcal{R}_{3}^{+}$ and we can claim that $V_{1} - V_{2} >0$, which contradicts the optimality assumption of $(\vec{\check{\mathbf{x}}},\vec{\check{\mathbf{c}}})$.

\textit{Case 2: $ \dot{n}_{r}^{s}\ge \mid \mathcal{S} \mid$.} In this case, all hexagons can be divided into two categories: 
\begin{align*}
    \mathcal{R}_1 & = \{j \mid \left \lfloor \frac{d_{r,j}^{k}}{\sum_{i \in \mathcal{S}} d_{r,i}^{k}} \times n_{r}^{s} \right \rfloor \ge n_{r,j}^{d} \} \\
    \mathcal{R}_2 & = \{j \mid \left \lfloor \frac{d_{r,j}^{k}}{\sum_{i \in \mathcal{S}} d_{r,i}^{k}} \times n_{r}^{s} \right \rfloor < n_{r,j}^{d} \}  
\end{align*}
The matching vehicle allocation $\vec{\mathbf{x}}^{*}$ obtained by Algorithm \ref{middle_algo} and the corresponding $\vec{\mathbf{c}}^{*}$, can be expressed as follows:
\begin{itemize}
    \item For $j \in \mathcal{R}_1, x_{r,j}^{*} = n_{r,j}^{d}, c_{r,j}^{*} =  \frac{d_{rj}^{k}}{\sum_{i \in \mathcal{S}} d_{r,i}^{k}} \times n_{r}^{s} - n_{r,j}^{d} $
    \item For $j \in \mathcal{R}_{2}, x_{r,j}^{*} \ge \left \lceil \frac{d_{r,j}^{k}}{\sum_{i \in \mathcal{S}} d_{r,i}^{k}} \times n_{r}^{s} \right \rceil, c_{r,j}^{*} = 0$.
\end{itemize}

For any given $(\vec{\mathbf{x}},\vec{\mathbf{c}})$, constraint \eqref{eq29} in MVA model imposes a restriction on $c_{r,j}$ such that $c_{r,j}  \ge  \frac{d_{r,j}^{k}}{\sum_{i \in \mathcal{S}} d_{r,i}^{k}} \times n_{r}^{s} - x_{r,j}$. In the case of hexagons within $\mathcal{R}_{1}$, the maximum value of $x_{r,j}$ is limited to $n_{r,j}^{d}$, and $c_{r,j}^{*}$ reaches its lower bound $\frac{d_{r,j}^{k}}{\sum_{i \in \mathcal{S}} d_{r,i}^{k}} \times n_{r}^{s} - n_{r,j}^{d}$. Additionally, the constraint \eqref{eq32} enforces that $c_{r,j}^{*}$, where $j \in \mathcal{R}_2$, attains its minimum possible value. Hence, we can assert that the minimum value of $\sum_{j \in \mathcal{R}} c_{r,j}$ is attained at the solution $(\vec{\mathbf{x}}^{*},\vec{\mathbf{c}}^{*})$.

Based on the aforementioned analysis, we can conclusively affirm that Algorithm \ref{middle_algo} is capable of attaining the optimal solution. Among the steps involved in Algorithm \ref{middle_algo}, the most intricate one pertains to the sorting the values of $\theta_j, j\in \mathcal{S}$. The worst case scenario arises when there exists an excess of unserved demands for each hexagon originating from hexagon $r$, $\mid \mathcal{S} \mid = \mid \mathcal{R} \mid$. Thus, the computational complexity of this worst case amounts to $O(\mid \mathcal{R} \mid^{2})$ when the bubble sort is employed. \hfill $\square$

\subsection{Proof of Proposition \ref{execution_pro}}

In the extended graph $\mathcal{G} (\mathcal{V}_{r} \cup \mathcal{C}_r \cup \mathcal{Z}_r,E)$, the cost associated with the edge $(v_{q}^{r}, c_{p}^{r})$ is computed as $w_{q,p}^{r}\times G(\nu_{p}^{r})$. An edge connects the customer $c_{p}^{r}$ to her destination node $z_{\theta_{p}^{r}}^{r}$, and the cost of edge $(c_{p}^{r},z_{\theta_{p}^{r}}^{r})$ is set to zero. No edge exists between customer $c_{p}^{r}$ and node $z_{k}^{r},\forall k \neq \theta_{p}^{r}, k\in \mathcal{R}$. The capacity of each edge is set to one unit. Within the extended graph $\mathcal{G}(\mathcal{V}_{r} \cup \mathcal{C}_r \cup \mathcal{Z}_{r},E)$, each node is associated with a value denoted as $b(i)$:
$$b(i) = \left\{
\begin{array}{rcl}
1, & & {i \in \mathcal{V}_r}\\
0, & & {i \in \mathcal{C}_r}\\
-x_{r,i}, & &{i \in \mathcal{Z}_r}
\end{array} \right.
$$

Let $y_{q,p}^{r}$ represent the flow on edge $(v_{q}^{r},c_{p}^{r})$, and $y_{p,k}^{r}$ denote the flow on edge $(c_{p}^{r},z_{k}^{r})$. The minimum cost flow problem in the extended graph can be expressed in its general form as follows. 

\begin{align}
    \min_{\vec{\bm{y}}} \quad & \sum_{c_{p}^{r} \in \mathcal{C}_{r}} \sum_{v_{q}^{r} \in \mathcal{V}_{r}} w_{q,p}^{r} \times \mathnormal{G}(\nu_{p}^{r}) \times y_{q,p}^{r} + \sum_{c_{p}^{r} \in \mathcal{C}_{r}} 0\times y_{p,\theta_{p}^{r}}^{r} \nonumber \\
\mbox{s.t.} \quad &  \sum_{c_{p}^{r} \in \mathcal{C}_r} y_{q,p}^{r} = 1 &\forall v_p^{r} \in \mathcal{V}_{r}\label{eq201} \\
 & \sum_{v_{q}^{r} \in \mathcal{V}_{r}} y_{q,p}^{r} - y_{p,\theta_{p}^{r}}^{r} = 0 & \forall c_{q}^{r} \in \mathcal{C}_{r} \label{eq202}\\
 & - \sum_{c_{p}^{r}:(c_{p}^{r},z_{k}^{r}) \in E} y_{p,k}^{r} = - x_{r,k} & \forall z_{k}^{r} \in \mathcal{Z}_{r} \label{eq203} \\
 & 0\le y_{q,p}^{r} \le 1, 0 \le y_{p,k}^{r} \le 1 & \forall v_{q}^{r} \in \mathcal{V}_{r}, c_{p}^{r}\in \mathcal{C}_{r}, z_{k}^{r} \in \mathcal{Z}_{r} \label{eq204} 
\end{align}

In the aforementioned minimum cost flow problem, the objective function and constraint \eqref{eq201} are the same as those in VOM. Constraint \eqref{eq202} stipulates that $\sum_{v_{q}^{r} \in \mathcal{V}_{r}} y_{q,p}^{r} = y_{p,\theta_{p}^{r}}^{r} \le 1$, which is equivalent to constraint \eqref{eq34}. By combining the flow conservation constraint \eqref{eq202} and constraint \eqref{eq203}, we can deduce that $x_{r,k} = \sum_{c_{p}^{r}:(c_{p}^{r},z_{k}^{r}) \in E} y_{p,k}^{r} =  \sum_{c_{p}^{r} \in \mathcal{C}_{r}} y_{p,k}^{r}\mathbf{I}(\theta_{p}^{r} = k) =\sum_{c_{p}^{r} \in \mathcal{C}_{r}} (\sum_{v_{q}^{r}\in \mathcal{V}_{r}} y
_{q,p}^{r})\mathbf{I}(\theta_{p}^{r} = k) = \sum_{v_{q}^{r} \in \mathcal{V}_{r}}\sum_{c_{p}^{r} \in \mathcal{C}_{r}} y_{q,p}^{r}\mathbf{I}(\theta_{p}^{r}=k)$. Note that this expression is analogous to constraint  \eqref{eq36}. Thus, it can be asserted that the linear-programming relaxation of VOM is equivalent to the minimum cost flow problem in the extended graph $\mathcal{G}(\mathcal{V}_{r} \cup \mathcal{C}_{r} \cup \mathcal{Z}_{r},E)$. 

Subsequently, we establish the equivalence between the optimal solution of VOM and that of its linear-programming relaxation. The node-edge incidence matrix exhibits the property of total unimodularity for where every square non-singular submatrix has a determinant of 0,+1, or -1 \citep{schrijver2003combinatorial}. This property ensures that for any problem instance with integer RHS values for all constraints, its optimal solutions are integer \citep{papadimitriou1998combinatorial}. Thus, the optimal solution of the minimum cost flow problem in the extended graph $\mathcal{G}(\mathcal{V}_{r} \cup \mathcal{C}_{r} \cup \mathcal{Z}_{r},E)$ is binary, indicating that the optimal solution of VOM can be obtained by solving its linear-programming relaxation problem. \hfill $\square$

\section{Additional illustration for the MMA framework}
\label{addi_PEMA}
\setcounter{table}{0}
\renewcommand{\thetable}{C-\arabic{table}}
\renewcommand*{\theHtable}{\thetable}
\setcounter{figure}{0}
\renewcommand{\thefigure}{C-\arabic{figure}}
\renewcommand*{\theHfigure}{\thefigure}

\subsection{Toy Example to Illustrate Two-step Mechanism in Execution Layer}
\label{toy_illu}

This subsection presents a toy example involving three regions $A,B,C$ to provide a clearer illustration of the core idea underlying the two-step mechanism. We suppose that the strategic layer suggests that ten vehicles to serve the passengers from $A$ to $A$, four vehicles from $A$ to $B$ and six vehicles from $A$ to $C$ in interval $k$. Before the current matching interval, we assume that one demand from $A$ to $A$, one demand from $A$ to $B$, and three demands from $A$ to $C$ have been successfully fulfilled. At the end of the current matching interval, there are 30, 20, and 20 demands heading for $A$, $B$, and $C$ waiting to be matched, while there are only five vacant vehicles in $A$. In such a scenario, we leverage the two-step mechanism to assign these five vehicles to the 70 waiting demands. The first step determines the number of vehicles to be matched with the demands from $A$ to each region in the current matching interval. This allocation is proportional to the uncompleted targets and may not exceed the number of waiting demands, ensuring that the cumulative number of matched demands in interval $k$ approximates the strategic guidance (see Table \ref{Illustration_for_first_step} for an illustration). Thus, the system assigns three vehicles to serve the passengers traveling to $A$, one vehicle to serve a passenger to $B$, and one vehicle to pick up a passenger to $C$ in the current matching interval. These results impose constraints on the batch matching in the second step, which jointly determine the specific vehicle-order assignment with the objective of minimizing the weighted pick-up time. 

\begin{table}[!ht]
  \footnotesize
  \centering
  \caption{Illustration of the first step in two-step mechanism.}
  \begin{tabularx}{\textwidth}{m{2.1cm}m{2.1cm}m{2.3 cm}m{2.6cm}m{2cm}m{3.4cm}}
    \hline
    \textbf{Destination} & \textbf{Target $M_{A,j}^{k}$} &\textbf{Completed targets} &\textbf{Uncompleted target} &\textbf{Waiting passengers} & \textbf{Allocation results}\\
    \hline
    A & 10 & 1 & 9 & 30 & $5 \times \frac{9}{9+3+3} =3$\\
    B & 4 & 1 & 3 & 20 & $5 \times \frac{3}{9+3+3} =1$\\
C & 6 & 3 & 3 & 20 & $5 \times \frac{3}{9+3+3} =1$ \\
    \hline
  \end{tabularx}
\label{Illustration_for_first_step}
\end{table}

\subsection{Greedy Algorithm for Relocation}
\label{greedy_relo}
This subsection presents a simple greedy algorithm to determine the number of vehicles relocating from hexagon $r$ to $j$, denoted as $z_{r,j}$, in accordance with the guidance provided by the strategic layer at the end of strategic interval $k$.

\begin{algorithm}[h] 
    \footnotesize
    \caption{The greedy algorithm to determine $z_{r,j}, \forall j \in \mathcal{R}$} 
    \label{greedy_algo} 
    \begin{algorithmic}[1] 
    \REQUIRE ~~\\ 
    The suggested relocation strategies $E_{r,j}^{k}, \forall j\in\mathcal{R}$\\
    The number of vacant vehicles in hexagon $r$, $l_{k,r}^{s}$\\
    \ENSURE The number of relocated vehicles from hexagon $r$ to $j$, $z_{r,j}$\\ 
    \STATE Initialization: $z_{r,j}$ = $\left \lfloor  \frac{E_{r,j}^{k}}{\sum_{i \in \mathcal{R}}E_{r,i}^{k}}  \times l_{k,r}^{s} \right\rfloor$
    \STATE Calculate the number of available vehicles after pre-assignment, which is denoted by $ n = l_{k,r}^{s} - \sum_{j\in \mathcal{R}}z_{r,j}$
    \STATE For $n$ regions with top $n$ largest $E_{r,j}^{k}$, update $z_{r,j} = z_{r,j} + 1$
    \end{algorithmic}
\end{algorithm}

\section{Supplement of Numerical Studies}
\label{Addi_numerical}
\setcounter{table}{0}
\renewcommand{\thetable}{D-\arabic{table}}
\renewcommand*{\theHtable}{\thetable}
\setcounter{figure}{0}
\renewcommand{\thefigure}{D-\arabic{figure}}
\renewcommand*{\theHfigure}{\thefigure}

\subsection{Supplement of Toy Model}
\label{addi_toy_res}
\subsubsection{Experiment Design of Toy Network.}
\label{toy_model_design}
\paragraph{}
We design a toy network to demonstrate the effectiveness of the MMA framework in addressing the spatiotemporal order dispatching problem. The toy network consists of three regions  $A,B,C$ which are square-shaped regions centered on $(\frac{3}{2},\frac{3}{2}),(\frac{3}{2},\frac{19}{2}),(\frac{27}{2},\frac{3}{2})$ with a side length of three. The pick-up distance and the travel distance are measured using the Euclidean distance with a detour ratio of 1.3. The vehicle speed remains constant at 30km/h throughout the simulation. An order will be canceled if not being served to any drivers for a long time, with the cancellation time modeled as an exponential distribution with time-varying mean, denoted as $\varphi_{t}^{d},\forall t\in \mathcal{T}$. Each driver is initialized according to the initial location and the time of its first entry into the platform. The driver's subsequent actions are completely determined by assignments, idle movements, and offline behaviors. We assume that vacant vehicles remain at the drop-off location of the last served request without cruising. A driver may decide to go offline if being vacant for a long time. The maximum waiting time for drivers is modeled by an exponential distribution with time-varying mean  $\varphi_{t}^{s},\forall t \in \mathcal{T}$. 

For data generation, we aim to simulate a scenario with imbalanced demand and supply. Therefore, the order request generation intervals and the intervals of drivers' first entry into the platform are sampled from the two-component mixture of Gaussians and then truncated to integers. The specific generation times are uniformly sampled within their corresponding interval. Moreover, the origins and destinations of orders, as well as the initial locations of drivers, are randomly sampled from a uniform distribution defined on the respective regions. The mathematical expression of the Gaussian mixture distribution is shown in Equation \eqref{eq40}:
\begin{align}
    p(x \lvert \eta) = \epsilon_{1}p_{1}(x\lvert \mu_{1},\sigma_{1}^{2}) + \epsilon_{2}p_{2}(x\lvert \mu_{2},\sigma_{2}^{2}) \label{eq40}
\end{align}
where $\eta = (\epsilon_{1},\epsilon_{2},\mu_{1},\sigma_{1},\mu_{2},\sigma_{2})$, and $\epsilon_{1} + \epsilon_{2} = 1$. Afterwards, an imbalanced transition probability matrix, denoted as $P = (p_{i \rightarrow j})$ is designed, where $p_{i \rightarrow j}$ represents the proportion of generated requests heading for region $j$ from $i$ relative to the total generated requests in region $i$. The setting parameters are shown in Table \ref{table:design_parameter}. 

\begin{table}[!ht]
  \footnotesize
  \centering
  \renewcommand\arraystretch{1.2}
  \caption{Numerical parameters.}
    \begin{tabularx}{\textwidth}{m{2.6cm}m{1.9cm}m{2.6cm}m{2.2cm}m{2cm}m{4.0cm}}
    \hline
    \multicolumn{6}{c}{\textbf{The parameters in the toy model}}\\
    \hline
    \multirow{4}*{Demand data} & Region & Demand quantity & \multicolumn{2}{l}{Generation interval distribution} & Transition probability\\
    & A & 5000 & \multicolumn{2}{l}{(0.5,0.5,30,30,100,30)} & (0.2,0.3,0.5) \\
    & B & 5000 & \multicolumn{2}{l}{(0.5,0.5,96,60,126,60)} & (0.3,0.2,0.5) \\
    & C & 5000 & \multicolumn{2}{l}{(0.7,0.3,36,30,96,50)} & (0.2,0.2,0.6) \\
    \hline
    \multirow{4}*{Supply data} & Region & Supply quantity & \multicolumn{2}{l}{Generation interval distribution} \\
    & A & 300 & \multicolumn{3}{l}{(0.7,0.3,36,20,108,20)}\\
    & B & 300 & \multicolumn{3}{l}{(0.5,0.5,50,20,108,20)}\\
    & C & 300 & \multicolumn{3}{l}{(0.7,0.3,43,20,108,20)}\\
    \hline
    \multirow{2}*{System Parameters} & Vehicle speed & Planning intervals & \multicolumn{2}{l}{Length of a matching interval}& Length of a strategic interval  \\
    & 30km/h & 9 & \multicolumn{2}{l}{10 seconds} & 10 minutes \\
    \hline
    Offline behavior & 0:00 -- 6:00  & 6:00 -- 10:00& 10:00 --17:00 & 17:00 -- 21:00 & 21:00 --24:00 \\
    $\varphi_{t}^{s}$  (s) & $1200$ & $1800$ & $900$ &$1800$ &$1200$ \\
    $\varphi_{t}^{d}$  (s) & $1500$ & $1200$ & $1800$ &$1200$ &$1500$ \\
    \hline
    \end{tabularx}\label{table:design_parameter}
\end{table}



\subsubsection{Sensitivity Analysis of Hyper Parameters.}
\label{sensi_toy_model}
\paragraph{}
The strategic-layer decision-making process in the MMA framework aims to optimize the number of completed requests, the relocation cost, and the penalty of region imbalance. To facilitate the optimization, the multi-objective optimization is transformed into a single-objective optimization by the weighted sum method with weight 1, $\alpha$, and $\beta$. The selection of $\alpha$ and $\beta$ has a significant impact on both platform revenue and relocation times. This subsection performs sensitivity analyses on $\alpha$ and $\beta$ under the toy model. Figure \ref{fig:toy_sensitivity analysis} depicts the impact of parameters on the major measures.

Parameter $\alpha$ represents the relative importance assigned to the relocation cost in comparison to the number of completed requests. An increase in $\alpha$ leads to a decrease in relocation times and vehicle utilization. Consequently, the number of completed requests decreases and the pick-up distance raises with the increment of $\alpha$. Meanwhile, the parameter $\beta$ denotes the penalty weight for regional imbalance. A larger $\beta$ tends to balance the regional gap between demand and supply across different regions by relocating vacant vehicles from the oversupplied region to the undersupplied region in anticipation of future demand. As demonstrated in Figure \ref{fig:toy_sensitivity analysis}, a significant increase in the number of completed requests is observed with higher $\beta$ values.

\begin{figure}[!htb]
\centering
\subfigure[Variation of requests completion] {\includegraphics[width=5.12cm]{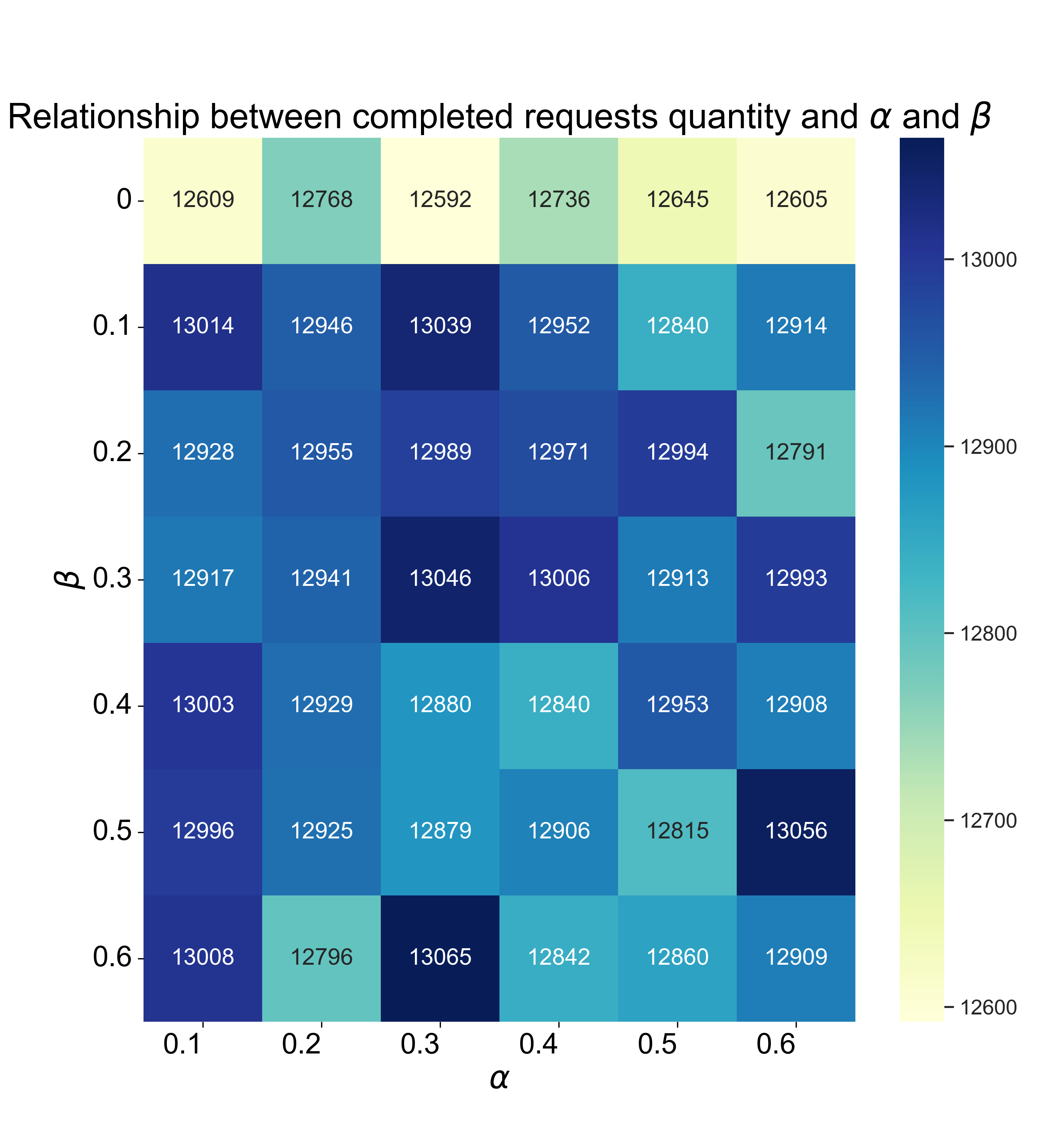}} 
\subfigure[Variation of relocation times] {\includegraphics[width=5.12cm]{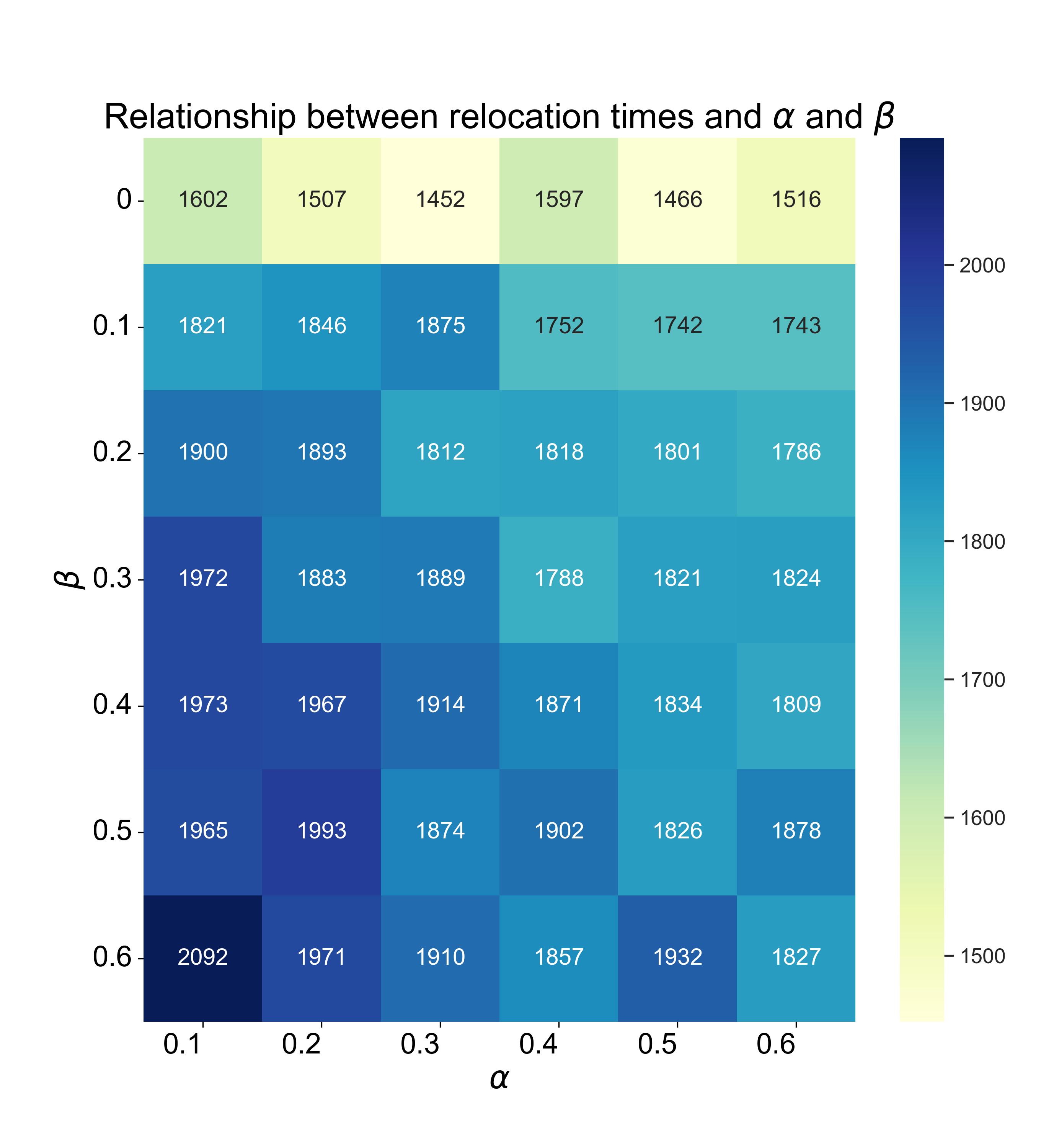}}
\subfigure[Variation of pick-up distance] {\includegraphics[width=5.12cm]{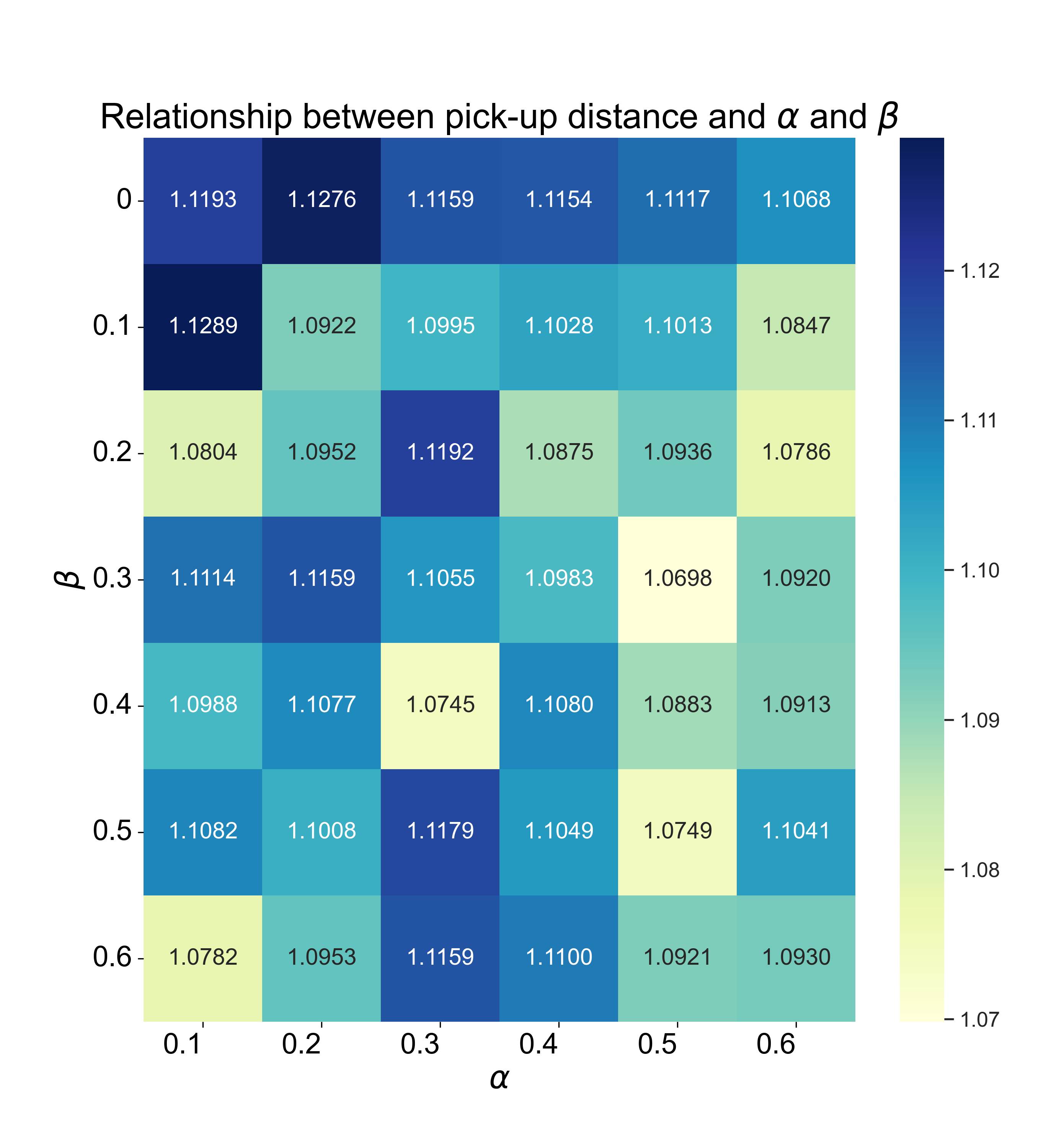}} 
\caption{The impact of $\alpha$ and $\beta$ on the major measures.} 
\label{fig:toy_sensitivity analysis}  
\end{figure}

\subsection{Data Analysis of Realistic Network.}
\label{addi_real_net}

The real-world dataset encompasses historical vehicle data from January $1^{st}$ to $7^{th}$, 2021, and demand data from September $1^{st}$ to November $17^{th}$, 2020. The demand data comprises information including order ID, generation time, origin and destination coordinates. The vehicle trajectory data includes driver ID, timestamp, latitude and longitude, as well as the relevant state (offline, vacant, and occupied). The raw data analysis serves two primary purposes: the first one is to simulate and replicate the behavior of requests and drivers in the ride-hailing simulator, and the second one is to construct the prediction model applied in the strategic-layer. 

To facilitate the evaluation of the MMA framework, a ride-hailing simulator is developed. We employ Uber H3 hexagonal hierarchical index grid system \citep{uber} to partition the city into hexagonal grids with an approximate side length of 3.229 kilometers. The spatial distribution of raw demands exhibits sparsity, with approximately 65.37$\%$ of requests concentrated within the central 19 hexagons. Therefore, our focus lies on these densely populated 19 hexagons, which offer sufficient grounds for verifying the effectiveness of the proposed framework. Figure \ref{fig:demand_data_description}(a) displays the urban area of Chengdu, wherein the red hexagons represent the studied area in the realistic network experiment. Furthermore, Figure \ref{fig:demand_data_description}(b) depicts the spatial distribution of demand origins, highlighting the prevailing spatial imbalance among the hexagons. 

\begin{figure}[!htb]
\centering
\subfigure[Operational hexagons in simulator] {\includegraphics[width=7cm]{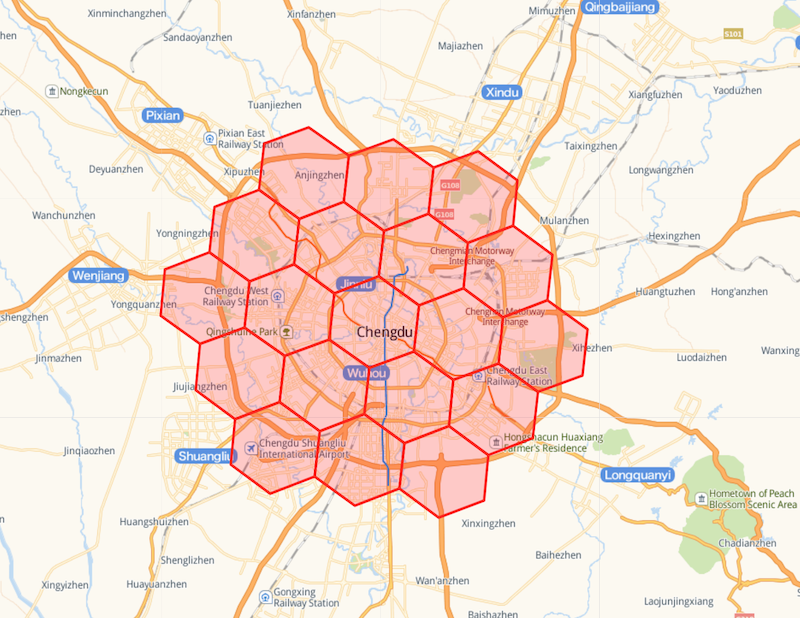}} \hspace{6mm}
\subfigure[Spatial distribution of the requests on one day] {\includegraphics[width=7cm]{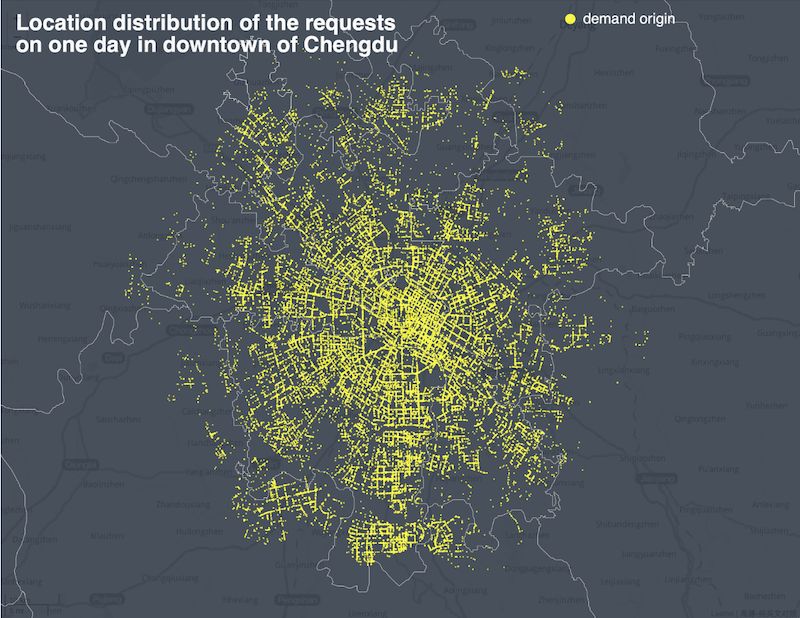}}
\caption{The operational hexagons and spatial distribution of demand origins.} 
\label{fig:demand_data_description}
\end{figure}

The raw supply data undergoes an extensive analysis to address the challenges posed by the large volume of records and the intricacies of driver behavior. Upon scrutinizing the online time and the number of completed requests per vehicle, it becomes evident that a notable proportion of vehicles exhibit abnormal behavior. For instance, some drivers are rarely online or fail to serve vehicles despite their availability. We employ the following criteria to identify and exclude drivers who are incapable of providing effective service: (1) the total online time is less than 1,000 minutes from January $1^{st}$ to $7^{th}$, 2021 and (2) the occupied time does not exceed 20$\%$ of the total online time. The remaining 27.19$\%$ of drivers, referred to \textit{loyal drivers}, are responsible for serving 80.30$\%$ of the completed requests. In our simulation, we randomly select the appropriate number of vehicles from the pool of loyal drivers so that the simulator's request completion rate is close to the actual completion rate under the same matching strategy. The initial locations and first entry times of drivers are bootstrapped from the historical records of loyal drivers, while the subsequent movements of drivers depend on the matching and relocation algorithm. Following the centralized strategy, drivers currently occupied proceed toward the destination of their assigned request, while others either remain in their current position or relocate to the target hexagon. The simulation also encompasses the departure behavior of passengers and drivers. In the absence of available data, we introduce the following assumptions to ensure congruence with reality: (1) drivers are subject to a maximum working duration of 12 hours per day, and (2) the maximum waiting times for both drivers and passengers are modeled by an exponential distribution with time-dependent parameters that are estimated through maximum likelihood estimation.

Meanwhile, the MMA framework adopts a rolling horizon mechanism to prevent myopic decision-making by incorporating explicit forecast information. In our simulation, we employ a series of Lasso regression models with massive explanatory variables, as stated in Section \ref{strategic}, to predict the number of new-emerging vehicles and demands. For the demand prediction models, we use data from September $1^{st}$ to October $31^{st}$, 2020 as a training dataset to estimate the parameters in prediction models. In addition, we construct the training and test datasets for supply prediction model by bootstrapping historical supply data. It is worth noting that the selection of prediction models is not limited to a specific method and should be tailored to the available data. We use \textit{Error Rate} (ER), as introduced by \citet{Tong2017}, to assess the prediction performance.
\begin{align}
    \mathrm{ER} & = \frac{\sum_{i=1}^{N} \lvert p_{i}-y_{i} \rvert}{\sum_{i=1}^{N} y_{i}} 
\end{align}
Here, $N$ is the size of the test dataset, while $p_{i}$ and $y_{i}$ represent the estimation and the ground truth of the $i^{th}$ sample, respectively. Table \ref{table:prediction_res} summarizes the prediction results, indicating that predictions are less accurate for intervals further beyond the current strategic interval. The strategic layer leverages the prediction information to devise strategies for the subsequent nine intervals, with only the strategy for the current interval being implemented. As time progresses, the system undergoes re-prediction to formulate strategies for future intervals. Consequently, inaccurate predictions for future intervals exert limited influence on the strategy formulation process.

\begin{table}[!ht]
    \centering
    \renewcommand\arraystretch{1.2}
    \caption{Prediction results of demand and supply.}
    \footnotesize
    \begin{tabular}{llllllll}
    \hline
    \multirow{2}*{} & \multicolumn{3}{c}{Demand} & & \multicolumn{3}{c}{Supply} \\
    \cline{2-4} 
    \cline{6-8}
         &  Train $R^{2}$ &  Test $R^{2}$ & ER & & Train $R^{2}$ & Test $R^{2}$ & ER\\
    \hline
        Next one horizon &  0.9599 & 0.9583 & 0.1708 & & 0.7702 & 0.7710 & 0.5391\\ 
        Next two horizon & 0.9392 & 0.9487 & 0.1828 & & 0.7623 & 0.7606 & 0.5520\\
        Next three horizon & 0.9171 & 0.9397 & 0.1915 & &0.7583 & 0.7557 & 0.5604 \\
        Next four horizon & 0.8962 & 0.9304 & 0.1973 & & 0.7548 & 0.7562 & 0.5622 \\
        Next five horizon & 0.8750 & 0.9229 & 0.2018 & & 0.7529 & 0.7567 & 0.5664\\
        Next six horizon & 0.8565 & 0.9146 & 0.2070 & & 0.7512 & 0.7558 & 0.5667 \\
        Next seven horizon & 0.8394 & 0.9075 & 0.2115 & & 0.7503 & 0.7558 & 0.5694 \\
        Next eight horizon & 0.8240 & 0.9012 & 0.2144 & & 0.7488 & 0.7553 & 0.5703 \\
        Next nine horizon & 0.8110 & 0.8957 & 0.2177 & & 0.5186 & 0.5098 & 0.6874 \\
    \hline
    \end{tabular}
    \label{table:prediction_res}
\end{table}


%
%
%

\end{APPENDICES}


\end{document}